\documentclass[10pt,twocolumn,letterpaper]{article}

\usepackage[pagenumbers]{iccv}      % To produce the REVIEW version

\usepackage{amsmath,amsfonts}
\usepackage{algorithmic}
\usepackage{array}
\usepackage{textcomp}
\usepackage{stfloats}
\usepackage{url}
\usepackage{verbatim}
\usepackage{graphicx}
\usepackage{amssymb}
\usepackage{bm}
\usepackage{multirow}
\usepackage{color}
\usepackage{tabularx}
\usepackage{float}
\usepackage{fdsymbol}
\usepackage[switch]{lineno}
\usepackage{bbding}
\usepackage{booktabs}
\usepackage{lipsum}
\usepackage[accsupp]{axessibility}

\newcommand\blfootnote[1]{%
\begingroup
\renewcommand\thefootnote{}\footnote{#1}%
\addtocounter{footnote}{-1}%
\endgroup
}
\definecolor{cvprblue}{rgb}{0.21,0.49,0.74}
\usepackage[pagebackref,breaklinks,colorlinks,allcolors=cvprblue]{hyperref}

\def\paperID{8294} % *** Enter the Paper ID here
\def\confName{ICCV}
\def\confYear{2025}
\def\R{\mathbb{R}}
\title{Benchmarking and Learning Multi-Dimensional Quality Evaluator \\ for Text-to-3D Generation}

\author{Yujie Zhang$^{1*}$ \quad Bingyang Cui$^{1*}$ \quad Qi Yang$^{2}$ \quad Zhu Li$^{2}$ \quad Yiling Xu$^{1\dagger}$ \\
$^{1}$ Shanghai Jiao Tong University\quad \textsuperscript{2} University of Missouri-Kansas City  \\
{\tt\small $^{1}$\{yujie19981026, ccccby0813, yl.xu\}@sjtu.edu.cn , $^{2}$\{qiyang, lizhu\}@umkc.edu}}
\begin{document}
\maketitle
\pagestyle{empty}
\thispagestyle{empty}

\begin{abstract}
Text-to-3D generation has achieved remarkable progress in recent years, yet evaluating these methods remains challenging for two reasons: i) Existing benchmarks lack fine-grained evaluation on different prompt categories and evaluation dimensions. ii) Previous evaluation metrics only focus on a single aspect (\textit{e.g.,} text-3D alignment) and fail to perform multi-dimensional quality assessment. To address these problems, we first propose a comprehensive benchmark named MATE-3D. The benchmark contains eight well-designed prompt categories that cover single and multiple object generation, resulting in 1,280 generated textured meshes. We have conducted a large-scale subjective experiment from four different evaluation dimensions and collected 107,520 annotations, followed by detailed analyses of the results. Based on MATE-3D, we propose a novel quality evaluator named HyperScore. Utilizing hypernetwork to generate specified mapping functions for each evaluation dimension, our metric can effectively perform multi-dimensional quality assessment. HyperScore presents superior performance over existing metrics on MATE-3D, making it a promising metric for assessing and improving text-to-3D generation. The project is available at \url{https://mate-3d.github.io/}.

\end{abstract}

\blfootnote{* Equal Contribution.}
\blfootnote{$\dagger$ Corresponding Author.}

\vspace{-0.3cm}
\section{Introduction}\label{sec:introduction}

Remarkable advances in text-to-3D generative methods have been witnessed in recent years \cite{liu2024comprehensive}. Given textual descriptions (\textit{i.e.}, prompts), these methods have the capability to generate contextually relevant 3D representations, such as textured mesh \cite{textmesh}, Neural Radiance Field (NeRF) \cite{he2024mmpi,huang2023nerftexture,mildenhall2021nerf}, and 3D Gaussian Splatting (3DGS) \cite{kerbl20233dgs,ren2023dreamgaussian4d,ben2024dreamgaussian,yi2024gaussiandreamer}. Unlike traditional distortions in 3D data processing (\textit{e.g.}, noise, lossy compression), generated 3D representations may suffer from some unique degradations, such as misalignment with prompt semantics and multi-view inconsistency (\textit{i.e.}, the Janus problem) \cite{hong2023debiasing,dreamfusion}. To better guide the method design and support fair comparisons, it is important to explore subjective-aligned quality evaluation for text-to-3D generation methods.

\begin{figure}[t]
    \centering
    \begin{subfigure}[t]{0.5\textwidth}
        \centering
        \includegraphics[width=0.8\linewidth]{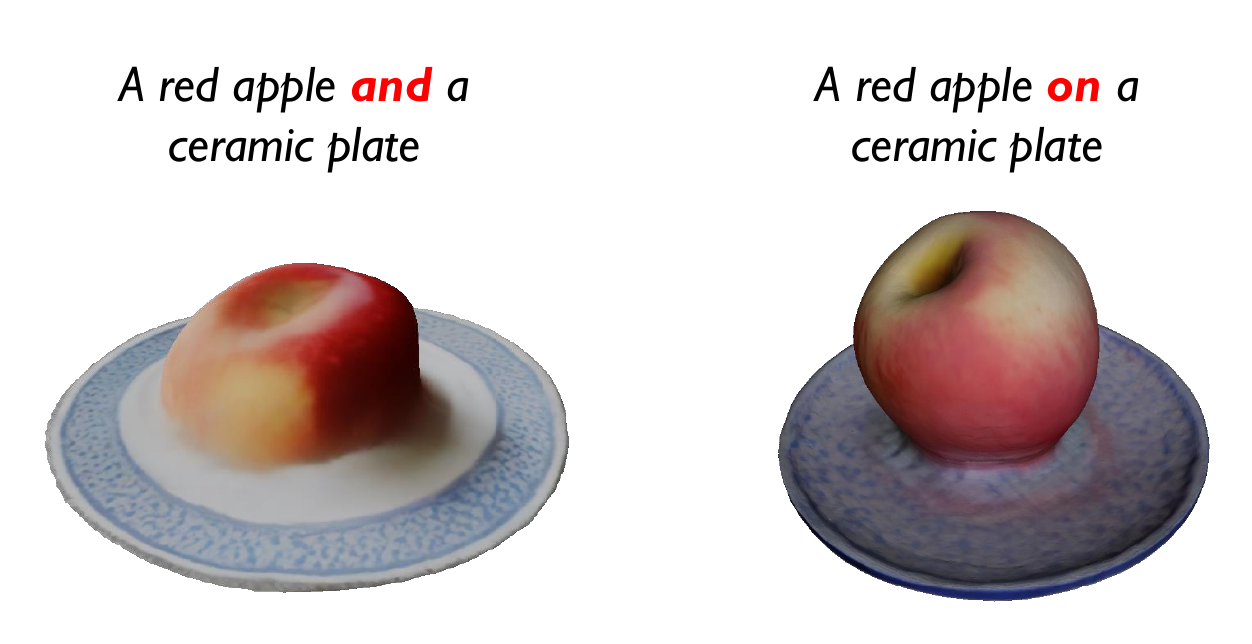}
        \caption{}
        \label{fig:teaser1}
    \end{subfigure}  

    \begin{subfigure}[t]{0.5\textwidth}
        \centering
        \includegraphics[width=\linewidth]{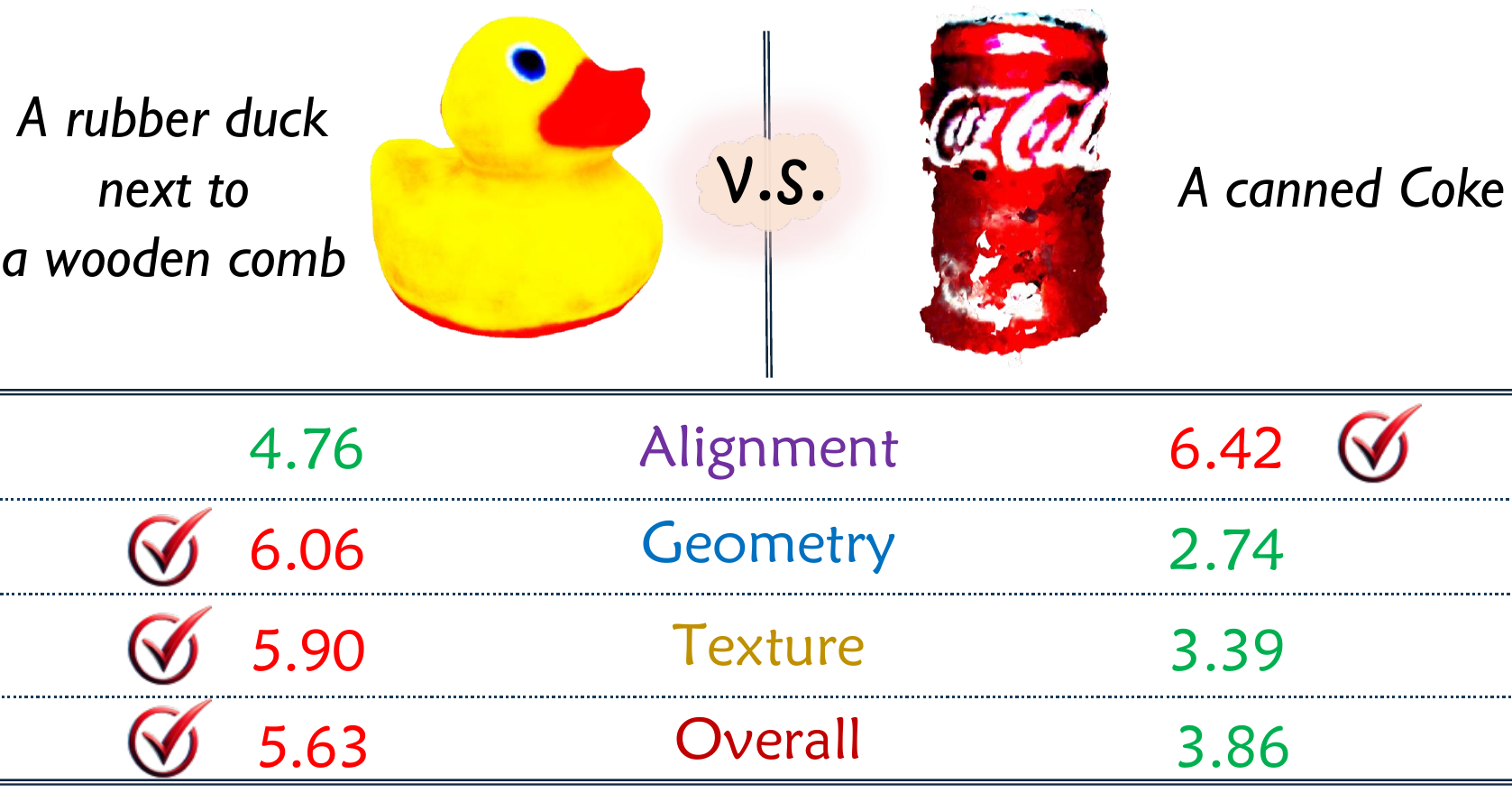}
        \caption{}
        \label{fig:teaser2}
    \end{subfigure} 
    \caption{(a) Diverse generation results from similar prompts. (b) Quality evaluation from multiple perspectives.}
    \label{fig:teaser}
    \vspace{-0.4cm}
\end{figure}

Previous works have conducted text-to-3D evaluation by proposing new benchmarks. For example, $\mathrm{T}^3$Bench \cite{he2023t3bench} constructs three prompt classes with increasing scene complexity, and \textit{Multiple objects}, followed by collecting 630 texture meshes annotated from two dimensions, \textit{i.e.}, quality and alignment. However, these benchmarks have two main drawbacks. i) \textbf{Lack of Prompt Diversity}. Although these benchmarks \cite{he2023t3bench, hui2024makeashape, wu2024gpt4v} have classified prompts based on textual complexity or object quantity, their categorization generally lacks an adequate degree of distinction. In fact, we observe that some similar prompts usually lead to visually different results, as shown in \cref{fig:teaser1}.
Therefore, a fine-grained categorization is essential to gain a comprehensive understanding of model capabilities. ii) \textbf{Limited Evaluation Dimension.} As shown in \cref{fig:teaser2}, the overall quality of the generated 3D representation is influenced by different factors, including geometry fidelity, texture details, and text-3D alignment. A single score is insufficient to perform detailed comparison between different cases. 
%Although some research has conducted multi-dimensional evaluation, they only select preferences for sample pairs generated from the same prompt, which brings difficulty for comparison among samples originated from different prompts. 
Consequently, it is necessary to construct a comprehensive benchmark with higher prompt diversity and richer evaluation dimensions.

Another challenge in text-to-3D evaluation is the lack of accurate and automated metrics to assess the quality of generated results. Existing approaches \cite{wang2024prolificdreamer} rely primarily on subjective user studies, which are constrained by high costs in terms of time, money, and rigorous testing environment. Some previous work \cite{mohammad2022clipmesh,textmesh,xu2023dream3d} has attempted to address this by rendering the 3D objects into single or multiple 2D images and using CLIP- \cite{radford2021learning} or BLIP- \cite{li2022blip} based metrics \cite{hessel2021clipscore,schuhmann2022laion,xu2024imagereward,wu2023human} to measure the alignment between the prompts and the rendered images. However, these metrics focus on the text-3D correspondence, ignoring other critical aspects of subjective perception, such as geometry and texture fidelity. In fact, human subjects can dynamically adjust their focus and decision process based on the evaluation dimension (\textit{e.g}., prioritize shape and 
contour for geometry evaluation), resulting in diverse scores with changing perspectives. Consequently, single-dimensional evaluators may fail to capture rich quality-related factors.
To comprehensively evaluate text-to-3D generation, it is essential to develop robust metrics that reflect multi-dimensional human perception.

To solve the above problems, we establish a comprehensive benchmark named \textbf{M}ulti-Dimension\textbf{A}l \textbf{T}ext-to-3D Quality \textbf{E}valuation Benchmark (MATE-3D), which is the first contribution of this paper. MATE-3D first categorizes the prompts based on object quantity. For single object generation, we define four sub-categories based on the prompt complexity and creativity, including \textit{``Basic''}, \textit{``Refined''}, \textit{``Complex''} and \textit{``Fantastical''}. For multiple object generation, the prompts contain at least two objects with specified relationships between the objects.
Based on the type of relationships, we define four sub-categories denoted by \textit{``Grouped''}, \textit{``Spatial''}, \textit{``Action''}, and \textit{``Imaginative''}. Utilizing large language models, we design 160 prompts for the eight sub-categories in total. We use these prompts as the input of eight prevalent text-to-3D methods and acquire 1,280 generated samples, which are transformed into a unified representation, \textit{i.e.,} textured mesh. Then, we conduct a comprehensive subjective experiment and each textured mesh is annotated by 21 human subjects from four evaluation dimensions, including \textit{semantic alignment}, \textit{geometry quality}, \textit{texture quality}, and \textit{overall quality}. A total of $1,280\times4\times21=107,520$ annotations are collected, and we further obtain the Mean Opinion Score (MOS) for each evaluation dimension of the generated meshes. 
Finally, we conduct detailed analyses of subjective scores to provide useful insight for future text-to-3D research.

Furthermore, we propose a novel quality evaluator named HyperScore for text-to-3D generation, which can perform multi-dimensional assessment by utilizing a hypernetwork to generate specified mapping function for each evaluation dimension, resulting in the second contribution. More concretely, we first use a pre-trained CLIP model to extract visual and textual features from rendered images of textured meshes and prompts. Then, we create a sequence of learnable tokens and feed them into the
textual encoder to obtain multiple condition features that represent information relevant to evaluation dimensions. Using these condition features, we implement two strategies to achieve fine-grained quality evaluation across dimensions. i) 
We utilize the condition features to compute the fusion weights for various patches in the visual feature, which mimics the focus shift during evaluation. ii) We feed the condition features into a hypernetwork \cite{ha2022hypernetworks} to generate diverse weights of a mapping head, thereby simulating different decision-making processes aligned with changing evaluation dimensions. Experiments on MATE-3D  show that HyperScore outperforms existing metrics in all evaluation dimensions. 
We summarize the contributions as follows:
\begin{itemize}
    \item We establish a new benchmark, MATE-3D, for evaluating text-to-3D methods. MATE-3D contains 1,280 textured meshes generated from eight prompt categories and each sample is annotated from four evaluation dimensions.  
    \item We propose a multi-dimensional quality evaluator, HyperScore, for text-to-3D generation. Utilizing the learnable condition features, we achieve differentiated quality predictions for multiple evaluation dimensions. 
    \item Extensive experiments show that the proposed HyperScore presents superior performance on different evaluation dimensions than existing evaluation metrics.  

\end{itemize}

\begin{figure*}[t]
    \centering
    \begin{subfigure}[t]{0.25\textwidth}
        \centering
        \includegraphics[width=\linewidth]{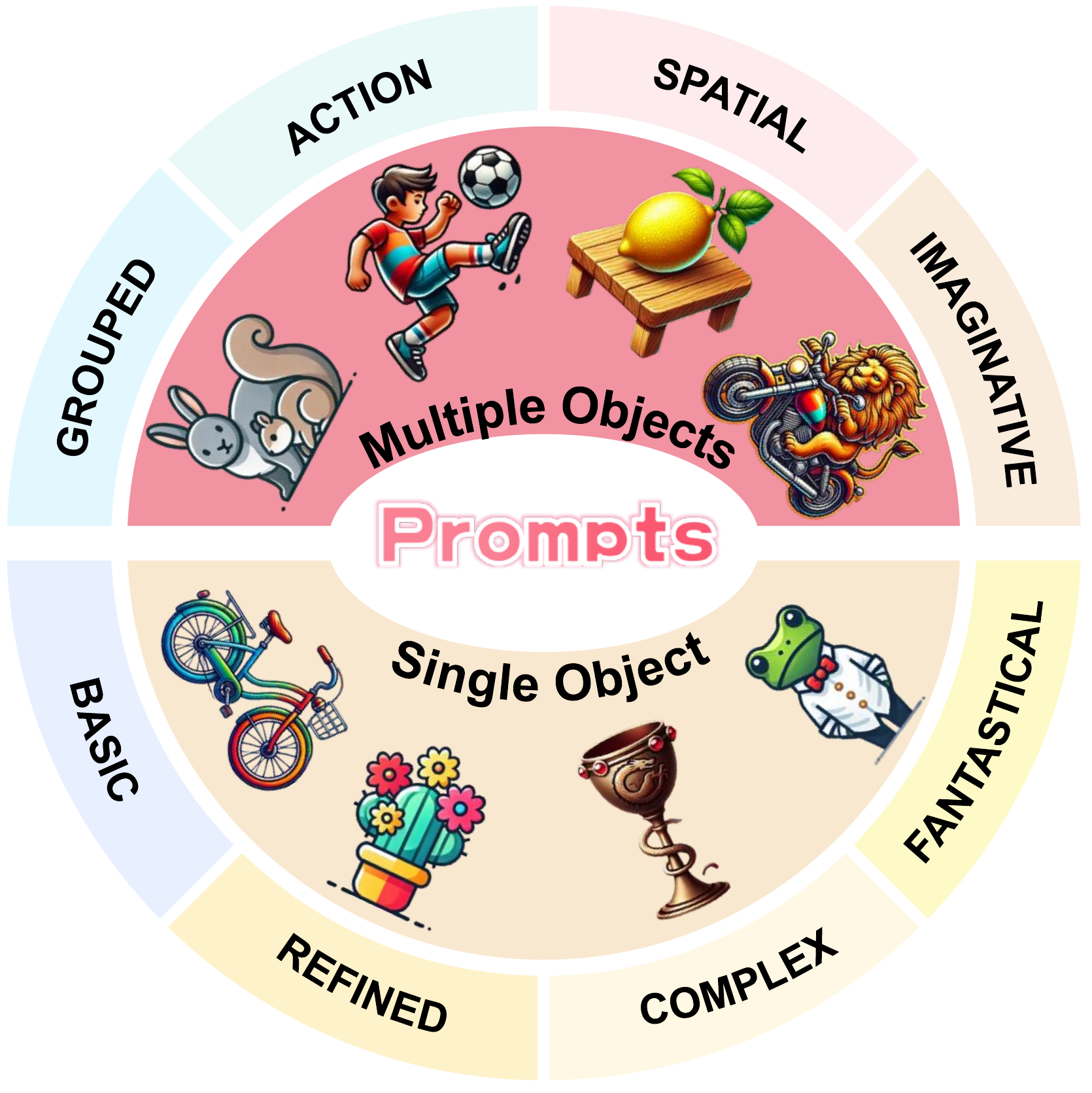}
        \caption{}
        \label{fig:challenge}
    \end{subfigure}  
    \hspace{0.01\textwidth} % 控制子图之间的水平间距
    \begin{subfigure}[t]{0.72\textwidth}
        \centering
        \includegraphics[width=\linewidth]{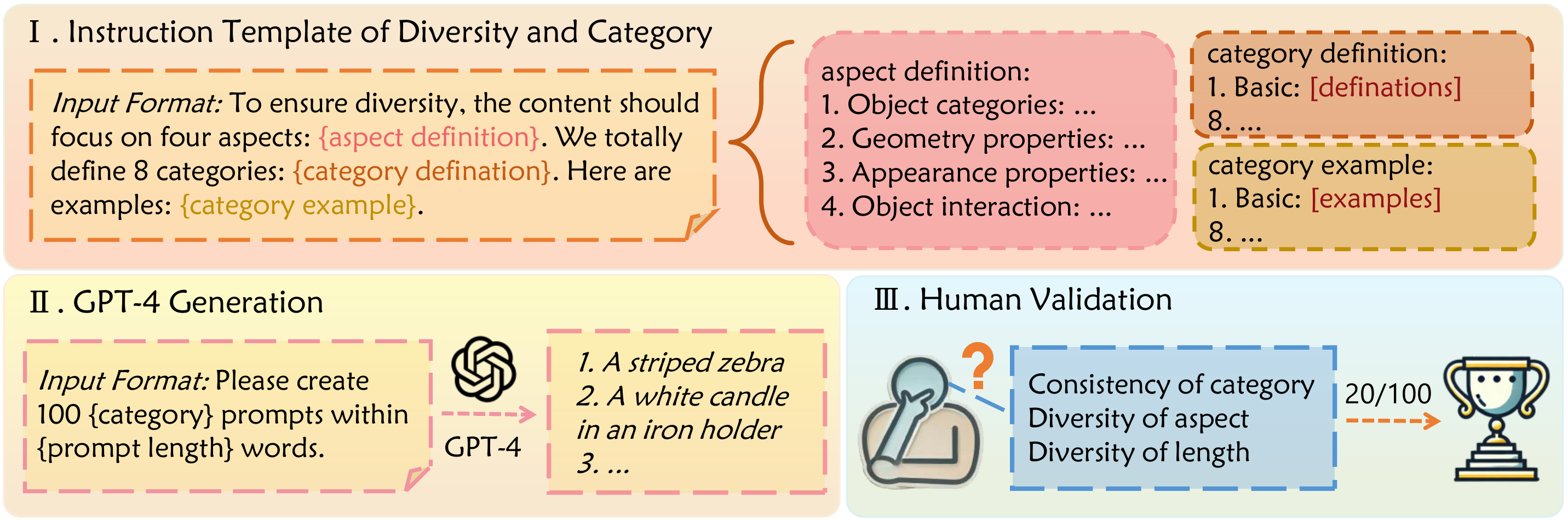}
        \caption{}
        \label{fig:prompts}
    \end{subfigure} 
    \caption{(a) The overview of eight prompt categories; (b) The illustration of prompt generation pipeline.}
    \label{fig:challenge_prompt}
    \vspace{-0.5cm}
\end{figure*}

\section{Related Works}\label{sec:related work}

\subsection{Text-to-3D Generation}

Generating 3D objects conditioned on texts has become a challenging yet prominent research area in recent years. With the emergence of large-scale 3D datasets \cite{deitke2023objaversexl,deitke2023objaverse}, some methods, such as Point-E \cite{nichol2022pointe} and Shap-E \cite{jun2023shape}, first train the network using 3D datasets and then generate 3D assets in a feed-forward manner \cite{zhang2024clay,fu2022shapecrafter,gao2022get3d,zhang20233dshape2vecset,hessel2021clipscore,hong2023lrm,chen2025lara}. Since the quantity of image-text pairs \cite{li2024aigiqa20k,liagiqa3k} is much larger than the 3D counterparts, Dreamfusion \cite{dreamfusion} pioneers the paradigm of optimizing 3D generation with pre-trained 2D diffusion models, which leverages Score Distillation Sampling (SDS) to optimize a NeRF. Based on this paradigm, some works \cite{magic3d,latentnerf,wang2024prolificdreamer,seo2023let,ma2024scaledreamer,zhu2023hifa} introduce a coarse-to-fine optimization strategy with two stages and improve both speed and quality. Recent studies \cite{ren2023dreamgaussian4d,ben2024dreamgaussian,yi2024gaussiandreamer,jaganathan2024ice, yi2024gaussiandreamerpro} incorporate 3DGS into generative 3D content creation, achieving acceleration compared to NeRF-based approaches. There also emerge hybrid 3D generative methods \cite{li2024instant3d,shi2024mvdream,chen2024sculpt3d,yang2024viewfusion,seo2024retrievalaugmented,ye2023consistent1to3,liu2024vqa, dong2024interactive3d, tang2024lgm, yan2025consistentflowdistillationtextto3d, liu2024sherpa3d, yan2024dreamview, chen2024vp3d} by combining the advantage of 3D native and 2D prior-based generative methods. The typical example is One-2-3-45++ \cite{one2345++}, which generates 3D representations by training a 3D diffusion model with the input of 2D prior-based multi-view images. 

%Although promising advances have been reported with existing methods, the quality of text-to-3D generation is still far from satisfactory. For example, due to Stable Diffusion’s bias toward 2D front views, its 3D outputs tended to repeat front views from different angles rather than generating coherent 3D objects. These problems influence the visual perception from different quality dimensions. To provide meaningful guidance for model design choices and support fair comparisons, it is necessary to explore human-centric quality evaluation for 3D generation.

\subsection{Text-to-3D Evaluation}
To facilitate the text-to-3D evaluation, several benchmarks have been developed. $\mathrm{T}^3$Bench \cite{he2023t3bench} devises three prompt suites incorporating diverse 3D scenes and with increasing complexity, including \textit{Single object}, \textit{Single object with surroundings}, and \textit{Multiple objects}, and then creates 630 samples scored from two dimensions, \textit{i.e.,} quality and alignment. Another benchmark \cite{wu2024gpt4v} leverages GPT-4 to generate prompts with varying degrees of creativity and complexity, then evaluates preferences between 234 sample pairs generated from identical prompts across five dimensions.  Overall, these benchmarks face limitations, such as insufficient prompt diversity and the absence of multi-dimensional absolute scores, emphasizing the need for a more comprehensive benchmark. Furthermore, existing evaluation metrics \cite{hessel2021clipscore,li2022blip,schuhmann2022laion,xu2024imagereward,wu2023human, li2024evaluating, ye2024dreamreward} are either limited to single-dimensional evaluations or dependent on simple pairwise comparisons, which are insufficient to face the nuanced challenges of text-to-3D generation. Consequently, there is an urgent need for a comprehensive metric capable of assessing multi-dimensional quality.

There exists some work targeted for multi-dimensional evaluation in text-to-image generation. An intriguing way, MPS \cite{zhang2024learning}, introduces the condition features upon CLIP model to learn diverse dimensions. We inherit the merits of MPS and further introduce two effective strategies to achieve a fine-grained quality evaluation across dimensions.

%To facilitate text-to-3D evaluation, several benchmarks have been developed. $\mathrm{T}^3$Bench devises three prompt suites incorporating diverse 3D scenes and with increasing complexity, including \textit{Single object}, \textit{Single object with surroundings}, and \textit{Multiple objects}, followed by collecting 1260 scores from two dimensions, \textit{i.e.,} quality and alignment. Another benchmark \cite{} leverages GPT-4 to generate prompts with varying degrees of creativity and complexity, then evaluates preferences between sample pairs generated from identical prompts across five dimensions.  Overall, these benchmarks face limitations, such as insufficient prompt diversity and the absence of multi-dimensional absolute scoring, underscoring the need for a more comprehensive benchmark. 

%As for text-to-3D evaluation metrics, the existing metrics mainly rely on CLIP and BLIP to measure the similarity between text and rendered images of generated 3D representations. \cite{} combines multi-view text-image scores and regional convolution to measure quality and uses multi-view captioning and GPT-4 evaluation to measure text-3D consistency. \cite{} instructs GPT-4V to compare two 3D samples based on user-defined criteria. In general, these metrics are either limited to single-dimensional evaluations or rely on simple pairwise comparisons, which are insufficient to face the nuanced challenges of text-to-3D generation. Consequently, there is an urgent need for a comprehensive metric capable of assessing multi-dimensional quality in this rapidly evolving field.

\section{Benchmark Construction and Analysis}\label{sec:database}

\subsection{Data Preparation}

\begin{figure}[t]
    \centering
    \begin{subfigure}[t]{0.48\textwidth}
        \centering
        \includegraphics[width=\linewidth]{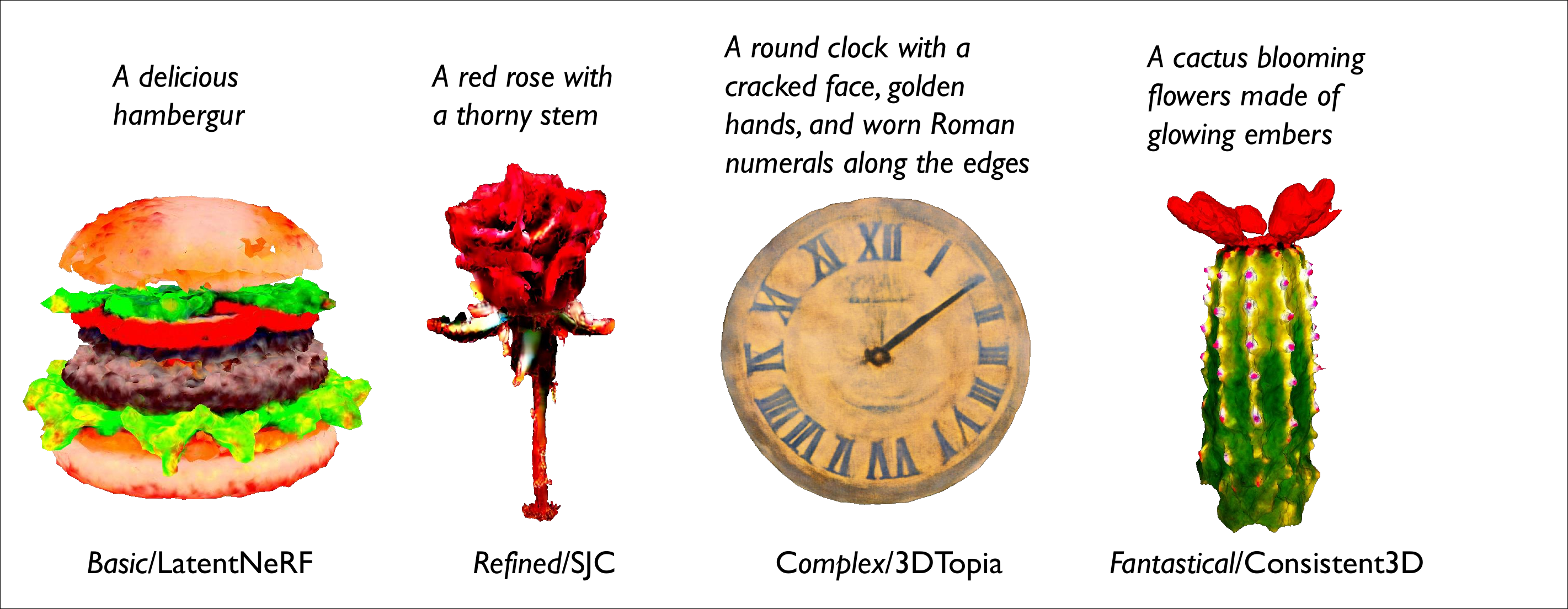}
        \label{fig:sample1}
    \end{subfigure}  
    \vspace{0.1cm}
    \begin{subfigure}[t]{0.48\textwidth}
        \centering
        \includegraphics[width=\linewidth]{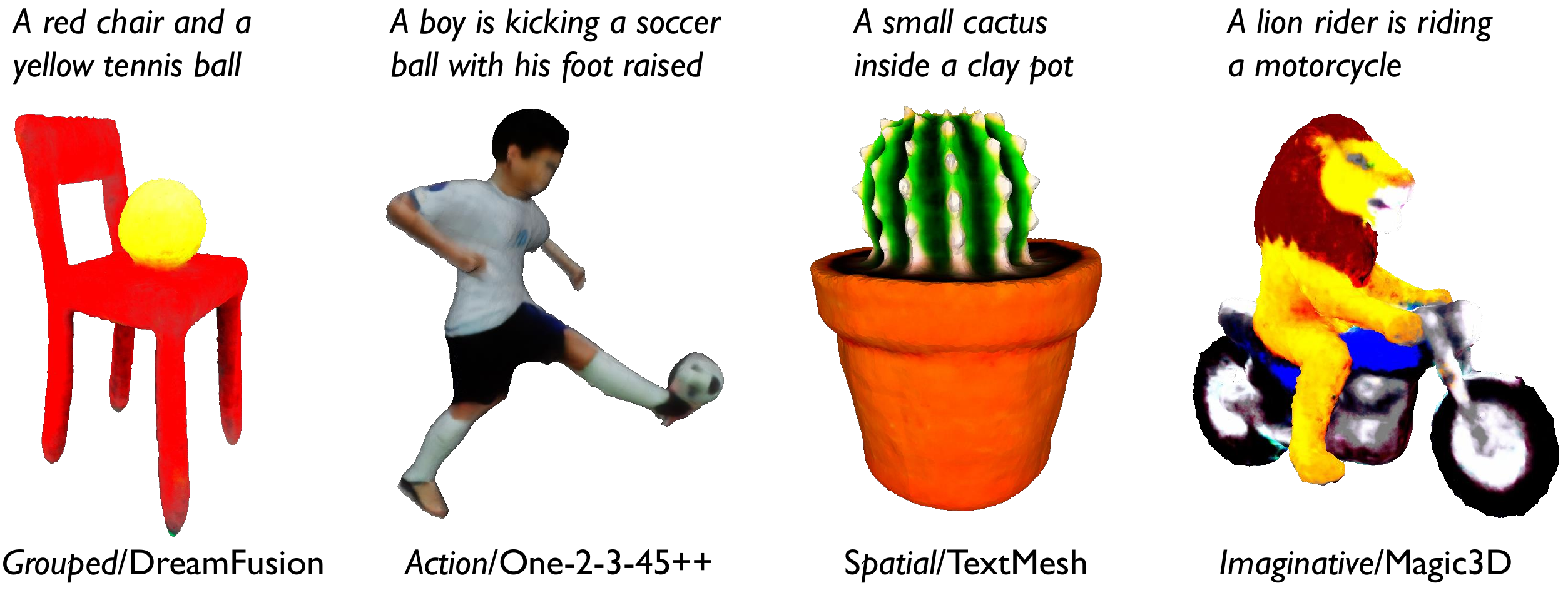}
        \label{fig:sample2}
    \end{subfigure}
   \caption{Samples from the database generated by eight different generative methods.}
   \label{fig:samples}
   \vspace{-0.5cm}
\end{figure}

\textbf{Prompt Categorization.} 
To enhance the comprehensiveness and diversity of constructed prompts, we thoroughly consider the variety of scenarios present in the real world. We first classify text prompts based on object quantity. For single object generation, we define four sub-categories including \textit{``Basic''}, \textit{``Refined''}, \textit{``Complex''} and \textit{``Fantastical''}, where the former three categories have increasing textual complexity while the last category focuses on generating object with high creativity.
For multiple object generation, the text prompt is composed of at least two objects with specified relationships between the objects. Based on the type of the relationships, we define four sub-categories denoted by \textit{``Grouped''}, \textit{``Action''}, \textit{``Spatial''}, and \textit{``Imaginative''}. The \textit{Grouped} category describes multiple objects connected by ``and''; The \textit{Action} and \textit{Spatial} categories describe action interactions (\textit{e.g.}, ``hold'', ``watch'', ``play with'') and spatial (\textit{e.g.} ``on'', ``near'', ``on the bottom of'' ), respectively; \textit{``Imaginative''} describe multiple objects with interactions, where objects or interactions should be creative. \cref{fig:samples} shows eight exemplary prompts corresponding to different categories. We detail the definition of each category in the appendix.

\textbf{Prompt Generation.} Based on the defined categories, we generate well-structured prompts for each category with the assistance of GPT-4 \cite{openaigpt} as shown in \cref{fig:prompts}. First, we follow \cite{wu2024gpt4v} to provide an instruction to ensure that GPT-4 understands the task requirements. To promote prompt diversity, the instruction first emphasizes four aspects to consider: object categories, geometry properties, appearance properties, and object interactions. Then, the instruction defines the above eight categories with examples to ensure that the prompts generated by GPT-4 closely align with each category. Based on the instruction, we utilize GPT-4 to create a large pool of candidate prompts tailored to the identified challenges. Finally, we manually filter out any duplicate or inappropriate prompts to ensure consistency and diversity within the benchmark. Overall, we select 20 prompts from a pool of 100 candidates for each category, resulting in a total of 160 prompts. 

\textbf{Mesh Generation.} After the prompt generation, we adopt eight recent text-to-3D generative methods, including DreamFusion \cite{dreamfusion}, Magic3D \cite{magic3d}, Score Jacobian Chaining (SJC) \cite{sjc}, TextMesh \cite{textmesh}, 3DTopia \cite{3dtopia}, Consistent3D \cite{consistent3d}, LatentNeRF \cite{latentnerf}, and One-2-3-45++ \cite{one2345++} to generate 3D representations by using open source code and default weight, which are then transformed into textured meshes. Finally, we obtain 1,280 textured meshes. \cref{fig:samples} shows the mesh samples in the database along with their corresponding prompts and generative methods.

\subsection{Subjective Study}

% \textbf{Evaluation Dimension.} To evaluate the quality of the  textured meshes in the database and obtain Mean Opinion Scores (MOS), a subjective experiment is conducted following the guidelines of ITU-R BT.500-14. Due to the unnatural property of AIG-3D textured mesh and different text prompts having different target mesh spaces, it is unreasonable to use one score, i.e., ``quality" to represent human visual preferences. For natural textured mesh, researchers typically evaluate its quality from the perspectives of ``geometry" and ``texture". These two perspectives are also applicable to AIG-3D textured mesh. In this paper, we propose four perspectives including alignment, geometry, texture, and overall to comprehensively assess the quality of AIG-3D samples. By employing these four perspectives, we aim to provide a more nuanced understanding of human preferences. These four perspectives are defined as follows:

\textbf{Evaluation Dimension.}  The quality of the generated textured meshes is influenced by different factors. To provide a more nuanced understanding of human perception, we define four evaluation dimensions that need annotation:
\begin{itemize}
    \item \textbf{Alignment.} Subjects should measure the semantic consistency of the generated meshes with the prompts, including evaluating whether the generated meshes accurately match the prompts (\textit{e.g}., quantity, attributes, location, relationship) and whether there is missing or redundant content.
    \item \textbf{Geometry.} Subjects should measure the geometry fidelity of textured meshes, including measuring the shape, contour, size, and whether there is missing or redundant structure in the generated meshes.
    \item \textbf{Texture.} Subjects should measure the texture details of textured meshes, including evaluating the color, material, and whether the appearance of meshes is vibrant and of high resolution.
    \item \textbf{Overall.} Subjects should conduct a holistic assessment of the samples by integrating the insights from different perspectives.  
\end{itemize}

\begin{figure}[t]
  \centering
   \includegraphics[width=0.9\linewidth]{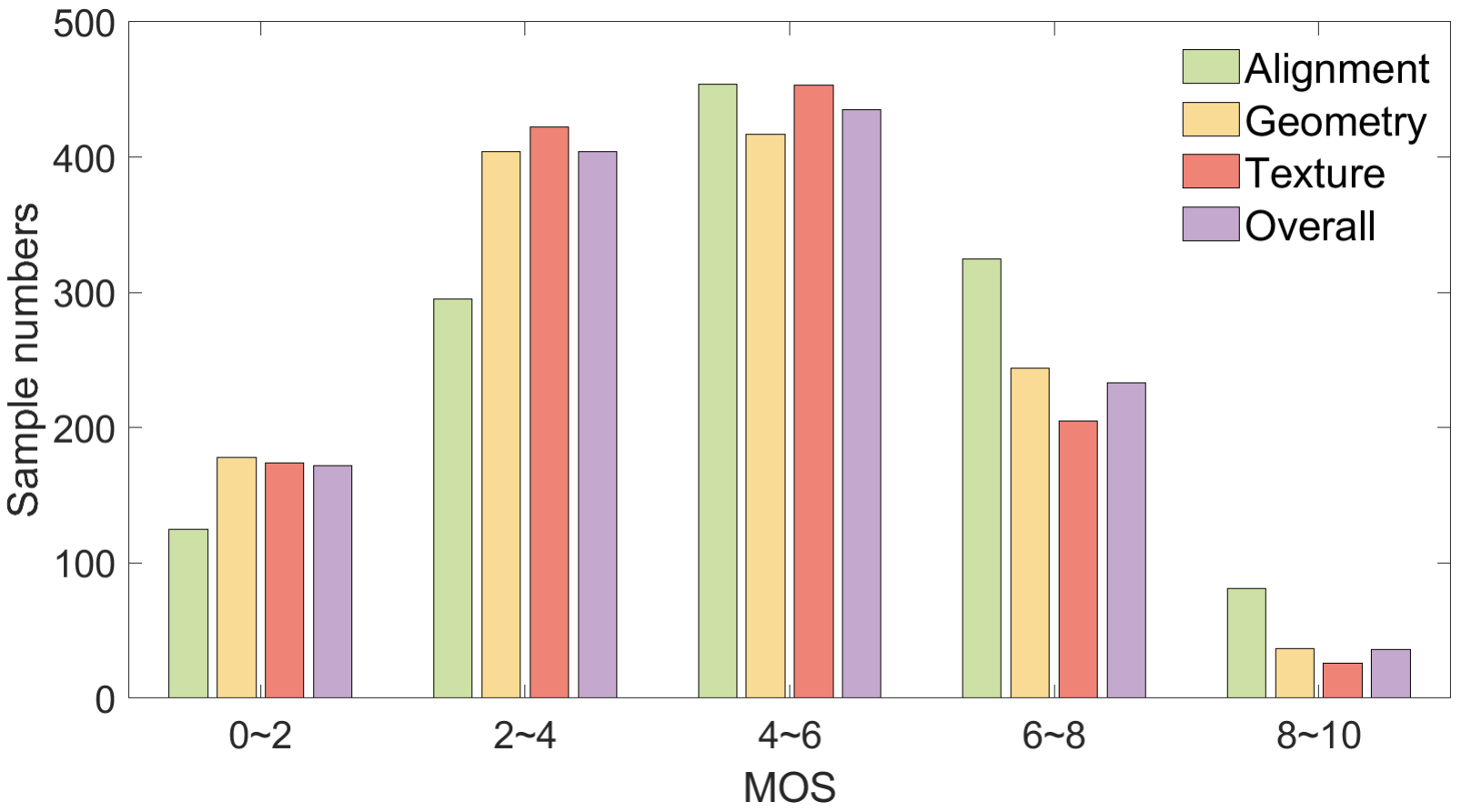}
   \caption{Distribution of MOS for four dimensions.}
   \label{fig:mos}
\vspace{-0.3cm}
\end{figure}

\textbf{Experiment Setup.} To obtain the MOS of each mesh, we invite 21 subjects to score the samples from the above four dimensions. We employ the 11-level impairment scale (ranging from 0 to 10) proposed by ITU-T P.910 \cite{p910} as the voting methodology.  After collecting the subjective scores, we implement the outlier detection method outlined in ITU-R BT.500 \cite{bt500} to identify and exclude outliers from the raw data. Finally, the MOS for each dimension of one textured mesh is calculated by averaging the filtered raw scores.
%as follows:
% \begin{equation}
%     MOS_{j}^{i} =\frac{1}{L}\sum_{l=1}^{L}r_{j,l}^{i}
% \end{equation}
% where $r_{j,l}^{i}$ indicates the score given by the $l$-th subject for the $j$-th sample from the $i$-th quality dimension; $L$ denotes the number of valid subjects. 
\cref{fig:mos} shows the distribution of MOS for alignment, geometry, texture, and overall quality. We see that the benchmark covers broad quality ranges for multiple dimensions.

\subsection{Observations}

\begin{figure}[t]
    \centering
    \begin{subfigure}[t]{0.22\textwidth}
        \centering
        \includegraphics[width=\linewidth]{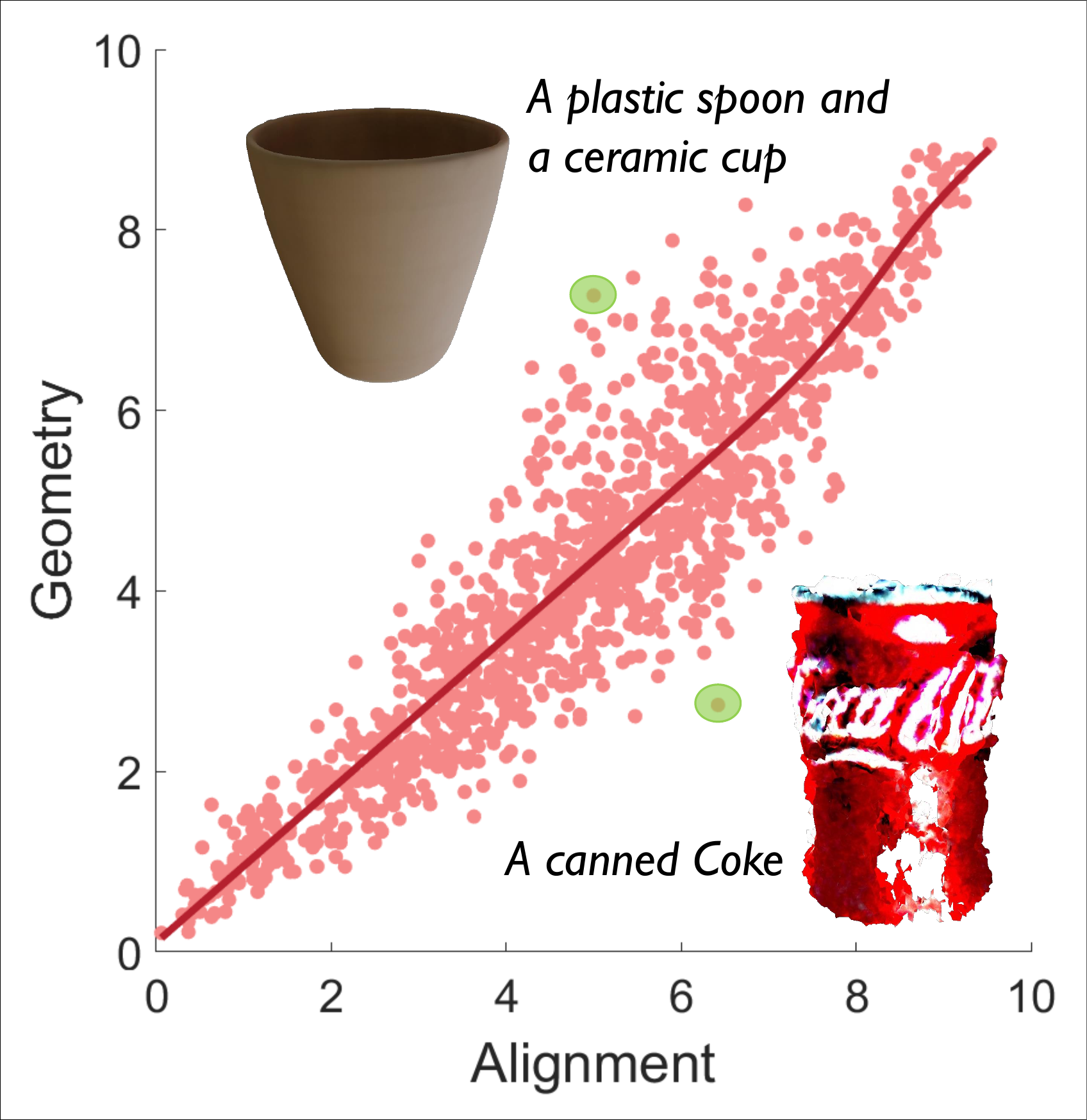}
        \caption{}
        \label{fig:alignment_geometry}
    \end{subfigure}  
    \hspace{0.005\textwidth} % 控制子图之间的水平间距
    \begin{subfigure}[t]{0.22\textwidth}
        \centering
        \includegraphics[width=\linewidth]{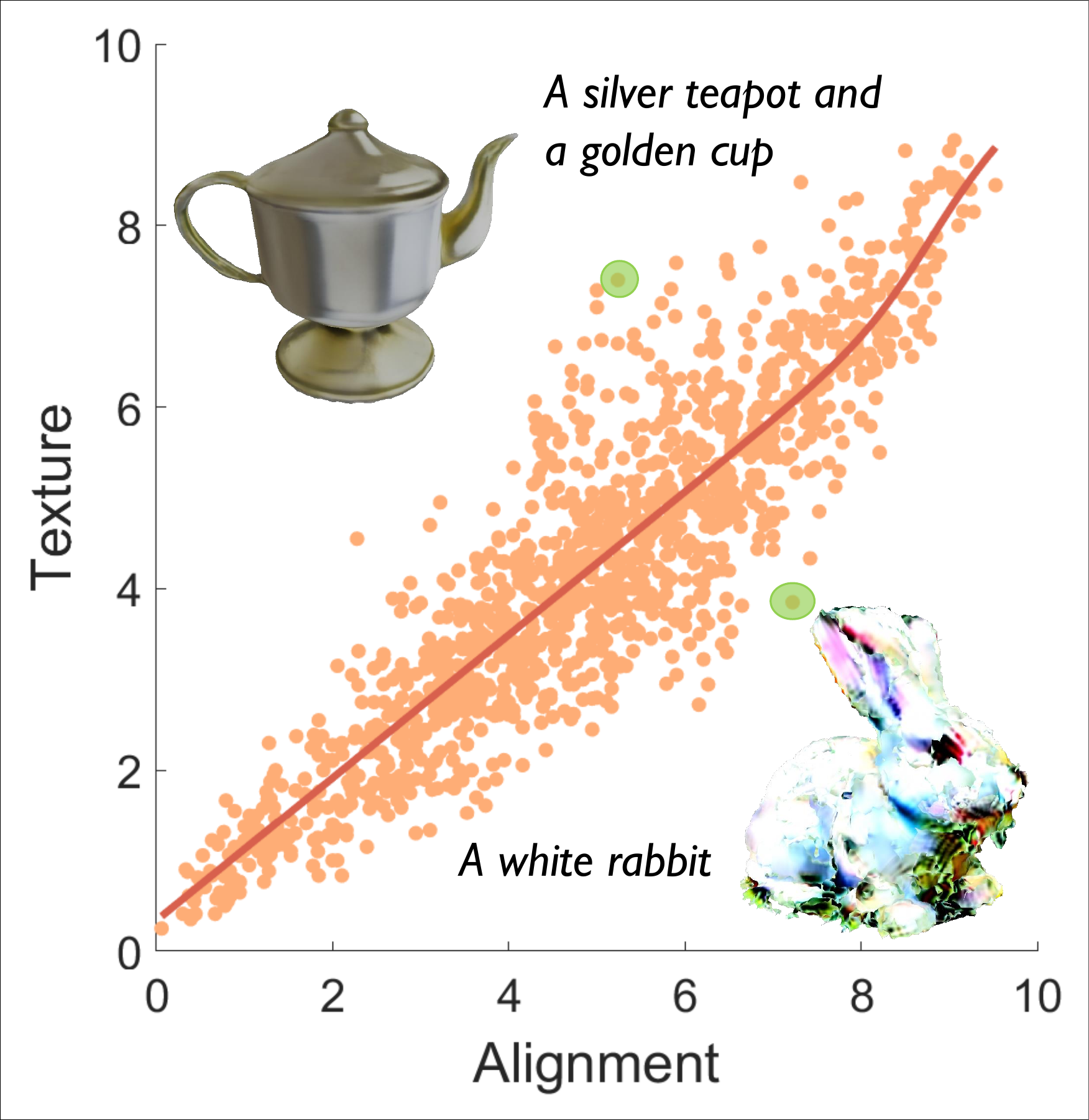}
        \caption{}
        \label{fig:alignment_texture}
    \end{subfigure}
    \vspace{0.001\textwidth}
    \begin{subfigure}[t]{0.22\textwidth}
        \centering
        \includegraphics[width=\linewidth]{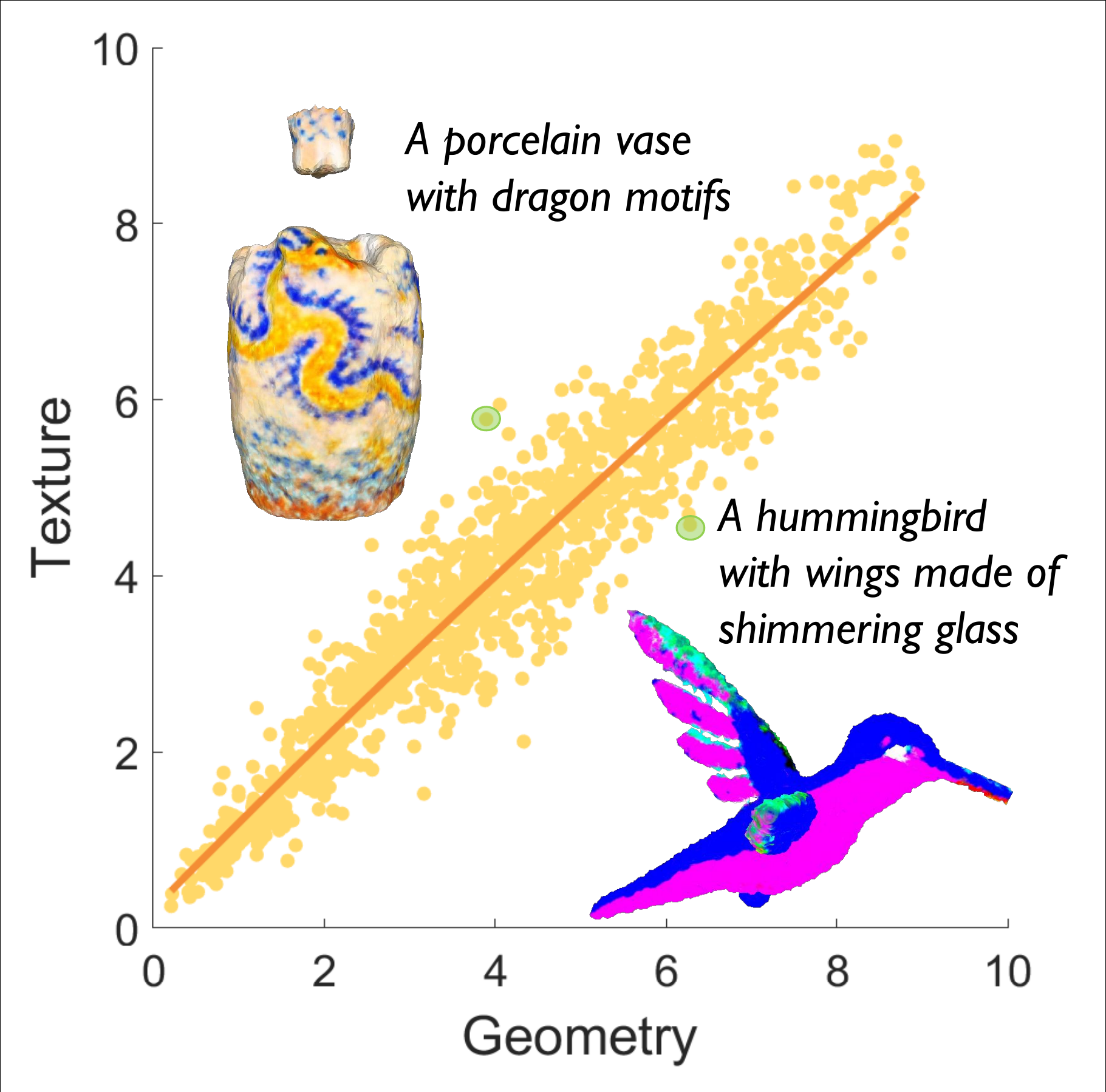}
        \caption{}
        \label{fig:geometry_texture}
    \end{subfigure}  
    \hspace{0.005\textwidth} % 控制子图之间的水平间距
    \begin{subfigure}[t]{0.24\textwidth}
        \centering
        \includegraphics[width=\linewidth]{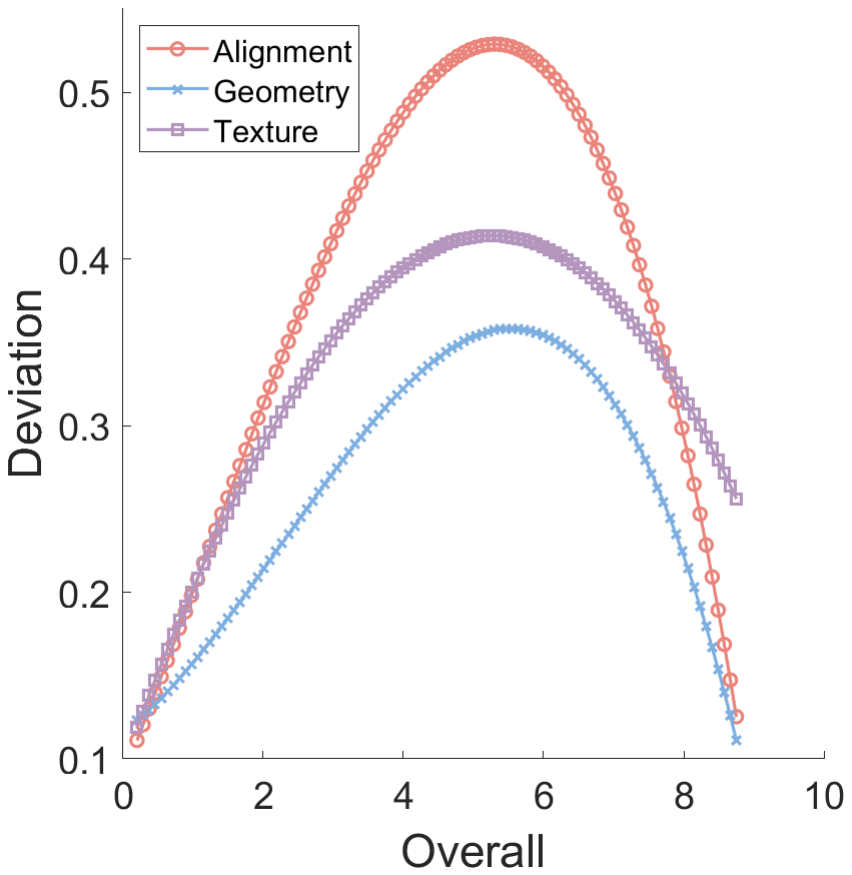}
        \caption{}
        \label{fig:deviation_overall}
    \end{subfigure}
    
    \caption{(a) Alignment \textit{vs.} Geometry; (b) Alignment \textit{vs.} Texture; (c) Geometry \textit{vs.} Texture; (d) Deviations among the other three dimensions and overall quality.}
    \label{fig:deviations}
    \vspace{-0.3cm}
\end{figure}

We conduct comprehensive analyses for the subjective scores of MATE-3D.  We first investigate the relationships among different evaluation dimensions. \cref{fig:alignment_geometry} - \cref{fig:geometry_texture} illustrate scatter plots that depict the correlations among three dimensions (\textit{i.e.}, alignment, geometry, and texture). From \cref{fig:alignment_geometry} and \cref{fig:alignment_texture},
we observe a relatively low correlation between the alignment and the geometry or texture score at the medium quality range (\textit{e.g.,} $[4,\ 6]$). The reason is that the alignment evaluation mostly focuses on semantic consistency while the geometry or texture evaluation relies on specified details. For instance, in \cref{fig:alignment_geometry}, the mesh generated from the prompt ``A plastic spoon and a ceramic cup" includes only a cup, resulting in a low alignment score. However, the cup itself demonstrates good geometry quality. In contrast, the sample corresponding to the prompt ``A canned Coke" achieves a relatively high alignment score but suffers from an uneven surface and an incomplete shape, leading to a low geometry score. From \cref{fig:geometry_texture}, the mutual influence between the geometry and texture scores is more pronounced, as the two dimensions influence each other during the generation process. Finally, in \cref{fig:deviation_overall}, we calculate the deviations between the other three dimensions and the overall score after non-linear regression, which indicates the consistency between each dimension and the overall quality. We can see that the geometry quality is most closely related to the overall quality. It is reasonable because geometry deformations (\textit{e.g.}, incomplete shape, floaters) directly affect 3D perception.
The alignment quality appears to have the least correlation with the overall quality because it relatively ignores some details that human subjects care about.

\begin{table*}[t]
\centering
\caption{Average scores of each generative method on eight prompt categories. The values of 1-8 represent the rank in each row. }
\label{tab:challenge_performance}
\resizebox{0.8\textwidth}{!}{
\begin{tabular}{c|cccc|cccc}
\toprule
\multirow{2}{*}{Method} & \multicolumn{4}{c|}{Single Object}  & \multicolumn{4}{c}{Multiple Object} \\ \cline{2-9}
& Basic & Refined & Complex & Fantastic & Grouped & Action & Spatial & Imaginative 
 \\ \midrule

DreamFusion &4.99(1) &4.96(2) &4.64(4) &3.24(8) &3.45(7) &3.73(6) &4.47(3) &3.75(5) \\

Magic3D &5.26(1) &4.89(6) &5.06(3) &4.95(4) &3.92(8) &4.73(7) &5.06(2) &4.93(5) \\

SJC &3.28(2) &3.48(1) &2.95(4) &2.92(6) &2.50(8) &2.77(7) &2.92(5) &3.25(3) \\

TextMesh &4.08(6) &4.56(2) &4.79(1) &4.22(4) &3.22(8) &3.68(7) &4.17(5) &4.26(3) \\

3DTopia &5.06(1) &4.91(2) &4.83(3) &3.89(6) &4.05(4) &2.80(8) &3.89(5) &2.95(7) \\

Consistent3D &4.15(5) &4.92(1) &4.40(2) &4.32(4) &3.37(8) &3.64(6) &4.39(3) &3.47(7) \\

LatentNeRF &3.20(7) &3.76(3) &3.47(5) &4.16(1) &2.94(8) &3.48(4) &3.46(6) &4.01(2) \\

One-2-3-45++ &7.79(1) &7.69(2) &6.50(5) &6.60(4) &6.49(6) &5.65(8) &6.81(3) &6.13(7) \\

\bottomrule
\end{tabular}}
\end{table*}

\begin{figure}[t]
    \centering
    \includegraphics[width=0.75\linewidth]{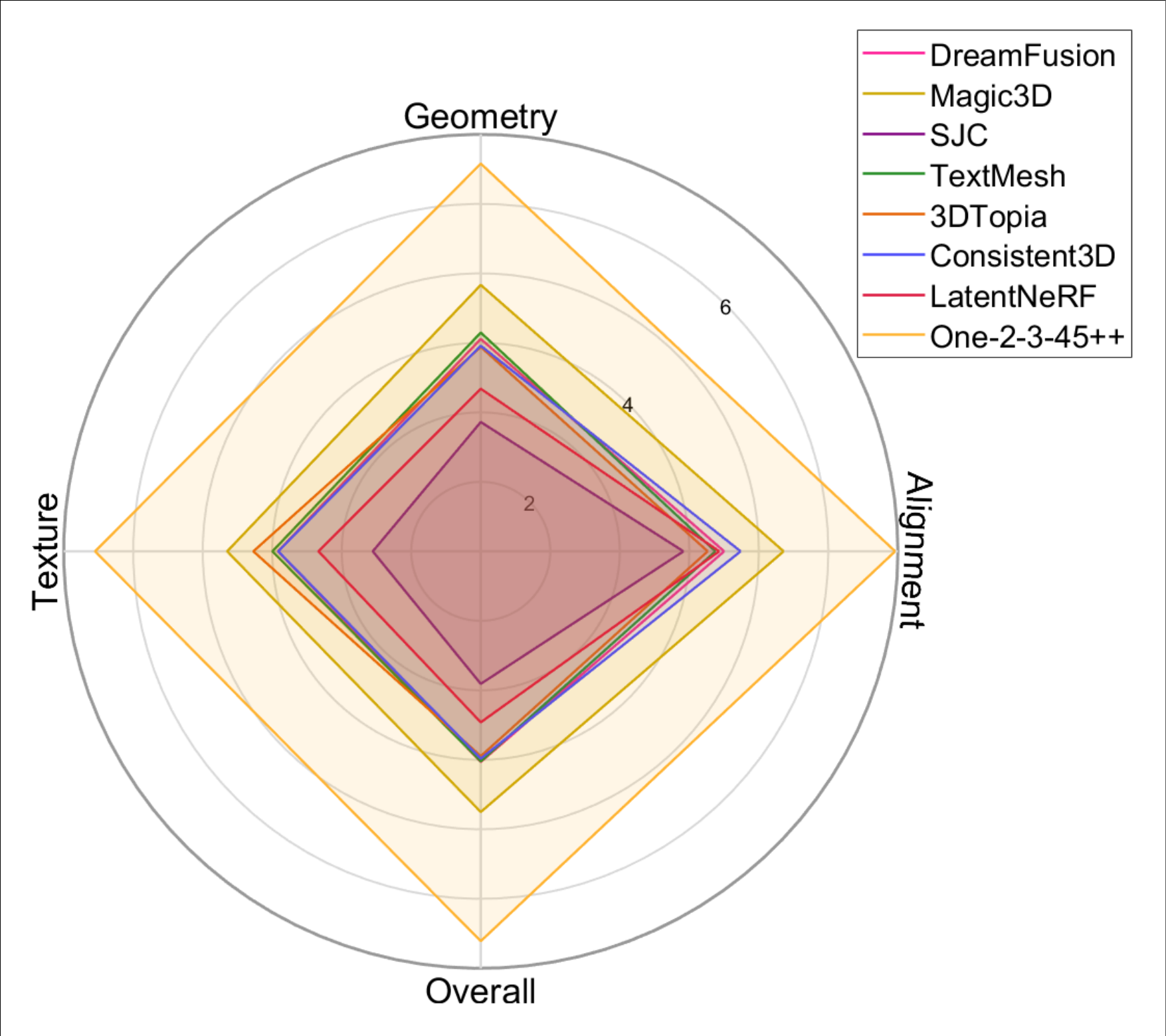}
   \caption{Average scores of different methods on four dimensions.}
   \label{fig:model}
   \vspace{-0.5cm}
\end{figure}

We further illustrate the average scores of each generative method on the four evaluation dimensions in \cref{fig:model}. We can see that One-2-3-45++ performs best on all dimensions, while SJC provides the worst performance. This is primarily because SJC usually causes an incomplete geometry shape and noisy floaters, making it difficult to generate high-quality samples. Other generative methods also face some difficulties during the generation process. For example, TextMesh often generates 3D objects with overly simplistic shapes, which limits their complexity and realism. 3DTopia tends to generate single object, leading to comparatively lower scores in multiple object generation. One-2-3-45++ tends to generate flatter geometric structures and blurry textures. Moreover, except for One-2-3-45++, all generative methods can cause Janus problem due to the bias of the used 2D diffusion models towards 2D front views. The bias results in repeating front views from different angles rather than generating coherent 3D objects.

To investigate the ability of the methods to generate different scenes, we report the average scores (in terms of overall quality) of each method on eight prompt categories in \cref{tab:challenge_performance}. We have the following findings: i) All methods achieve better performance for single object generation than multiple object generation. In fact, existing methods struggle in constructing the correct relationships among objects and usually miss some contents during generation, restricting their performance in multiple object generation. ii) For the sub-categories of single object generation, the generation quality of most methods decreases with increasing prompt complexity (from \textit{Basic} to \textit{Refined} to \textit{Complex}). However, there are some exceptions. For example, the generated samples of TextMesh often exhibit too simple geometric structures. Consequently, its generation for the \textit{Basic} category may be too simplistic to discern, leading to lower scores. iii) For the sub-categories of multiple object generation, it is observed that most methods provide better performance on the \textit{Spatial} category, followed by the \textit{Action} and \textit{Grouped} categories.
It seems that these methods are more capable of modeling specific spatial interactions. In contrast, the \textit{Action} prompts may introduce ambiguity with actions like ``chase" or ``run" because these verbs often indicate a greater distance, while the \textit{Grouped} category usually results in overlapping or missing objects. iv) The \textit{Fantastic} and \textit{Imaginative} categories primarily emphasize the creativity of the prompts, allowing a wide range of object details and interactions. As a result, different methods perform inconsistently on the two categories.

\begin{figure}
    \centering
    \includegraphics[width=\linewidth]{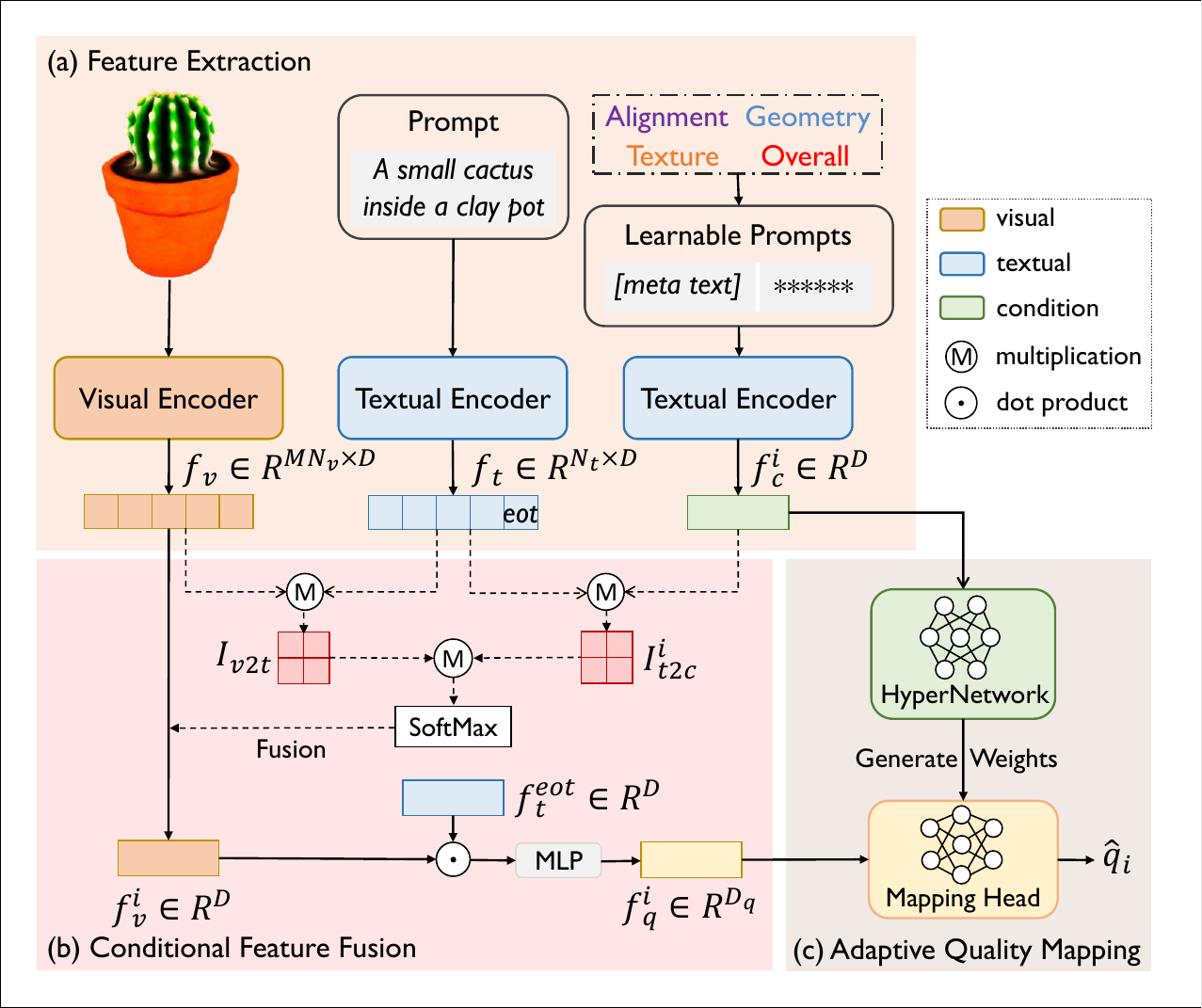}
    \caption{The framework of the proposed evaluator.}
    \label{fig:framework}
    \vspace{-0.5cm}
\end{figure}

\section{Evaluator for Text-to-3D Generation}

\subsection{Problem Formulation}
Given a generated mesh $x$ and its corresponding prompt $t$, existing single-dimensional evaluators directly map them to the subjective score $q$. The procedure can be described as $\hat{q}=\psi\left ( \phi(x,t) \right ) $, where $\hat{q}$ is the predicted score; $\phi\left ( \cdot \right )$ denotes a feature extractor such as CLIP and $\psi\left ( \cdot \right )$ represents a mapping function such as the cosine similarity function in CLIPScore \cite{hessel2021clipscore}. Given multiple targets $\left \{ q_i \right \}_{i=1}^{K}$ from $K$ dimensions, a direct approach is to separately use a single-head network for each dimension. However, it is not economical to train and retain dozens of different expert models.
Therefore, we prefer a one-for-all evaluator. For the purpose, a solution is a multi-task network with a shared feature extractor and multiple mapping heads, that is,
\begin{equation}
    \hat{q}_i =\psi_i\left ( \phi\left ( x,t \right )|\theta_i \right ),
\end{equation}
where $\psi_i\left (\cdot|\theta_i \right )$ denotes the mapping head for the $i$-th dimension with the parameters $\theta_i$. However, directly optimizing such a multi-task network may fail to exploit differentiated perception rules depending on the evaluation dimensions. Therefore, we further transform the task as follows:
\begin{equation}
    \hat{q}_i=\psi\left ( \phi\left ( x,t \right )|\pi\left ( f_c^i \right ) \right ),
\end{equation}
where $\pi\left ( \cdot \right )$ denotes a hypernetwork to generate $\theta_i$ for a unified mapping head; $f_c^i$ denotes a condition feature related to the $i$-th evaluation dimension. By injecting the condition features, we can better characterize the differences among evaluation perspectives.

% The framework of our evaluator can divided into three stages: i) Feature Extraction; ii) Conditional Feature Fusion; iii) Adaptive Quality Mapping. We will detail these stages in the following.

We illustrate the framework of our evaluator in \cref{fig:framework}, which can be divided into three stages: i) \textbf{Feature Extraction}. We first use a transformer-based CLIP model to extract visual and textual features from rendered images of textured meshes and prompts. Meanwhile, we feed a sequence of learnable prompts into the textual encoder to obtain multiple condition features corresponding to various dimensions. ii) \textbf{Conditional Feature Fusion}. We utilize the condition features to compute fusion weights for various patches in the visual features. The fused visual feature is then merged with the textual feature to obtain the final quality feature. iii) \textbf{Adaptive Quality Mapping}. We feed the condition features into a hypernetwork to generate diverse weights of the mapping head, and then regress the quality feature into the final score of the specified dimension.

\subsection{Feature Extraction}
For a test mesh $x$, we first render it into multi-view images as $\left \{ x_{r}^{m} \right \}_{m=1}^{M}$, where $M$ denotes the number of viewpoints. Then we resort to a pre-trained CLIP model for feature extraction, which is composed of a transformer-based visual encoder $\phi_v\left ( \cdot \right )$ and a frozen textual encoder $\phi_t\left ( \cdot \right )$. To better capture local distortion patterns, we use the tokens of all image patches as the output of $\phi_v\left ( \cdot \right )$. The feature extraction process can be formulated as:
\begin{equation}
    f_v^m =\phi_v\left ( x_r^m \right ) \in\R^{N_v\times D}, \ 
    f_t =\phi_t\left ( t\right )\in\R^{N_t\times D},
\end{equation}
where $N_v$ denotes the token number of image (\textit{i.e.,} the patch number) and  $N_t$ denotes the token number of text. We further concatenate the visual features of different viewpoints to derive $f_v\in\R^{MN_v\times D}$.

Previous research \cite{hou2023towards,zhang2024learning} demonstrates that specific tags (\textit{e.g.,} ``shape, edge" for geometry evaluation) can effectively capture dimension-related information. To avoid meticulous design for fixed tags, we only define $K$ meta text about different evaluation dimensions (\textit{e.g.}, \textit{``alignment quality'', ``geometry quality'', ``texture quality'', ``overall quality''} for MATE-3D) and acquire the corresponding meta tokens. We further introduce the learnable prompts following CoOp \cite{zhou2022learning} and insert the meta tokens in the front of a squeeze of learnable tokens to acquire $K$ tokenized prompts as $\left \{ t_c^i \right \}_{i=1}^{K}$. Subsequently, these tokenized prompts will pass through the frozen textual encoder to obtain a list of condition features denoted by $\left \{ f_c^i\in\R^D \right \}_{i=1}^K$, where we only retain the EOT (end of text) token as the output because it is often used to represent the whole text.

\begin{table*}[t]
\centering
\caption{Performance Comparison on MATE-3D.  The best metric in each column is marked in \textbf{boldface}.}
\label{tab:overall_performance}
\resizebox{0.9\textwidth}{!}{
\begin{tabular}{c|ccc|ccc|ccc|ccc}
\toprule
\multirow{2}{*}{Metric} & \multicolumn{3}{c|}{Alignment}  & \multicolumn{3}{c|}{Geometry} & \multicolumn{3}{c|}{Texture} & \multicolumn{3}{c}{Overall}  \\ \cline{2-13}
& PLCC & SRCC & KRCC & PLCC & SRCC & KRCC & PLCC & SRCC & KRCC  & PLCC & SRCC & KRCC \\ \midrule

CLIPScore \cite{hessel2021clipscore} &0.535	&0.494	&0.347	&0.520	&0.496	&0.347	&0.563	&0.537	&0.379	&0.541	&0.510	&0.359

\\
BLIPScore \cite{li2022blip}  &0.569	&0.533	&0.377	&0.570	&0.542	&0.382	&0.608	&0.578	&0.413	&0.586	&0.554	&0.393
\\
Aesthetic Score \cite{schuhmann2022laion} &0.185	&0.099	&0.066	&0.270	&0.160	&0.107	&0.288	&0.172	&0.114	&0.228	&0.150	&0.100
\\

ImageReward \cite{xu2024imagereward} &{0.675}	&{0.651}	&0.470	&{0.624}	&{0.591}	&{0.422}	&{0.650}	&{0.612}	&{0.441}	&{0.657}	&{0.623}	&{0.448}
\\

DreamReward \cite{ye2024dreamreward} &0.533	&0.526	&0.368	&0.528	&0.501	&0.349	&0.544	&0.513	&0.359	&0.551	&0.524	&0.367 

\\
HPS v2 \cite{wu2023human} &0.477	&0.416	&0.290	&0.455	&0.412	&0.286	&0.469	&0.423	&0.296	&0.465	&0.420	&0.293
\\
CLIP-IQA \cite{wang2023exploring} 	&0.151	&0.039	&0.025	&0.223	&0.125	&0.084	&0.264	&0.163	&0.108	&0.213	&0.115	&0.076
\\
Q-Align \cite{wu2023q} &0.223	&0.199	&0.135	&0.389	&0.355	&0.300	&0.371	&0.431	&0.300	&0.371	&0.340	&0.234

\\ \midrule
 
ResNet50 + FT \cite{he2016deep} &0.577	&0.551	&0.387	&0.669	&0.653	&0.471	&0.691	&0.672	&0.487	&0.649	&0.632	&0.452
\\
ViT-B + FT \cite{dosovitskiy2020image} &0.544	&0.515	&0.360	&0.662	&0.642	&0.461	&0.691	&0.676	&0.492	&0.633	&0.614	&0.439
\\
SwinT-B + FT \cite{liu2021swin} &0.534	&0.506	&0.359	&0.626	&0.610	&0.438	&0.654	&0.640	&0.465	&0.608	&0.592	&0.425 \\
DINO v2 + FT \cite{oquab2023dinov2} &0.664	&0.642	&0.461	&0.761	&0.739	&0.550	&0.787	&0.771	&0.579	&0.746	&0.728	&0.538

\\
MultiScore &0.657	&0.638	&0.458	&0.719	&0.703	&0.516	&0.746	&0.729	&0.540	&0.714	&0.698	&0.511 \\

\textbf{HyperScore}   &\bf{0.754}	&\bf{0.739}	&\bf{0.547}	&\bf{0.793}	&\bf{0.782}	&\bf{0.588}	&\bf{0.822}	&\bf{0.811}	&\bf{0.619}	&\bf{0.804}	&\bf{0.792}	&\bf{0.600}

\\ \bottomrule
\end{tabular}}
\vspace{-0.3cm}
\end{table*}

\subsection{Conditional Feature Fusion}

As aforementioned, human evaluators adaptively focus on different aspects of scenes depending on the evaluation dimensions. Therefore, different patches in multiple viewpoints do not contribute equally to the final prediction; meanwhile, the contribution of each patch may change across dimensions. Based on the insight, we implement a conditional feature fusion strategy.

Specifically, we first normalize the visual, textual and condition features to obtain $\Tilde{f_v}$, $\Tilde{f}_t$, and $\left \{ \Tilde{f}_c^i \right \}_{i=1}^{K}$. For each evaluation dimension, we define two correlation matrices as: 
\begin{equation}
    \begin{aligned} I_{v2t}=\Tilde{f}_v\cdot\Tilde{f}_t^T\in\R^{MN_v\times N_t},\ I_{t2c}^i=\Tilde{f}_t\cdot\left(\Tilde{f}_c^i \right)^T\in\R^{N_t},
    \end{aligned}
\end{equation}
where $I_{v2t}$ quantifies the degree of correlation between each image patch and each text token, whereas $I_{t2c}^i$ determines the relative importance of each text token for the $i$-th condition. Then, we perform the conditional feature fusion for the $i$-th evaluation dimension as:
\begin{equation}\label{eq:fusion_weight}
    f_{v,c}^i = \mathrm{SoftMax}\left ( I_{v2t}\cdot I_{t2c}^i \right )\cdot f_v \in\R^{D}, 
\end{equation}
where $\mathrm{SoftMax}\left ( \cdot \right )$ denotes the softmax function. Based on Eq. \eqref{eq:fusion_weight}, the more relevant the patch is to the condition, the higher the weight. Finally, we derive the quality feature of the $i$-th dimension as: 
\begin{equation}\label{eq:feature_aggregate}
    f_q^i = \mathrm{MLP}\left ( f_{v,c}^i\odot f_t^{eot} \right )\in\R^{D_q},
\end{equation}
where $\odot$ denotes the element-wise multiplication and $\mathrm{MLP}\left ( \cdot \right )$ represents a simple multi-layer perception; $f_t^{eot}\in\R^D$ indicates the EOT token in $f_t$.

\subsection{Adaptive Quality Mapping}
After obtaining the quality feature for the specified condition, we need to transform it into the corresponding quality score using a mapping function $\psi\left ( \cdot \right )$ parameterized by the weights $\theta_i$. Previous studies \cite{su2020blindly} have indicated that the mapping function can represent the decision procedure used to assess quality. We further argue that as the evaluation dimension considered varies, the decision procedure also varies. Therefore, we define $\psi\left ( \cdot \right )$ as fully connected layers and use a hypernetwork $\pi\left ( \cdot \right )$ to generate $\theta_i$, \textit{i.e.}, $\theta_i = \pi\left ( f_c^i \right )$.
Two types of network parameters, \textit{i.e.}, fully connected layer weights and biases, are generated (please refer to the appendix for the detailed architecture of $\psi\left ( \cdot \right )$ and $\pi\left ( \cdot \right )$). Finally, we can infer the quality score of the $i$-th dimension as:
\begin{equation}
    \hat{q}_i = \psi\left ( f_q^i|\pi\left ( f_c^i \right ) \right ).
\end{equation}

% The final loss function for training is defined as
% \begin{equation}
%     L = \frac{1}{K}\sum_{i=1}^K(\hat{q}_{i}-q_i)^2 + \frac{1}{T}\sum_{i\neq j, \{i,j\}\neq K}\max(\epsilon, \mathrm{cos}(f_c^i, f_c^j)).
% \end{equation}

\section{Experiment}

\subsection{Implement Details}
\textbf{Training Strategy.} We implement our metric by PyTorch and conduct training
and testing on the NVIDIA 3090 GPUs. All textured meshes are
rendered into $M=6$ projected images with a spatial resolution of 512 × 512 by PyTorch3D. More details for network implementation can be found in the appendix.

\textbf{Training-test Split.}
We apply a 5-fold cross-validation for MATE-3D while ensuring that there is no prompt overlap between the training and testing sets. Specifically, the ratios of prompts in the training set and test
set are 128:32. For each fold, the performance on the test set with minimal training loss is recorded, and the average performance across all folds is recorded as the final result.

% \textbf{Training Strategy}
% We train our HyperScore on MATE-3D for 30 epochs with a batch size of 8. During the training and testing process, all rendered images are resized into the resolution of $224\times224$. We use the Adam \cite{} optimizer with weight decay $1e-4$. The learning rate is set separately as $2e-6$ and $2e-4$ for the pre-trained visual encoder and other parts, and is reduced by a rate of 0.9 every 5 epochs. 

% \textbf{Network Details}
% The used visual encoder is Vision Transformer~\cite{} with 16 × 16 patch embeddings (namely ViT-B/16) in CLIP-Visual. The visual/textual/condition features have a dimension of $D=512$ while the quality feature has a dimension of $D_q=224$.

\textbf{Evaluation Metrics.} We utilize Pearson Linear Correlation
Coefficient (PLCC), Spearman Rank order Correlation Coefficient (SRCC), and Kendall’s Rank order Correlation Coefficient (KRCC) as performance criteria. Better metrics should have higher SRCC, KRCC, and PLCC. 
%Before calculating PLCC, we follow \cite{fittingfunction} to map the objective score to the subject score using a logistic function.

% \begin{figure}
%     \centering
%     \includegraphics[width=0.9\linewidth]{image_my/qualitative_sample.pdf}
%     \caption{Exemplary samples with their MOSs and the predicted scores of different metrics. The four scores in each row denote alignment, geometry, texture, and overall quality, respectively.}
%     \label{fig:qualitative_sample}
%     \vspace{-0.3cm}
% \end{figure}

\begin{figure}
    \centering
    \includegraphics[width=\linewidth]{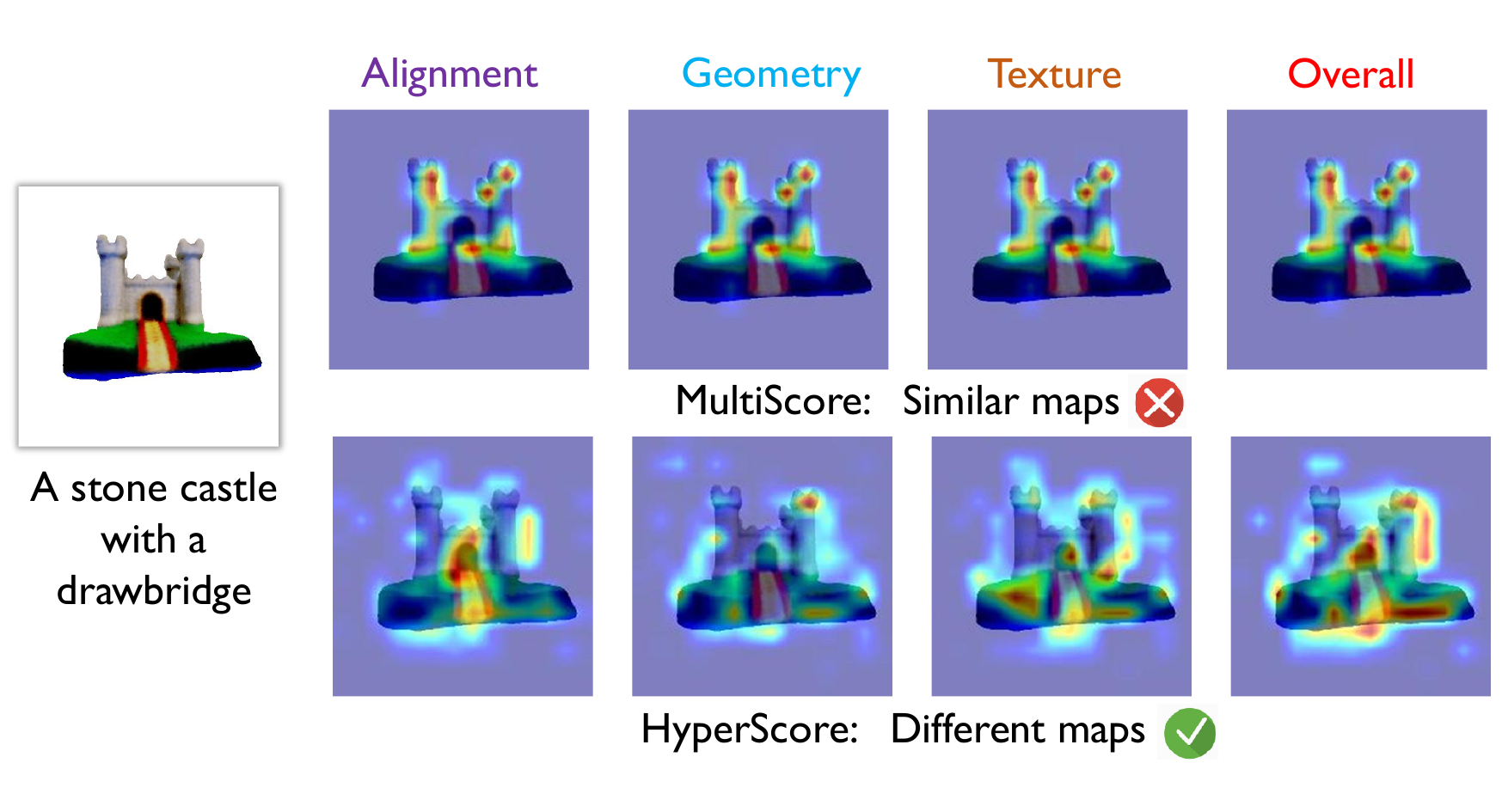}
    \caption{Visual explanations generated by XGrad-CAM for different evaluation dimensions.}
    \label{fig:gradcam}
    \vspace{-0.5cm}
\end{figure}

% \begin{figure}
%     \centering
%     \includegraphics[width=\linewidth]{image/scatter plot.pdf}
%     \caption{The framework of the proposed multi-dimensional quality evaluator.}
%     \label{fig:framework}
% \end{figure}

\subsection{Performance Comparison}
\textbf{Quantitative Analysis.}
Table \ref{tab:overall_performance} lists the experimental results on MATE-3D. We select widely used zero-shot metrics for text-to-image(or 3D) evaluation, including CLIPscore \cite{hessel2021clipscore}, BLIPScore \cite{li2022blip}, Aesthetic Score \cite{schuhmann2022laion}, ImageReward \cite{xu2024imagereward}, HPS v2 \cite{wu2023human}, CLIP-IQA \cite{wang2023exploring} and Q-Align \cite{wu2023q}. Note that we average the scores of multiple rendered viewpoints for these metrics. In addition, we also equip some prevalent backbones including ResNet50 \cite{he2016deep}, ViT-B \cite{dosovitskiy2020image}, SwinT-B \cite{liu2021swin} pre-trained on ImageNet \cite{deng2009imagenet} and ViT-B in DINO v2 \cite{oquab2023dinov2} with multiple regression heads and fine-tune them on MATE-3D. Finally, we propose a baseline named MultiScore that has the same feature extractor as HyperScore but only applies a multi-task learning strategy.
%and the best metric in each column are marked in \textbf{boldface}.

From \cref{tab:overall_performance}, we have the following observations: i) The proposed HyperScore exhibits outstanding performance across all evaluation dimensions, which demonstrates its effectiveness for multi-dimensional assessment. Especially, HyperScore presents a noticeable gain over MultiScore, justifying the potential of conditional learning strategy. ii)  ImageReward performs best among these zero-shot metrics, which may be attributed to its rank-based training strategy. iii) Aesthetic Score provides inferior results, probably due to its preference for aesthetic quality. CLIP-IQA also performs poorly, possibly because they are designed for natural images instead of rendered images of textured meshes.

\textbf{Visualization}
For better illustration, we present a qualitative analysis using a prevalent tool (\textit{i.e.}, XGrad-CAM \cite{fu2020axiom}) for network explanation. As illustrated in \cref{fig:gradcam}, the XGrad-CAM maps of MultiScore (top row) and HyperScore (bottom row) reflect what makes the network perform decisions for different evaluation dimensions.  We can see that the multi-head network fails to distinguish different dimensions while HyperScore exhibits different focus areas corresponding to changes in evaluation dimensions, which validates the effectiveness of the proposed strategy. Notably, for HyperScore, the XGrad-CAM maps related to the geometry evaluation pay more attention to the shape and structure of objects. In contrast, the texture and overall quality evaluations have a wider range of focus, likely because they require more appearance details to make decisions.

% \textbf{Qualitative Analysis.}
% For better illustration, we also conduct a qualitative analysis on several examples shown in \cref{fig:qualitative_sample}. From the figure, we can see: i) One sample has diverse scores for different quality
% dimensions, highlighting the drawbacks of traditional evaluators (\textit{e.g.}, CLIPScore) because they can only provide a single score. ii) MultiScore gives similar scores for different dimensions, which affects the judgment for different factors. For the first sample, MultiScore gives a lower score for texture quality than for overall quality (3.88 \textit{vs.} 4.19), which contradicts the subjective rating (4.81 \textit{vs.} 4.43). In contrast, HyperScore achieves a more discriminated prediction and presents an accurate ranking between the above two dimensions (4.87 \textit{vs.} 4.64), further demonstrating the effectiveness of our metric.

\begin{table}[t]

  \centering
  \caption{Ablation study for the key modules. `\Checkmark' or `\XSolidBrush' means the setting is preserved or discarded.}

  \resizebox{\linewidth}{!}{
    \begin{tabular}{c|cc|cccc}
    \toprule
    Index & CFF & AQM   & Alignment & Geometry & Texture & Overall\\
    \midrule
    (1)     & \XSolidBrush     & \XSolidBrush  &0.638	&0.703	&0.729	&0.698
 \\
     (2) & \Checkmark     & \XSolidBrush       &0.660	&0.730	&0.760	&0.724

 \\
    (3)     & \XSolidBrush      & \Checkmark    &0.721	&0.762	&0.792	&0.776

    \\
    (4) & \Checkmark     & \Checkmark      &\bf{0.739}	&\bf{0.782}	&\bf{0.811}	&\bf{0.792}

  \\ \midrule
\multicolumn{3}{c|}{Separately Trained} &0.737	&0.770	&0.798	&0.778

  \\
    \bottomrule
    \end{tabular}}%
  \label{tab:ablation_module}%
   \vspace{-0.6cm}
\end{table}%

\subsection{Ablation Study}
We further investigate the contribution of different components and report the results (in terms of SRCC) in \cref{tab:ablation_module}.  In the table, CFF and AQM denote conditional feature fusion and adaptive quality mapping, respectively. The baseline (denoted by (1)) applies a multi-task learning strategy and simply averages all patches in the visual feature. \textit{Separately Trained}
denotes separately trained networks for each dimension. From the table, we can see: i) Comparing (2) or (3) with (1), both CFF and AQM contribute to the performance, demonstrating the effectiveness of the two modules. Especially, the AQM brings noticeable gain over the baseline, justifying the use of hypernetwork in our metric. ii) Seeing (2)-(4), the full model performs better than using only single module, which verifies the complementarity of the two modules. iii) Comparing (1), (4) and \textit{Separately Trained}, directly training multi-head network results into performance drop compared to
separately trained networks for each dimension  while HyperScore can even achieve better performance, justifying the proposed strategy. 

% \textbf{Ablation Study for the Aggregation Strategy.}
% We test the performance of HyperScore with different aggregation strategies between
% visual and textual features in \cref{eq:feature_aggregate}. Except for the used element-wise multiplication (denoted by $\odot$), we choose two other strategies, \textit{i.e.}, addition (denoted by $+$) and concatenation (denoted by $\oplus$). According to the results in \cref{tab:ablation_aggregation}, we can see that the element-wise multiplication performs the best on three quality dimensions, and the addition provides relatively inferior performance, which justifies our choice for the aggregation strategy.

% \textbf{Ablation Study for the View Count.} We test the performance of HyperScore under different $M$ values to investigate the influence of the number of viewpoints and report the results in \cref{tab:ablation_viewpoint}. From the table, it can be observed that as the $M$ value increases, the performance initially improves and then decreases. The reason may be that increasing $M$ provides more information for a better prediction when $M$ is a small number. However, when $M$ becomes large, the increase in $L$ can affect performance to some extent due to redundancy in information. Meanwhile, a large $M$ also incurs high computational complexity. Therefore, to achieve the balance between self-describing ability and time complexity, we consider $M=6$ as a suitable choice.

\section{Conclusion}
In this paper, we are dedicated to exploring a subjective-aligned quality evaluation for text-to-3D generation. For this purpose, we first establish a comprehensive benchmark named MATE-3D containing 1,280 textured mesh samples annotated from four evaluation dimensions. 
Based on MATE-3D, we propose a multi-dimensional quality evaluator named HyperScore. Based on the merits of hypernetwork, the evaluator can perform adaptive multi-dimensional prediction. Experimental results indicate that HyperScore is effective in evaluating text-to-3D generation.

\section*{Acknowledgments}
This paper is supported in part by National Natural Science Foundation of China (62371290), National Key R\&D Program of China (2024YFB2907204), the Fundamental Research Funds for the Central Universities of China, and STCSM under Grant (22DZ2229005). 

{
    \small
    \bibliographystyle{ieeenat_fullname}
    \bibliography{main}

\begin{thebibliography}{80}
\providecommand{\natexlab}[1]{#1}
\providecommand{\url}[1]{\texttt{#1}}
\expandafter\ifx\csname urlstyle\endcsname\relax
  \providecommand{\doi}[1]{doi: #1}\else
  \providecommand{\doi}{doi: \begingroup \urlstyle{rm}\Url}\fi

\bibitem[Chen et~al.(2025)Chen, Xu, Esposito, Tang, and Geiger]{chen2025lara}
Anpei Chen, Haofei Xu, Stefano Esposito, Siyu Tang, and Andreas Geiger.
\newblock Lara: Efficient large-baseline radiance fields.
\newblock In \emph{European Conference on Computer Vision (ECCV)}, 2025.

\bibitem[Chen et~al.(2024{\natexlab{a}})Chen, Yang, Yang, Feng, Fu, Foo, Lin, and Liu]{chen2024sculpt3d}
Cheng Chen, Xiaofeng Yang, Fan Yang, Chengzeng Feng, Zhoujie Fu, Chuan-Sheng Foo, Guosheng Lin, and Fayao Liu.
\newblock Sculpt3d: Multi-view consistent text-to-3d generation with sparse 3d prior.
\newblock In \emph{IEEE/CVF Conference on Computer Vision and Pattern Recognition (CVPR)}, 2024{\natexlab{a}}.

\bibitem[Chen et~al.(2024{\natexlab{b}})Chen, Pan, Yang, Yao, and Mei]{chen2024vp3d}
Yang Chen, Yingwei Pan, Haibo Yang, Ting Yao, and Tao Mei.
\newblock Vp3d: Unleashing 2d visual prompt for text-to-3d generation.
\newblock In \emph{IEEE/CVF Conference on Computer Vision and Pattern Recognition (CVPR)}, 2024{\natexlab{b}}.

\bibitem[Deitke et~al.(2023{\natexlab{a}})Deitke, Liu, Wallingford, Ngo, Michel, Kusupati, Fan, Laforte, Voleti, Gadre, VanderBilt, Kembhavi, Vondrick, Gkioxari, Ehsani, Schmidt, and Farhadi]{deitke2023objaversexl}
Matt Deitke, Ruoshi Liu, Matthew Wallingford, Huong Ngo, Oscar Michel, Aditya Kusupati, Alan Fan, Christian Laforte, Vikram Voleti, Samir~Yitzhak Gadre, Eli VanderBilt, Aniruddha Kembhavi, Carl Vondrick, Georgia Gkioxari, Kiana Ehsani, Ludwig Schmidt, and Ali Farhadi.
\newblock Objaverse-xl: A universe of 10m+ 3d objects.
\newblock In \emph{Advances in Neural Information Processing Systems (NIPS)}, 2023{\natexlab{a}}.

\bibitem[Deitke et~al.(2023{\natexlab{b}})Deitke, Schwenk, Salvador, Weihs, Michel, VanderBilt, Schmidt, Ehsanit, Kembhavi, and Farhadi]{deitke2023objaverse}
Matt Deitke, Dustin Schwenk, Jordi Salvador, Luca Weihs, Oscar Michel, Eli VanderBilt, Ludwig Schmidt, Kiana Ehsanit, Aniruddha Kembhavi, and Ali Farhadi.
\newblock Objaverse: A universe of annotated 3d objects.
\newblock In \emph{IEEE/CVF Conference on Computer Vision and Pattern Recognition (CVPR)}, 2023{\natexlab{b}}.

\bibitem[Deng et~al.(2009)Deng, Dong, Socher, Li, Li, and Fei-Fei]{deng2009imagenet}
Jia Deng, Wei Dong, Richard Socher, Li-Jia Li, Kai Li, and Li Fei-Fei.
\newblock Imagenet: A large-scale hierarchical image database.
\newblock In \emph{IEEE/CVF Conference on Computer Vision and Pattern Recognition (CVPR)}, 2009.

\bibitem[Dong et~al.(2024)Dong, Ding, Huang, Wang, Xue, and Xu]{dong2024interactive3d}
Shaocong Dong, Lihe Ding, Zhanpeng Huang, Zibin Wang, Tianfan Xue, and Dan Xu.
\newblock Interactive3d: Create what you want by interactive 3d generation.
\newblock In \emph{IEEE/CVF Conference on Computer Vision and Pattern Recognition (CVPR)}, 2024.

\bibitem[Dosovitskiy(2020)]{dosovitskiy2020image}
Alexey Dosovitskiy.
\newblock An image is worth 16x16 words: Transformers for image recognition at scale.
\newblock \emph{arXiv preprint arXiv:2010.11929}, 2020.

\bibitem[Fu et~al.(2020)Fu, Hu, Dong, Guo, Gao, and Li]{fu2020axiom}
Ruigang Fu, Qingyong Hu, Xiaohu Dong, Yulan Guo, Yinghui Gao, and Biao Li.
\newblock Axiom-based grad-cam: Towards accurate visualization and explanation of cnns.
\newblock \emph{arXiv preprint arXiv:2008.02312}, 2020.

\bibitem[Fu et~al.(2022)Fu, Zhan, Chen, Ritchie, and Sridhar]{fu2022shapecrafter}
Rao Fu, Xiao Zhan, Yiwen Chen, Daniel Ritchie, and Srinath Sridhar.
\newblock Shapecrafter: A recursive text-conditioned 3d shape generation model.
\newblock In \emph{Advances in Neural Information Processing Systems (NIPS)}, 2022.

\bibitem[Gao et~al.(2022)Gao, Shen, Wang, Chen, Yin, Li, Litany, Gojcic, and Fidler]{gao2022get3d}
Jun Gao, Tianchang Shen, Zian Wang, Wenzheng Chen, Kangxue Yin, Daiqing Li, Or Litany, Zan Gojcic, and Sanja Fidler.
\newblock Get3d: A generative model of high quality 3d textured shapes learned from images.
\newblock In \emph{Advances in Neural Information Processing Systems (NIPS)}, 2022.

\bibitem[Ha et~al.(2022)Ha, Dai, and Le]{ha2022hypernetworks}
David Ha, Andrew~M Dai, and Quoc~V Le.
\newblock Hypernetworks.
\newblock In \emph{International Conference on Learning Representations (ICLR)}, 2022.

\bibitem[He et~al.(2016)He, Zhang, Ren, and Sun]{he2016deep}
Kaiming He, Xiangyu Zhang, Shaoqing Ren, and Jian Sun.
\newblock Deep residual learning for image recognition.
\newblock In \emph{IEEE/CVF Conference on Computer Vision and Pattern Recognition (CVPR)}, 2016.

\bibitem[He et~al.(2023)He, Bai, Lin, Zhao, Hu, Sheng, Yi, Li, and Liu]{he2023t3bench}
Yuze He, Yushi Bai, Matthieu Lin, Wang Zhao, Yubin Hu, Jenny Sheng, Ran Yi, Juanzi Li, and Yong-Jin Liu.
\newblock T$^3$bench: Benchmarking current progress in text-to-3d generation.
\newblock \emph{arXiv preprint arXiv:2310.02977}, 2023.

\bibitem[He et~al.(2024)He, Wang, Hu, Zhao, Yi, Liu, and Wang]{he2024mmpi}
Yuze He, Peng Wang, Yubin Hu, Wang Zhao, Ran Yi, Yong-Jin Liu, and Wenping Wang.
\newblock Mmpi: a flexible radiance field representation by multiple multi-plane images blending.
\newblock In \emph{International Conference on Robotics and Automation (ICRA)}, 2024.

\bibitem[Hessel et~al.(2021)Hessel, Holtzman, Forbes, Le~Bras, and Choi]{hessel2021clipscore}
Jack Hessel, Ari Holtzman, Maxwell Forbes, Ronan Le~Bras, and Yejin Choi.
\newblock {CLIPS}core: A reference-free evaluation metric for image captioning.
\newblock In \emph{Empirical Methods in Natural Language Processing (EMNLP)}, 2021.

\bibitem[Hong et~al.(2024{\natexlab{a}})Hong, Tang, Cao, Shi, Wu, Chen, Wang, Pan, Lin, and Liu]{3dtopia}
Fangzhou Hong, Jiaxiang Tang, Ziang Cao, Min Shi, Tong Wu, Zhaoxi Chen, Tengfei Wang, Liang Pan, Dahua Lin, and Ziwei Liu.
\newblock 3dtopia: Large text-to-3d generation model with hybrid diffusion priors.
\newblock \emph{arXiv preprint arXiv:2403.02234}, 2024{\natexlab{a}}.

\bibitem[Hong et~al.(2023)Hong, Ahn, and Kim]{hong2023debiasing}
Susung Hong, Donghoon Ahn, and Seungryong Kim.
\newblock Debiasing scores and prompts of 2d diffusion for view-consistent text-to-3d generation.
\newblock In \emph{Advances in Neural Information Processing Systems (NIPS)}, 2023.

\bibitem[Hong et~al.(2024{\natexlab{b}})Hong, Zhang, Gu, Bi, Zhou, Liu, Liu, Sunkavalli, Bui, and Tan]{hong2023lrm}
Yicong Hong, Kai Zhang, Jiuxiang Gu, Sai Bi, Yang Zhou, Difan Liu, Feng Liu, Kalyan Sunkavalli, Trung Bui, and Hao Tan.
\newblock Lrm: Large reconstruction model for single image to 3d.
\newblock In \emph{International Conference on Learning Representations (ICLR)}, 2024{\natexlab{b}}.

\bibitem[Hou et~al.(2023)Hou, Lin, Fang, Wu, Chen, Liao, and Liu]{hou2023towards}
Jingwen Hou, Weisi Lin, Yuming Fang, Haoning Wu, Chaofeng Chen, Liang Liao, and Weide Liu.
\newblock Towards transparent deep image aesthetics assessment with tag-based content descriptors.
\newblock \emph{IEEE Transactions on Image Processing (TIP)}, 2023.

\bibitem[Huang et~al.(2023)Huang, Cao, Lai, Shan, and Gao]{huang2023nerftexture}
Yi-Hua Huang, Yan-Pei Cao, Yu-Kun Lai, Ying Shan, and Lin Gao.
\newblock Nerf-texture: Texture synthesis with neural radiance fields.
\newblock In \emph{ACM SIGGRAPH}, 2023.

\bibitem[Hui et~al.(2024)Hui, Sanghi, Rampini, Rahimi~Malekshan, Liu, Shayani, and Fu]{hui2024makeashape}
Ka-Hei Hui, Aditya Sanghi, Arianna Rampini, Kamal Rahimi~Malekshan, Zhengzhe Liu, Hooman Shayani, and Chi-Wing Fu.
\newblock Make-a-shape: a ten-million-scale 3{D} shape model.
\newblock In \emph{International Conference on Machine Learning (ICML)}, 2024.

\bibitem[Jaganathan et~al.(2024)Jaganathan, Huang, Irshad, Jampani, Raj, and Kira]{jaganathan2024ice}
Vishnu Jaganathan, Hannah~Hanyun Huang, Muhammad~Zubair Irshad, Varun Jampani, Amit Raj, and Zsolt Kira.
\newblock Ice-g: Image conditional editing of 3d gaussian splats.
\newblock \emph{arXiv preprint arXiv:2406.08488}, 2024.

\bibitem[Jun and Nichol(2023)]{jun2023shape}
Heewoo Jun and Alex Nichol.
\newblock Shap-e: Generating conditional 3d implicit functions.
\newblock \emph{arXiv preprint arXiv:2305.02463}, 2023.

\bibitem[Kerbl et~al.(2023)Kerbl, Kopanas, Leimk{\"u}hler, and Drettakis]{kerbl20233dgs}
Bernhard Kerbl, Georgios Kopanas, Thomas Leimk{\"u}hler, and George Drettakis.
\newblock 3d gaussian splatting for real-time radiance field rendering.
\newblock \emph{ACM Transactions On Graphics (TOG)}, 2023.

\bibitem[Kingma and Ba(2014)]{kingma2014adam}
Diederik~P Kingma and Jimmy Ba.
\newblock Adam: A method for stochastic optimization.
\newblock \emph{arXiv preprint arXiv:1412.6980}, 2014.

\bibitem[Li et~al.(2024{\natexlab{a}})Li, Lin, Pathak, Li, Fei, Wu, Xia, Zhang, Neubig, and Ramanan]{li2024evaluating}
Baiqi Li, Zhiqiu Lin, Deepak Pathak, Jiayao Li, Yixin Fei, Kewen Wu, Xide Xia, Pengchuan Zhang, Graham Neubig, and Deva Ramanan.
\newblock Evaluating and improving compositional text-to-visual generation.
\newblock In \emph{IEEE/CVF Conference on Computer Vision and Pattern Recognition (CVPR)}, 2024{\natexlab{a}}.

\bibitem[Li et~al.(2024{\natexlab{b}})Li, Kou, Gao, Cao, Sun, Zhang, Zhou, Zhang, Zhang, Wu, Liu, Min, and Zhai]{li2024aigiqa20k}
Chunyi Li, Tengchuan Kou, Yixuan Gao, Yuqin Cao, Wei Sun, Zicheng Zhang, Yingjie Zhou, Zhichao Zhang, Weixia Zhang, Haoning Wu, Xiaohong Liu, Xiongkuo Min, and Guangtao Zhai.
\newblock Aigiqa-20k: A large database for ai-generated image quality assessment.
\newblock In \emph{IEEE/CVF Conference on Computer Vision and Pattern Recognition Workshops (CVPRW)}, 2024{\natexlab{b}}.

\bibitem[Li et~al.(2024{\natexlab{c}})Li, Zhang, Wu, Sun, Min, Liu, Zhai, and Lin]{liagiqa3k}
Chunyi Li, Zicheng Zhang, Haoning Wu, Wei Sun, Xiongkuo Min, Xiaohong Liu, Guangtao Zhai, and Weisi Lin.
\newblock Agiqa-3k: An open database for ai-generated image quality assessment.
\newblock \emph{IEEE Transactions on Circuits and Systems for Video Technology (TCSVT)}, 2024{\natexlab{c}}.

\bibitem[Li et~al.(2022)Li, Li, Xiong, and Hoi]{li2022blip}
Junnan Li, Dongxu Li, Caiming Xiong, and Steven Hoi.
\newblock Blip: Bootstrapping language-image pre-training for unified vision-language understanding and generation.
\newblock In \emph{International Conference on Machine Learning (ICML)}, 2022.

\bibitem[Li et~al.(2024{\natexlab{d}})Li, Zhou, Liu, Keppo, Lin, Yan, and Xu]{li2024instant3d}
Ming Li, Pan Zhou, Jia-Wei Liu, Jussi Keppo, Min Lin, Shuicheng Yan, and Xiangyu Xu.
\newblock Instant3d: Instant text-to-3d generation.
\newblock \emph{International Journal of Computer Vision (IJCV)}, 2024{\natexlab{d}}.

\bibitem[Library()]{threejs}
Three.js Library.
\newblock https://threejs.org.

\bibitem[Lin et~al.(2023)Lin, Gao, Tang, Takikawa, Zeng, Huang, Kreis, Fidler, Liu, and Lin]{magic3d}
Chen-Hsuan Lin, Jun Gao, Luming Tang, Towaki Takikawa, Xiaohui Zeng, Xun Huang, Karsten Kreis, Sanja Fidler, Ming-Yu Liu, and Tsung-Yi Lin.
\newblock Magic3d: High-resolution text-to-3d content creation.
\newblock In \emph{IEEE/CVF Conference on Computer Vision and Pattern Recognition (CVPR)}, 2023.

\bibitem[Liu et~al.(2024{\natexlab{a}})Liu, Wu, Wei, Rao, and Duan]{liu2024sherpa3d}
Fangfu Liu, Diankun Wu, Yi Wei, Yongming Rao, and Yueqi Duan.
\newblock Sherpa3d: Boosting high-fidelity text-to-3d generation via coarse 3d prior.
\newblock In \emph{IEEE/CVF Conference on Computer Vision and Pattern Recognition (CVPR)}, 2024{\natexlab{a}}.

\bibitem[Liu et~al.(2024{\natexlab{b}})Liu, Huang, Huang, Chen, Hou, Tang, Liu, Ouyang, Zuo, Jiang, et~al.]{liu2024comprehensive}
Jian Liu, Xiaoshui Huang, Tianyu Huang, Lu Chen, Yuenan Hou, Shixiang Tang, Ziwei Liu, Wanli Ouyang, Wangmeng Zuo, Junjun Jiang, et~al.
\newblock A comprehensive survey on 3d content generation.
\newblock \emph{arXiv preprint arXiv:2402.01166}, 2024{\natexlab{b}}.

\bibitem[Liu et~al.(2024{\natexlab{c}})Liu, Shi, Chen, Zhang, Xu, Wei, Chen, Zeng, Gu, and Su]{one2345++}
Minghua Liu, Ruoxi Shi, Linghao Chen, Zhuoyang Zhang, Chao Xu, Xinyue Wei, Hansheng Chen, Chong Zeng, Jiayuan Gu, and Hao Su.
\newblock One-2-3-45++: Fast single image to 3d objects with consistent multi-view generation and 3d diffusion.
\newblock In \emph{IEEE/CVF Conference on Computer Vision and Pattern Recognition (CVPR)}, 2024{\natexlab{c}}.

\bibitem[Liu et~al.(2023)Liu, Wu, Van~Hoorick, Tokmakov, Zakharov, and Vondrick]{liu2023zero1to3}
Ruoshi Liu, Rundi Wu, Basile Van~Hoorick, Pavel Tokmakov, Sergey Zakharov, and Carl Vondrick.
\newblock Zero-1-to-3: Zero-shot one image to 3d object.
\newblock In \emph{International Conference on Computer Vision (ICCV)}, 2023.

\bibitem[Liu et~al.(2024{\natexlab{d}})Liu, Yang, Wu, Ren, Lin, Liu, Liu, and Shan]{liu2024vqa}
Yibo Liu, Zheyuan Yang, Guile Wu, Yuan Ren, Kejian Lin, Bingbing Liu, Yang Liu, and Jinjun Shan.
\newblock Vqa-diff: Exploiting vqa and diffusion for zero-shot image-to-3d vehicle asset generation in autonomous driving.
\newblock In \emph{European Conference on Computer Vision (ECCV)}, 2024{\natexlab{d}}.

\bibitem[Liu et~al.(2021)Liu, Lin, Cao, Hu, Wei, Zhang, Lin, and Guo]{liu2021swin}
Ze Liu, Yutong Lin, Yue Cao, Han Hu, Yixuan Wei, Zheng Zhang, Stephen Lin, and Baining Guo.
\newblock Swin transformer: Hierarchical vision transformer using shifted windows.
\newblock In \emph{IEEE/CVF Conference on Computer Vision and Pattern Recognition (CVPR)}, 2021.

\bibitem[Ma et~al.(2024)Ma, Wei, Zhang, Zhu, Lei, and Zhang]{ma2024scaledreamer}
Zhiyuan Ma, Yuxiang Wei, Yabin Zhang, Xiangyu Zhu, Zhen Lei, and Lei Zhang.
\newblock Scaledreamer: Scalable text-to-3d synthesis with asynchronous score distillation.
\newblock In \emph{European Conference on Computer Vision (ECCV)}, 2024.

\bibitem[Metzer et~al.(2023)Metzer, Richardson, Patashnik, Giryes, and Cohen-Or]{latentnerf}
Gal Metzer, Elad Richardson, Or Patashnik, Raja Giryes, and Daniel Cohen-Or.
\newblock Latent-nerf for shape-guided generation of 3d shapes and textures.
\newblock In \emph{IEEE/CVF Conference on Computer Vision and Pattern Recognition (CVPR)}, 2023.

\bibitem[Mildenhall et~al.(2020)Mildenhall, Srinivasan, Tancik, Barron, Ramamoorthi, and Ng]{mildenhall2021nerf}
Ben Mildenhall, Pratul~P Srinivasan, Matthew Tancik, Jonathan~T Barron, Ravi Ramamoorthi, and Ren Ng.
\newblock Nerf: Representing scenes as neural radiance fields for view synthesis.
\newblock In \emph{European Conference on Computer Vision (ECCV)}, 2020.

\bibitem[Mohammad~Khalid et~al.(2022)Mohammad~Khalid, Xie, Belilovsky, and Popa]{mohammad2022clipmesh}
Nasir Mohammad~Khalid, Tianhao Xie, Eugene Belilovsky, and Tiberiu Popa.
\newblock Clip-mesh: Generating textured meshes from text using pretrained image-text models.
\newblock In \emph{ACM SIGGRAPH Asia}, 2022.

\bibitem[Nichol et~al.(2022)Nichol, Jun, Dhariwal, Mishkin, and Chen]{nichol2022pointe}
Alex Nichol, Heewoo Jun, Prafulla Dhariwal, Pamela Mishkin, and Mark Chen.
\newblock Point-e: A system for generating 3d point clouds from complex prompts.
\newblock \emph{arXiv preprint arXiv:2212.08751}, 2022.

\bibitem[OpenAI(2023)]{openaigpt}
OpenAI.
\newblock Gpt-4 system card.
\newblock \emph{OpenAI}, 2023.

\bibitem[Oquab et~al.(2023)Oquab, Darcet, Moutakanni, Vo, Szafraniec, Khalidov, Fernandez, Haziza, Massa, El-Nouby, et~al.]{oquab2023dinov2}
Maxime Oquab, Timoth{\'e}e Darcet, Th{\'e}o Moutakanni, Huy Vo, Marc Szafraniec, Vasil Khalidov, Pierre Fernandez, Daniel Haziza, Francisco Massa, Alaaeldin El-Nouby, et~al.
\newblock Dinov2: Learning robust visual features without supervision.
\newblock \emph{arXiv preprint arXiv:2304.07193}, 2023.

\bibitem[Poole et~al.(2023)Poole, Jain, Barron, and Mildenhall]{dreamfusion}
Ben Poole, Ajay Jain, Jonathan~T. Barron, and Ben Mildenhall.
\newblock Dreamfusion: Text-to-3d using 2d diffusion.
\newblock In \emph{International Conference on Learning Representations (ICLR)}, 2023.

\bibitem[Radford et~al.(2021)Radford, Kim, Hallacy, Ramesh, Goh, Agarwal, Sastry, Askell, Mishkin, Clark, et~al.]{radford2021learning}
Alec Radford, Jong~Wook Kim, Chris Hallacy, Aditya Ramesh, Gabriel Goh, Sandhini Agarwal, Girish Sastry, Amanda Askell, Pamela Mishkin, Jack Clark, et~al.
\newblock Learning transferable visual models from natural language supervision.
\newblock In \emph{International Conference on Machine Learning (ICML)}, 2021.

\bibitem[Recommendation(2019)]{bt500}
ITU-R~BT.500 Recommendation.
\newblock Methodologies for the subjective assessment of the quality of television images.
\newblock 2019.

\bibitem[Recommendation(1999)]{p910}
ITU-T~P.910 Recommendation.
\newblock Subjective video quality assessment methods for multimedia applications.
\newblock 1999.

\bibitem[Ren et~al.(2023)Ren, Pan, Tang, Zhang, Cao, Zeng, and Liu]{ren2023dreamgaussian4d}
Jiawei Ren, Liang Pan, Jiaxiang Tang, Chi Zhang, Ang Cao, Gang Zeng, and Ziwei Liu.
\newblock Dreamgaussian4d: Generative 4d gaussian splatting.
\newblock \emph{arXiv preprint arXiv:2312.17142}, 2023.

\bibitem[Schuhmann et~al.(2022)Schuhmann, Beaumont, Vencu, Gordon, Wightman, Cherti, Coombes, Katta, Mullis, Wortsman, et~al.]{schuhmann2022laion}
Christoph Schuhmann, Romain Beaumont, Richard Vencu, Cade Gordon, Ross Wightman, Mehdi Cherti, Theo Coombes, Aarush Katta, Clayton Mullis, Mitchell Wortsman, et~al.
\newblock Laion-5b: An open large-scale dataset for training next generation image-text models.
\newblock In \emph{Advances in Neural Information Processing Systems (NIPS)}, 2022.

\bibitem[Seo et~al.(2024{\natexlab{a}})Seo, Hong, Jang, Kim, Kwak, Lee, and Kim]{seo2024retrievalaugmented}
Junyoung Seo, Susung Hong, Wooseok Jang, In{\`e}s~Hyeonsu Kim, Min-Seop Kwak, Doyup Lee, and Seungryong Kim.
\newblock Retrieval-augmented score distillation for text-to-3d generation.
\newblock In \emph{International Conference on Machine Learning (ICML)}, 2024{\natexlab{a}}.

\bibitem[Seo et~al.(2024{\natexlab{b}})Seo, Jang, Kwak, Kim, Ko, Kim, Kim, Lee, and Kim]{seo2023let}
Junyoung Seo, Wooseok Jang, Min-Seop Kwak, Hyeonsu Kim, Jaehoon Ko, Junho Kim, Jin-Hwa Kim, Jiyoung Lee, and Seungryong Kim.
\newblock Let 2d diffusion model know 3d-consistency for robust text-to-3d generation.
\newblock In \emph{International Conference on Learning Representations (ICLR)}, 2024{\natexlab{b}}.

\bibitem[Shi et~al.(2024)Shi, Wang, Ye, Long, Li, and Yang]{shi2024mvdream}
Yichun Shi, Peng Wang, Jianglong Ye, Mai Long, Kejie Li, and Xiao Yang.
\newblock Mvdream: Multi-view diffusion for 3d generation.
\newblock \emph{arXiv preprint arXiv:2308.16512}, 2024.

\bibitem[Su et~al.(2020)Su, Yan, Zhu, Zhang, Ge, Sun, and Zhang]{su2020blindly}
Shaolin Su, Qingsen Yan, Yu Zhu, Cheng Zhang, Xin Ge, Jinqiu Sun, and Yanning Zhang.
\newblock Blindly assess image quality in the wild guided by a self-adaptive hyper network.
\newblock In \emph{IEEE/CVF Conference on Computer Vision and Pattern Recognition (CVPR)}, 2020.

\bibitem[Tang et~al.(2024{\natexlab{a}})Tang, Chen, Chen, Wang, Zeng, and Liu]{tang2024lgm}
Jiaxiang Tang, Zhaoxi Chen, Xiaokang Chen, Tengfei Wang, Gang Zeng, and Ziwei Liu.
\newblock Lgm: Large multi-view gaussian model for high-resolution 3d content creation.
\newblock In \emph{European Conference on Computer Vision (ECCV)}, 2024{\natexlab{a}}.

\bibitem[Tang et~al.(2024{\natexlab{b}})Tang, Ren, Zhou, Liu, and Zeng]{ben2024dreamgaussian}
Jiaxiang Tang, Jiawei Ren, Hang Zhou, Ziwei Liu, and Gang Zeng.
\newblock Dreamgaussian: Generative gaussian splatting for efficient 3d content creation.
\newblock In \emph{International Conference on Learning Representations (ICLR)}, 2024{\natexlab{b}}.

\bibitem[Tsalicoglou et~al.(2024)Tsalicoglou, Manhardt, Tonioni, Niemeyer, and Tombari]{textmesh}
Christina Tsalicoglou, Fabian Manhardt, Alessio Tonioni, Michael Niemeyer, and Federico Tombari.
\newblock Textmesh: Generation of realistic 3d meshes from text prompts.
\newblock In \emph{International Conference on 3D Vision (3DV)}, 2024.

\bibitem[Wang et~al.(2023{\natexlab{a}})Wang, Du, Li, Yeh, and Shakhnarovich]{sjc}
Haochen Wang, Xiaodan Du, Jiahao Li, Raymond~A. Yeh, and Gregory Shakhnarovich.
\newblock Score jacobian chaining: Lifting pretrained 2d diffusion models for 3d generation.
\newblock In \emph{IEEE/CVF Conference on Computer Vision and Pattern Recognition (CVPR)}, 2023{\natexlab{a}}.

\bibitem[Wang et~al.(2023{\natexlab{b}})Wang, Chan, and Loy]{wang2023exploring}
Jianyi Wang, Kelvin~CK Chan, and Chen~Change Loy.
\newblock Exploring clip for assessing the look and feel of images.
\newblock In \emph{AAAI Conference on Artificial Intelligence}, 2023{\natexlab{b}}.

\bibitem[Wang et~al.(2024)Wang, Lu, Wang, Bao, Li, Su, and Zhu]{wang2024prolificdreamer}
Zhengyi Wang, Cheng Lu, Yikai Wang, Fan Bao, Chongxuan Li, Hang Su, and Jun Zhu.
\newblock Prolificdreamer: high-fidelity and diverse text-to-3d generation with variational score distillation.
\newblock In \emph{Advances in Neural Information Processing Systems (NIPS)}, 2024.

\bibitem[Wu et~al.(2023{\natexlab{a}})Wu, Zhang, Zhang, Chen, Liao, Li, Gao, Wang, Zhang, Sun, et~al.]{wu2023q}
Haoning Wu, Zicheng Zhang, Weixia Zhang, Chaofeng Chen, Liang Liao, Chunyi Li, Yixuan Gao, Annan Wang, Erli Zhang, Wenxiu Sun, et~al.
\newblock Q-align: Teaching lmms for visual scoring via discrete text-defined levels.
\newblock \emph{arXiv preprint arXiv:2312.17090}, 2023{\natexlab{a}}.

\bibitem[Wu et~al.(2024{\natexlab{a}})Wu, Yang, Li, Zhang, Liu, Guibas, Lin, and Wetzstein]{wu2024gpt4v}
Tong Wu, Guandao Yang, Zhibing Li, Kai Zhang, Ziwei Liu, Leonidas Guibas, Dahua Lin, and Gordon Wetzstein.
\newblock Gpt-4v(ision) is a human-aligned evaluator for text-to-3d generation.
\newblock In \emph{IEEE/CVF Conference on Computer Vision and Pattern Recognition (CVPR)}, 2024{\natexlab{a}}.

\bibitem[Wu et~al.(2023{\natexlab{b}})Wu, Hao, Sun, Chen, Zhu, Zhao, and Li]{wu2023human}
Xiaoshi Wu, Yiming Hao, Keqiang Sun, Yixiong Chen, Feng Zhu, Rui Zhao, and Hongsheng Li.
\newblock Human preference score v2: A solid benchmark for evaluating human preferences of text-to-image synthesis.
\newblock \emph{arXiv preprint arXiv:2306.09341}, 2023{\natexlab{b}}.

\bibitem[Wu et~al.(2024{\natexlab{b}})Wu, Zhou, Yi, Yuan, and Zhang]{consistent3d}
Zike Wu, Pan Zhou, Xuanyu Yi, Xiaoding Yuan, and Hanwang Zhang.
\newblock Consistent3d: Towards consistent high-fidelity text-to-3d generation with deterministic sampling prior.
\newblock In \emph{IEEE/CVF Conference on Computer Vision and Pattern Recognition (CVPR)}, 2024{\natexlab{b}}.

\bibitem[Xu et~al.(2023)Xu, Wang, Cheng, Cao, Shan, Qie, and Gao]{xu2023dream3d}
Jiale Xu, Xintao Wang, Weihao Cheng, Yan-Pei Cao, Ying Shan, Xiaohu Qie, and Shenghua Gao.
\newblock Dream3d: Zero-shot text-to-3d synthesis using 3d shape prior and text-to-image diffusion models.
\newblock In \emph{IEEE/CVF Conference on Computer Vision and Pattern Recognition (CVPR)}, 2023.

\bibitem[Xu et~al.(2024)Xu, Liu, Wu, Tong, Li, Ding, Tang, and Dong]{xu2024imagereward}
Jiazheng Xu, Xiao Liu, Yuchen Wu, Yuxuan Tong, Qinkai Li, Ming Ding, Jie Tang, and Yuxiao Dong.
\newblock Imagereward: Learning and evaluating human preferences for text-to-image generation.
\newblock In \emph{Advances in Neural Information Processing Systems (NIPS)}, 2024.

\bibitem[Yan et~al.(2024)Yan, Gao, Yang, Wei, Xie, Wu, and Zheng]{yan2024dreamview}
Junkai Yan, Yipeng Gao, Qize Yang, Xihan Wei, Xuansong Xie, Ancong Wu, and Wei-Shi Zheng.
\newblock Dreamview: Injecting view-specific text guidance into text-to-3d generation.
\newblock In \emph{European Conference on Computer Vision (ECCV)}, 2024.

\bibitem[Yan et~al.(2025)Yan, Chen, and Wang]{yan2025consistentflowdistillationtextto3d}
Runjie Yan, Yinbo Chen, and Xiaolong Wang.
\newblock Consistent flow distillation for text-to-3d generation.
\newblock In \emph{International Conference on Learning Representations (ICLR)}, 2025.

\bibitem[Yang et~al.(2024)Yang, Zuo, Ramasinghe, Bazzani, Avraham, and van~den Hengel]{yang2024viewfusion}
Xianghui Yang, Yan Zuo, Sameera Ramasinghe, Loris Bazzani, Gil Avraham, and Anton van~den Hengel.
\newblock Viewfusion: Towards multi-view consistency via interpolated denoising.
\newblock In \emph{IEEE/CVF Conference on Computer Vision and Pattern Recognition (CVPR)}, 2024.

\bibitem[Ye et~al.(2024{\natexlab{a}})Ye, Liu, Li, Wang, Wang, Wang, Duan, and Zhu]{ye2024dreamreward}
Junliang Ye, Fangfu Liu, Qixiu Li, Zhengyi Wang, Yikai Wang, Xinzhou Wang, Yueqi Duan, and Jun Zhu.
\newblock Dreamreward: Text-to-3d generation with human preference.
\newblock In \emph{European Conference on Computer Vision (ECCV)}, 2024{\natexlab{a}}.

\bibitem[Ye et~al.(2024{\natexlab{b}})Ye, Wang, Li, Shi, and Wang]{ye2023consistent1to3}
Jianglong Ye, Peng Wang, Kejie Li, Yichun Shi, and Heng Wang.
\newblock Consistent-1-to-3: Consistent image to 3d view synthesis via geometry-aware diffusion models.
\newblock In \emph{International Conference on 3D Vision (3DV)}, 2024{\natexlab{b}}.

\bibitem[Yi et~al.(2024{\natexlab{a}})Yi, Fang, Wang, Wu, Xie, Zhang, Liu, Tian, and Wang]{yi2024gaussiandreamer}
Taoran Yi, Jiemin Fang, Junjie Wang, Guanjun Wu, Lingxi Xie, Xiaopeng Zhang, Wenyu Liu, Qi Tian, and Xinggang Wang.
\newblock Gaussiandreamer: Fast generation from text to 3d gaussians by bridging 2d and 3d diffusion models.
\newblock In \emph{IEEE/CVF Conference on Computer Vision and Pattern Recognition (CVPR)}, 2024{\natexlab{a}}.

\bibitem[Yi et~al.(2024{\natexlab{b}})Yi, Fang, Zhou, Wang, Wu, Xie, Zhang, Liu, Wang, and Tian]{yi2024gaussiandreamerpro}
Taoran Yi, Jiemin Fang, Zanwei Zhou, Junjie Wang, Guanjun Wu, Lingxi Xie, Xiaopeng Zhang, Wenyu Liu, Xinggang Wang, and Qi Tian.
\newblock Gaussiandreamerpro: Text to manipulable 3d gaussians with highly enhanced quality.
\newblock \emph{CoRR}, 2024{\natexlab{b}}.

\bibitem[Zhang et~al.(2023)Zhang, Tang, Nie\ss{}ner, and Wonka]{zhang20233dshape2vecset}
Biao Zhang, Jiapeng Tang, Matthias Nie\ss{}ner, and Peter Wonka.
\newblock 3dshape2vecset: A 3d shape representation for neural fields and generative diffusion models.
\newblock \emph{ACM Transactions On Graphics (TOG)}, 2023.

\bibitem[Zhang et~al.(2024{\natexlab{a}})Zhang, Wang, Zhang, Qiu, Pang, Jiang, Yang, Xu, and Yu]{zhang2024clay}
Longwen Zhang, Ziyu Wang, Qixuan Zhang, Qiwei Qiu, Anqi Pang, Haoran Jiang, Wei Yang, Lan Xu, and Jingyi Yu.
\newblock Clay: A controllable large-scale generative model for creating high-quality 3d assets.
\newblock \emph{ACM Transactions on Graphics (TOG)}, 2024{\natexlab{a}}.

\bibitem[Zhang et~al.(2024{\natexlab{b}})Zhang, Wang, Wu, Li, Gao, Zhang, and Wang]{zhang2024learning}
Sixian Zhang, Bohan Wang, Junqiang Wu, Yan Li, Tingting Gao, Di Zhang, and Zhongyuan Wang.
\newblock Learning multi-dimensional human preference for text-to-image generation.
\newblock In \emph{IEEE/CVF Conference on Computer Vision and Pattern Recognition (CVPR)}, 2024{\natexlab{b}}.

\bibitem[Zhou et~al.(2022)Zhou, Yang, Loy, and Liu]{zhou2022learning}
Kaiyang Zhou, Jingkang Yang, Chen~Change Loy, and Ziwei Liu.
\newblock Learning to prompt for vision-language models.
\newblock \emph{International Journal of Computer Vision (IJCV)}, 2022.

\bibitem[Zhu et~al.(2024)Zhu, Zhuang, and Koyejo]{zhu2023hifa}
Junzhe Zhu, Peiye Zhuang, and Sanmi Koyejo.
\newblock Hifa: High-fidelity text-to-3d generation with advanced diffusion guidance.
\newblock In \emph{International Conference on Learning Representations (ICLR)}, 2024.

\end{thebibliography}
}
\clearpage
\setcounter{figure}{0}
\renewcommand{\thefigure}{S\arabic{figure}}
\renewcommand{\thetable}{S\arabic{table}}
\section*{Appendix}
\appendix
\section{Overview}
This appendix includes additional details
on benchmark construction, evaluator implementation, and experiment results, which cannot be fully covered in the main paper due to limited space. In \cref{sec:appendix_benchmark}, we first introduce more details about the benchmark construction, including the definition of eight prompt categories and the instruction used in GPT-4. Then, we present more implementation details about the proposed evaluator, including network details and training strategy in \cref{sec:appendix_evaluator}. Finally, we provide additional experimental results to demonstrate the effectiveness of our metric in \cref{sec:appendix_experiment}.

\section{More Details on Benchmark Construction}\label{sec:appendix_benchmark}
\subsection{Definition of Different Prompt Categories}

Considering the number of objects and their complexity, creativity, and relationship, we define eight prompt categories in our benchmark. The definitions of these categories are reported as follows:

\begin{itemize}
    \item \textit{Basic.} Descriptions about a single object without detailed geometry or appearance details, where some global properties (\textit{e.g.}, color, shape, material) could be included. Examples: ``A green apple", ``A square table", and ``A wooden chair".
    \item \textit{Refined.} Descriptions about a single object with one simple specification of the geometry or appearance details. Examples: ``A yellow rubber duck has only one foot ", ``A plastic cup with a printed logo", and ``An imperial state crown of England".
    \item \textit{Complex.} Descriptions about a single object with more than two detailed specifications of geometry or appearance details. 
    Examples: ``A brown teddy bear, fur matted, one eye missing", ``A turtle standing on its hind legs, wearing a top hat and holding a cane", and ``An over-sized, porous, sphere-shaped birdcage, made of woven golden wires".
    \item \textit{Fantastical.} Descriptions about a single object with high creativities that are not, or are generally unlikely to be, found in the real world. Examples: ``A frog with a translucent skin displaying a mechanical heart beating", ``A white cat has three tails, made of white, purple and black crystals", and ``A tiger dressed as a doctor".
    \item \textit{Grouped.} Descriptions about multiple objects without interactions, where some global properties (\textit{e.g.}, color, shape, material) could be included. 
    Examples: ``A delicious hamburger and a green apple", ``A red pig and a huge drum", and ``A round table and a square chair". 
    \item \textit{Action.} Descriptions about multiple objects with interaction about action relations, where some geometry or appearance details could be included. Examples: ``A humanoid robot with a top hat is playing the cello", ``A black cat with white feet is sleeping peacefully beside a carved pumpkin", and ``A dog is eating a red apple with its tail raised".
    \item \textit{Spatial.} Descriptions about multiple objects with interactions about spatial relations, where some geometry or appearance details could be included. Examples: ``A red apple on a round ceramic plate", ``A strong football player wearing a number seven jersey next to a blue soccer", and ``A blue rose in a crystal, symmetrical vase".
    \item \textit{Imaginative.} Descriptions about multiple objects with interactions, where objects or interactions are not or are generally unlikely to be found in the real world. Examples: ``A panda with a wizard hat is reading a newspaper", ``A ghost is eating a golden apple", and ``A raccoon astronaut is holding his helmet".
\end{itemize}

\subsection{Prompt Generation}

To ensure the comprehensiveness and diversity of prompts generated by GPT-4 \cite{openaigpt}, we define four aspects to consider during the prompt generation: object categories, geometry properties, appearance properties, and object interactions. These aspects are detailed as follows:

\begin{itemize}
    \item \textbf{Object Categories.} To ensure that the generated prompts encompass the most common object classes relevant to text-to-3D tasks, we first establish clear definitions for object categories, such as ``live beings'', ``animals'', ``plants'' and so on. This approach ensures prompt diversity and avoids the generated prompts being limited to a single category. 
    
    \item \textbf{Geometry Properties.} To generate more vivid and realistic 3D objects, it is essential to include detailed descriptions of geometry properties in prompts. Objects can be characterized based on their volume, shape, or size, including terms such as ``symmetrical", ``cylinder", and ``small".

    \item \textbf{Appearance Properties.} Similar to geometry properties, appearance properties are essential for prompt generation. Common attributes include colors, which are frequently used to describe objects' appearance, and textures, which capture surface characteristics such as ``smooth", ``rough", and ``furry". Additionally, materials, such as ``metal", ``glass", and ``fabric", also play a significant role in defining appearance.

    \item \textbf{Object Interactions.} Object interactions can be classified into two types: spatial and action relationships. Spatial relationships utilize terms such as ``on" and ``below" to accurately define the positional relationships between objects. Action relationships describe the actions or behaviors of objects, including verbs like ``wear", ``watch", and ``hold". By integrating these words with different object categories, GPT-4 can generate prompts encompassing various interactions.
\end{itemize}

\begin{figure*}[t]
    \centering
    \includegraphics[width=\linewidth]{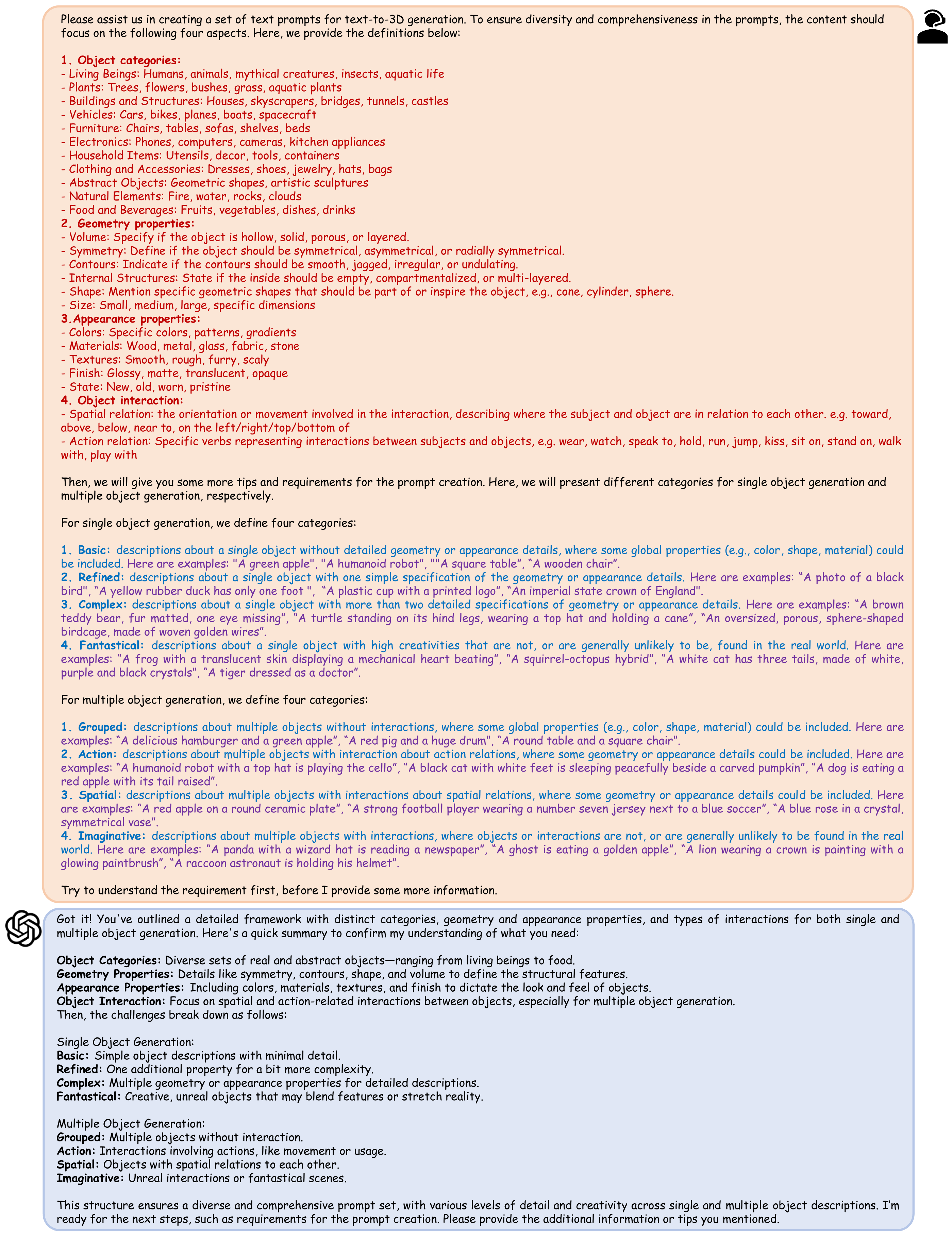}
   \caption{Examples of guiding GPT-4 for automatic prompt generation. (Part 1/2)}
   \label{fig:gpt1}
\end{figure*}

\begin{figure*}[t]
    \centering
    \includegraphics[width=\linewidth]{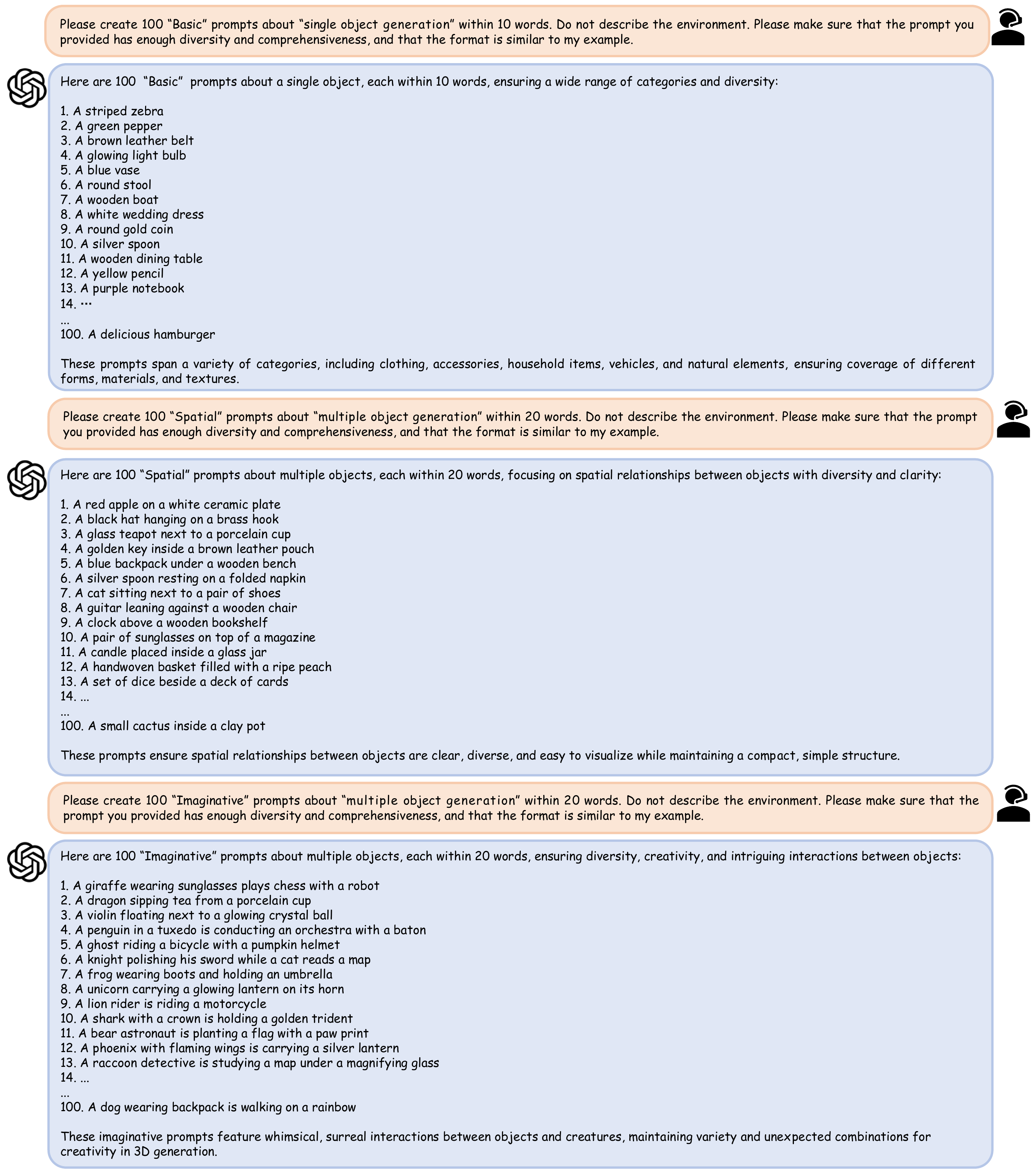}
   \caption{Examples of guiding GPT-4 for automatic prompt generation. (Part 2/2)}
   \label{fig:gpt2}
\end{figure*}

After defining eight prompt categories and four aspects, we design a universal template to provide GPT-4 with essential context for the prompt generation task. Detailed instructions are shown in \cref{fig:gpt1} and \cref{fig:gpt2}. Following these instructions, GPT-4 can gain a foundational understanding of the task and generate a list of prompts based on evaluator input. By editing prompt categories and lengths, GPT-4 can efficiently produce prompts across different categories.

\begin{figure}[t]
    \centering
    \begin{subfigure}[t]{0.48\textwidth}
        \centering
        \includegraphics[width=\linewidth]{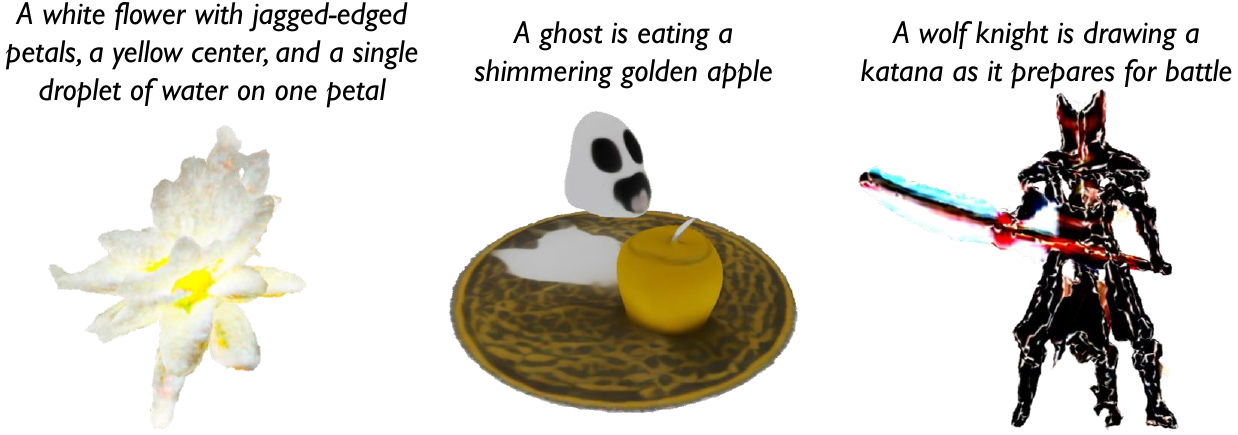}
        \label{fig:training1}
    \end{subfigure}  
    \vspace{0.1cm}
    \begin{subfigure}[t]{0.48\textwidth}
        \centering
        \includegraphics[width=\linewidth]{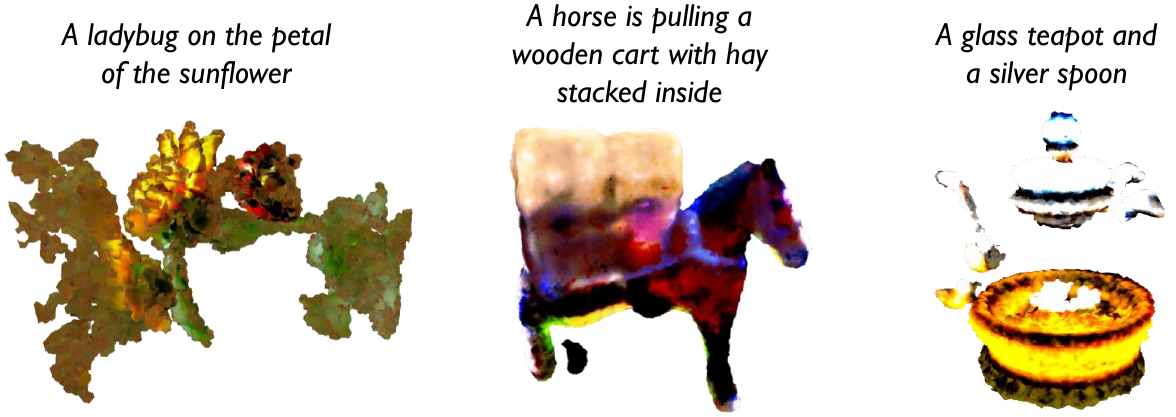}
        \label{fig:training2}
    \end{subfigure}
    \begin{subfigure}[t]{0.48\textwidth}
        \centering
        \includegraphics[width=\linewidth]{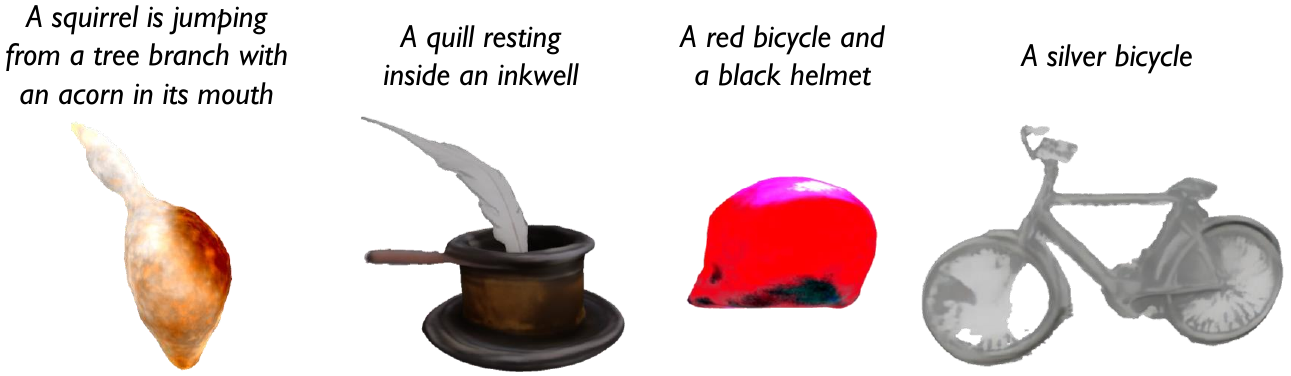}
        \label{fig:training3}
    \end{subfigure}
   \caption{Training samples used in our subjective experiment.}
   \label{fig:training}
   
\end{figure}

\subsection{Mesh Generation and Visualization }

In recent years, significant advancements have been made in 3D generative methods. In this paper, we employ DreamFusion \cite{dreamfusion}, Magic3D \cite{magic3d}, SJC \cite{sjc}, TextMesh \cite{textmesh}, 3DTopia \cite{3dtopia}, Consistent3D \cite{consistent3d}, LatentNeRF \cite{latentnerf}, and One-2-3-45++ \cite{one2345++} to generate 3D textured meshes. Here, we provide a brief introduction of each method. 
\begin{itemize}
    \item \textbf{DreamFusion} pioneers the paradigm of optimizing a unique 3D representation per text input or image, guided by powerful pre-trained 2D diffusion models.
    \item \textbf{Magic3D} introduces a coarse-to-fine optimization strategy with two stages. This approach improves both speed and quality. 
    \item \textbf{SJC} converts a pre-trained 2D diffusion generative model on images into a 3D generative model of radiance fields, without requiring access to any 3D data. 
    \item \textbf{TextMesh} employs a novel way to fine-tune the mesh texture, removing the effect of high saturation and improving the details of the output textured mesh. 
    \item \textbf{3DTopia} is a two-stage text-to-3D generation method. The first stage uses a diffusion model to quickly generate candidates. The second stage refines the assets chosen from the first stage. It can generate high-quality general 3D assets within 5 minutes using hybrid diffusion priors. 
    \item \textbf{Consistent3D} uses deterministic sampling priority to ensure that different generated results are visually more consistent and have higher details when generating 3D objects. Compared to traditional random sampling methods, this approach reduces variability in generated results, ensuring that the produced 3D objects align more consistently with the specified text prompt. 
    \item \textbf{LatentNeRF} guides the 3D generation process by encoding shape information in latent space. In this way, it can more effectively generate high-quality textures and shapes that match specific shapes while preserving details.
    \item \textbf{One-2-3-45++} advances multi-view 3D generation via an enhanced Zero123 \cite{liu2023zero1to3} module enabling simultaneous cross-view attention, alongside a multi-view conditioned 3D diffusion module performing coarse-to-fine textured mesh prediction over time.
\end{itemize}

\begin{figure*}[t]
    \centering
    \includegraphics[width=0.92\linewidth]{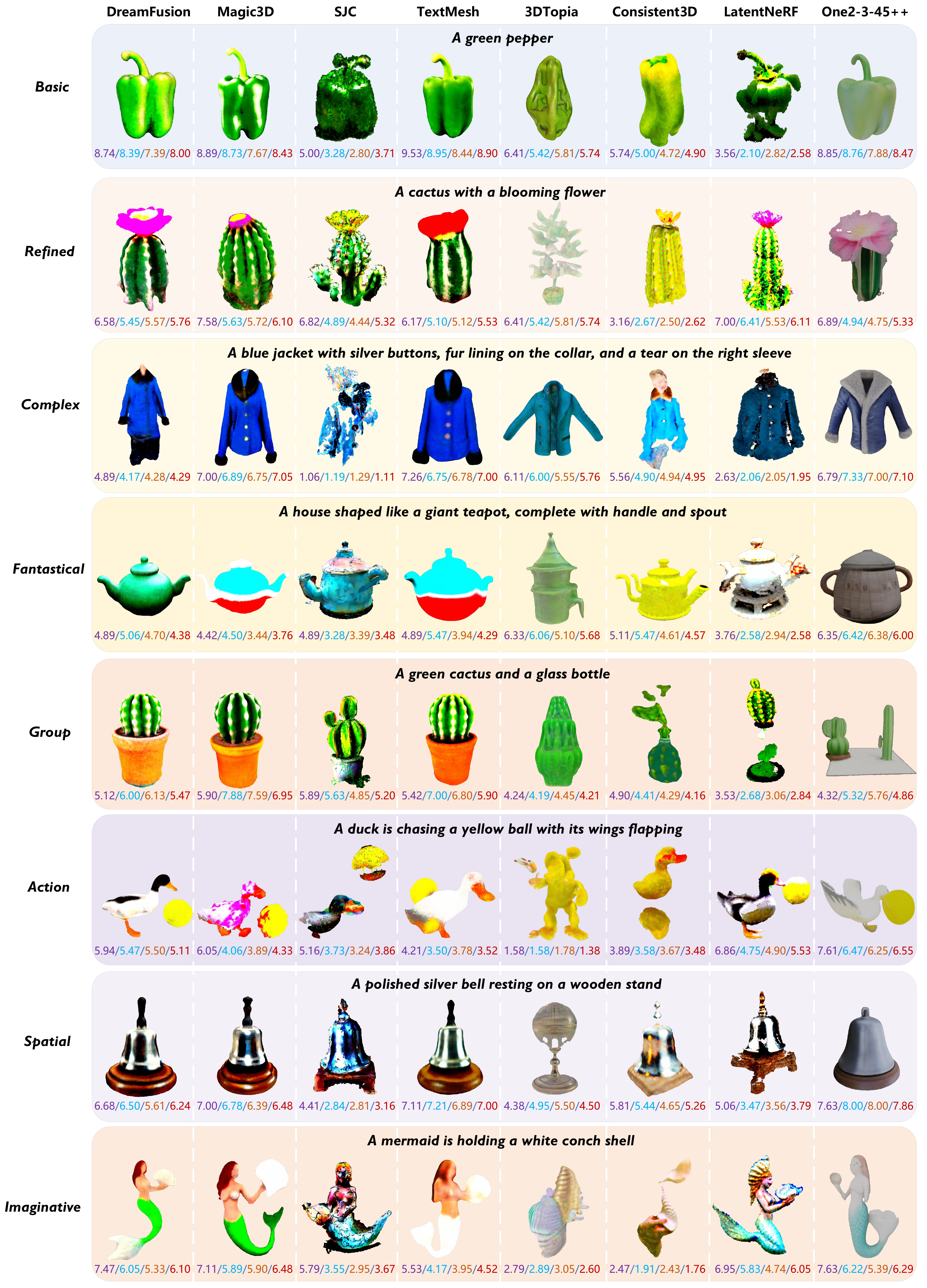}
    \caption{More results of eight generative methods for eight categories. The scores below each sample denote alignment, geometry, texture, and overall quality, respectively.}
    \label{fig:appendix_sample}
\end{figure*}

We present additional generated samples along with the evaluation scores for each method in \cref{fig:appendix_sample}.

\begin{table*}[t]
\centering
\caption{Comparison of the existing text-to-3D datasets. `\XSolidBrush' represents the scores are not available.}
\label{tab:dataset_comparison}
\resizebox{\textwidth}{!}{
\begin{tabular}{c|c|c|c|c|c|c}
\toprule
Benchmark & Prompt Categories & \begin{tabular}[c]{@{}c@{}}Number of\\Generative Methods\end{tabular} & Rating Dimensions & \begin{tabular}[c]{@{}c@{}}Number of\\Annotated Samples\end{tabular} & \begin{tabular}[c]{@{}c@{}}Number of\\Rating Score\end{tabular} & Annotation Type\\ \midrule

$\mathrm{T}^3$Bench \cite{he2023t3bench}& \begin{tabular}[c]{@{}c@{}c@{}}Single Object,\\Single Object with Surroundings,\\Multiple Objects\end{tabular} & 7 & \begin{tabular}[c]{@{}c@{}}Alignment,\\ Quality\end{tabular} & 630 & 630 $\times$ 2 $\times$ unknown & \begin{tabular}[c]{@{}c@{}}\XSolidBrush\\(Absolute Score)\end{tabular}\\ \midrule

GPTEval3D \cite{wu2024gpt4v} & \begin{tabular}[c]{@{}c@{}}Creativity,\\Complexity\end{tabular} & 13 & \begin{tabular}[c]{@{}c@{}c@{}c@{}c@{}}Text-Asset Alignment,\\3D Plausibility,\\Texture details,\\Geometry details,\\Texture–geometry coherency\end{tabular} & 234 pairs & 234 $\times$ 5 $\times$ 3 & \begin{tabular}[c]{@{}c@{}}\XSolidBrush\\(Preference Score)\end{tabular}\\ \midrule

MATE-3D & \begin{tabular}[c]{@{}c@{}c@{}c@{}}Basic,\ Refined,\\Complex,\ Fantastical,\\Grouped,\ Action,\\Spatial,\ Imaginative\end{tabular} & 8 & \begin{tabular}[c]{@{}c@{}c@{}c@{}}Alignment,\\Geometry,\\Texture,\\Overall\end{tabular} & 1,280 & 1,280 $\times$ 4 $\times$ 21 & \begin{tabular}[c]{@{}c@{}}\Checkmark\\(Absolute Score)\end{tabular}\\ \bottomrule
\end{tabular}}
\end{table*}

\begin{figure}[t]
    \centering
    \includegraphics[width=0.85\linewidth]{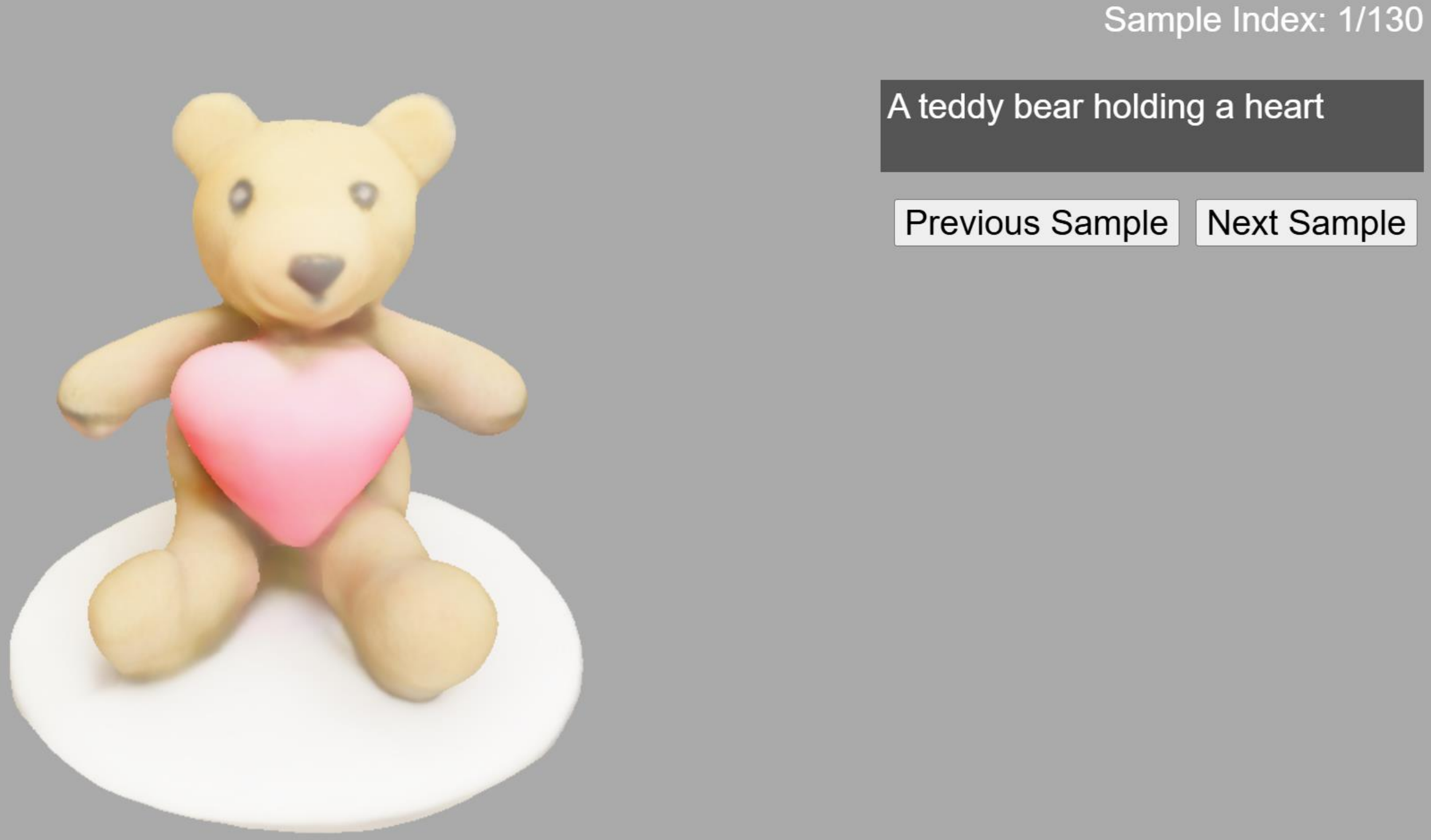}
   \caption{Illustration of the environment platform.}
   \label{fig:experiment_environment}
\end{figure}

\begin{table}[]
  \centering
  \caption{Performance comparison with evaluators in GPTEval3D and $T^3$Bench.}
  \resizebox{\linewidth}{!}{
    \begin{tabular}{c|cccc}
    \hline
     KRCC & {Alignment} & {Geometry} & {Texture} & {Overall}    \\
     \hline
    HyperScore  &0.517 &0.572 &0.622 &0.603

 \\
    GPTEval3D   &0.355	&0.356	&0.432	&0.391 \\
    \hline
    \end{tabular}}
  \label{tab:ablation_regionnum}%

\resizebox{\linewidth}{!}{
    \begin{tabular}{c|cccc}
    \hline
     SRCC & {Alignment} & {Geometry} & {Texture} & {Overall}    \\
     \hline
    HyperScore  &0.739	&0.782	&0.811	&0.792
 \\
    $T^3$Bench-Quality   &0.619	&0.497	&0.494	&0.540 \\
    $T^3$Bench-Alighment   &0.407	&0.366	&0.376	&0.383 \\
    \hline
    \end{tabular}}
  \label{tab:comparion_with_benchmark}%

\end{table}%

\subsection{Subjective Experiment Procedure}

\textbf{Training Session.} Before the subjective experiment, we scale the generated meshes proportionally to ensure that each mesh fits within a cube defined by the range $\left [ -1,\ 1 \right ] $. To enhance the reliability of the subjective scores, we use 10 samples whose corresponding prompts are excluded from MATE-3D to train subjects, helping them understand the rating rules. The training samples are selected to represent a comprehensive quality range, allowing viewers to understand the dataset comprehensively. We assign reference scores to the training samples and present the samples twice, requiring viewers to score them during the second viewing. If the scores given by the viewers show a high correlation with the reference scores, we conclude that they have grasped the rating principles and can provide reliable scores. Conversely, if viewers assign biased scores from the references, we repeat the training procedure until they provide reasonable results. The 10 training samples are illustrated in \cref{fig:training}.

\textbf{Experimental Environment.} To enable subjects to observe 3D objects from various viewpoints and provide more accurate ratings, we employ an interactive approach for the experiment. Our interactive renderer is developed as a web application utilizing the Three.js library \cite{threejs}. To effectively capture the scene, we employ an orthographic camera, with its field of view defined by the camera frustum. Subjects can control the camera's orientation through mouse movements, which enables them to adjust their viewpoints dynamically. To minimize the influence of the background on the subjects' evaluation, we set the background color to gray, specifically implemented using the code $scene.background = new THREE.Color(0xaaaaaa)$. The platform, as illustrated in \cref{fig:experiment_environment}, allows subjects to navigate between previous and next samples and rate scores from four dimensions. The subjective experiment is conducted on 27-inch AOC Q2790PQ monitors with a resolution of 2560×1440 in an indoor laboratory environment under standard lighting conditions.

\textbf{Outlier Detection.} In total, we generate 1,280 textured meshes from 160 prompts with eight generative methods. To mitigate visual fatigue associated with prolonged experiment durations, we randomly divide the 1,280 samples into 10 sessions. To detect outliers during the subjective experiment, each rating session includes one extremely low-quality sample and one duplicate sample as ``trapping samples". Consequently, each session comprises 130 samples. After collecting the subjective scores, we implement two consecutive steps to identify outliers from the raw data. First, we identify outliers based on the ``trapping samples" results. If a subject provides a high score that exceeds the expected threshold for the extremely low-quality sample or provides significantly different scores for the two duplicate samples, we exclude the raw scores of the subject from our analysis. Second, we apply the outlier detection method described in ITU-R BT.500 \cite{bt500} to conduct a further examination, removing any additional outliers identified through this process. As a result, four outliers are identified and eliminated from the subjective scores. Finally, we collect 17 scores for each sample.

\subsection{Comparison with Other Benchmarks}
Previous works \cite{he2023t3bench, wu2024gpt4v} have presented some text-to-3D benchmarks. We compare MATE-3D with the existing benchmarks from multiple perspectives in \cref{tab:dataset_comparison}. Note that for the column of ``Number of Annotated Samples", $\mathrm{T}^3$Bench and MATE-3D provide ``Absolute Score" for each sample, whereas GPTEval3D creates pairs of two samples and assigns ranking orders, referred to as ``Preference Score" within each pair. The column of ``Number of Rating Scores" is calculated as ``Number of Annotated Samples" $\times$ ``Number of Rating Dimensions" $\times$ ``Subject Number for Each Sample". Additionally, since the scores for $\mathrm{T}^3$Bench and GPTEval3D are not publicly available, we cannot evaluate the performance of objective metrics on these two benchmarks.

From \cref{tab:dataset_comparison}, it is clear that our benchmark presents several noticeable advantages. First, we incorporate a broader range of prompt categories, ensuring diversity and representativeness in the prompt generation. Second, we introduce four rating dimensions, offering a more comprehensive assessment of sample quality compared to $\mathrm{T}^3$Bench. Although GPTEval3D employs more dimensions, it only provides preferences for a limited number of pairs. Third, to promote robust and unbiased scoring, we annotate all generated samples and recruit 21 subjects to rate 1,280 samples across four dimensions. Additionally, we apply comprehensive outlier detection to refine the MOS, reducing the impact of anomalous ratings. We will make the MOS publicly accessible, facilitating further validation and experimentation by other researchers.

\begin{figure*}[t]
    \centering
    \begin{subfigure}[t]{0.24\textwidth}
        \centering
        \includegraphics[width=\linewidth]{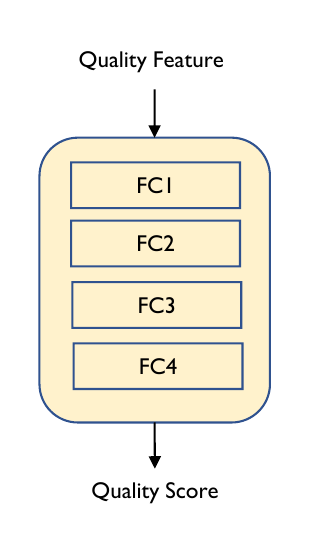}
        \caption{}
        \label{fig:network_detail1}
    \end{subfigure} 
    \begin{subfigure}[t]{0.56\textwidth}
        \centering
        \includegraphics[width=\linewidth]{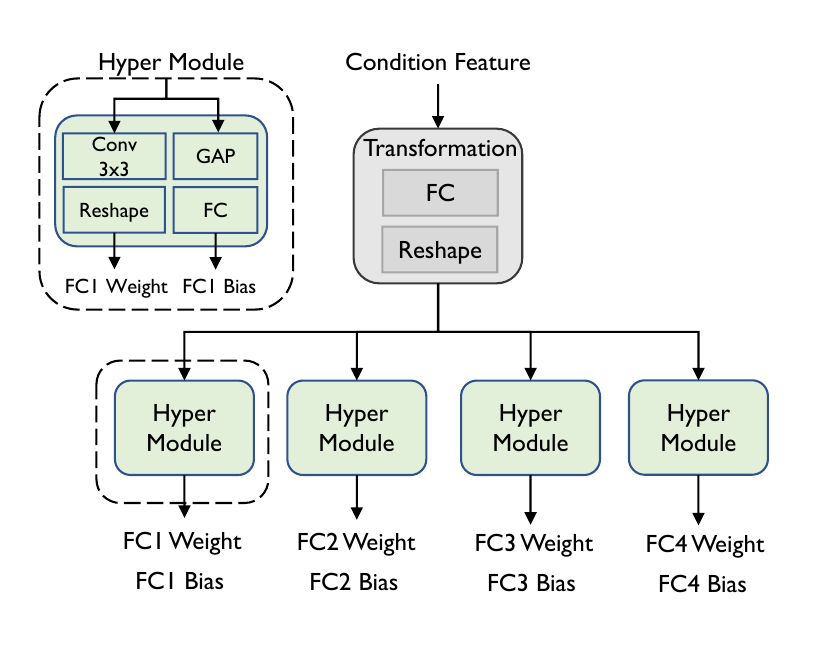}
        \caption{}
        \label{fig:network_detail2}
    \end{subfigure}

    \caption{(a) Architecture of the mapping head; (b) Architecture of the hypernetwork.}
    \label{fig:network_detail}
\end{figure*}

We further compare the evaluators proposed in GPTEval3D and T3Bench with HyperScore.  i) \textbf{Comparion with GPTEval3D}: The evaluator GPTEval3D is based on GPT-4v, which can only rank generated results from the same prompt in terms of five dimensions. However, directly ranking two samples in these methods can only know which one is better, but not how much better. In comparison, the proposed evaluator provides absolute scores for different samples, which can not only compare samples generated from different prompts but also know how much the quality difference between the two samples is. We report the KRCC results of HyperScore and GPTEval3D on MATE-3D in the first table of \cref{tab:comparion_with_benchmark}, we can see HyperScore perform better. ii) \textbf{Comparion with $T^3$Bench}: $T^3$Bench proposes two separate evaluators for quality and alignment measurement, which cannot handle multi-dimensional quality assessment.  We report the SRCC results of HyperScore and two evaluators in $T^3$Bench on MATE-3D in the second table of \cref{tab:comparion_with_benchmark}, we can see HyperScore can evaluate different dimensions more effectively than two evaluators in $T^3$Bench.

\section{More Details on Evaluator Implementation} \label{sec:appendix_evaluator}

\subsection{Network Details}

\textbf{Utilization of Pre-trained Model.} The used visual encoder in HyperScore is Vision Transformer~\cite{dosovitskiy2020image} with 16 × 16 patch embeddings (namely ViT-B/16) in CLIP-Visual. The textual encoder is also the pre-trained transformer in CLIP-Textual. The visual, textual, and condition features all have a size of $D=512$ while the quality feature has a size of $D_q=224$.

\textbf{Architecture of HyperNetwork and Mapping Head.} In the evaluator, we use a hypernetwork $\pi\left ( \cdot  \right ) $ to generate the parameter weights for a mapping head $\psi\left ( \cdot  \right ) $. We illustrate the network structure of  $\pi\left ( \cdot  \right ) $ and $\psi\left ( \cdot  \right ) $ in \cref{fig:network_detail}. Seeing \cref{fig:network_detail1}, the mapping head contains four fully connected (denoted by FC1-FC4) layers and takes the quality feature as input to generate the quality score. The input sizes of four FC layers are 224, 112, 56, and 28 as shown in \cref{tab:feature_size}. To inject the dimension-related information into the mapping head, we use the hypernetwork shown in \cref{fig:network_detail2} to generate the weights and biases for FC layers. More specifically, the condition feature after transformation is fed into four independent hyper modules, and each hyper module will output the weight and bias for the corresponding FC layer in the mapping head. We also declare the feature size for the weight and bias generated of the FC1 layer in \cref{tab:feature_size}.

\begin{figure*}
    \centering
    \includegraphics[width=\linewidth]{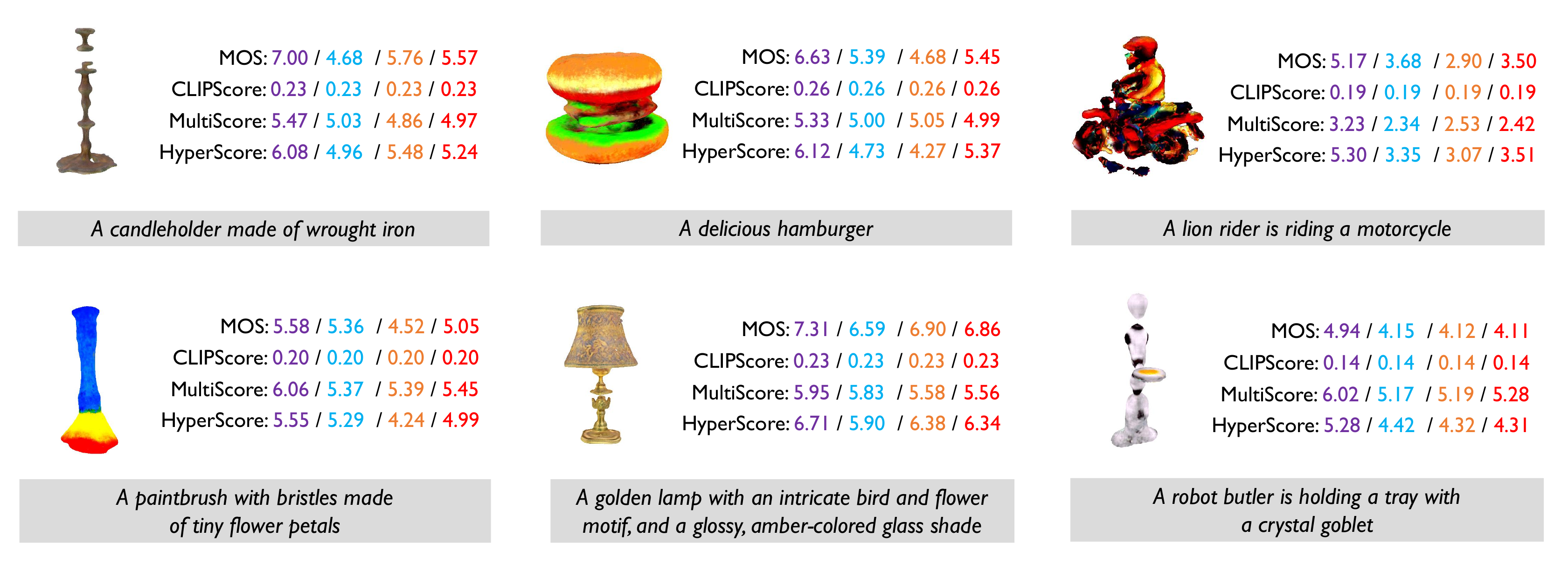}
    \caption{Additional exemplary samples with their MOSs and the predicted scores of different metrics. The four scores in each row denote alignment, geometry, texture, and overall quality, respectively.}
    \label{fig:qualitative_sample_appendix}
    \vspace{-0.3cm}
\end{figure*}

\subsection{Loss Function}
Given $K$ evaluation dimensions, we finally obtain $\left \{ \hat{q}_i \right \}_{i=1}^K$ as predictions. Denoting the subjective MOS as $\left \{ {q}_i \right \}_{i=1}^K$, we first define a regression loss as:
\begin{equation}
    L_{reg} = \frac{1}{KB}\sum_{b=1}^B\sum_{i=1}^K\left ( \hat{q}_i^b-q_i^b \right ) ^2,
\end{equation}
where the superscript ``$b$" denotes the $b$-th samples in a mini-batch with the size of $B$. 

Considering that different evaluation dimensions have various focuses, to 
avoid learning homogeneous features,
we define a feature disentangling loss for the condition features $\left \{ f_c^i \right \}_{i=1}^K$:
\begin{equation}
     L_{dis} = \frac{2}{K\left ( K-1 \right )}\sum_{i\neq j}\max\left ( \epsilon,\mathrm{cos}\left ( f_c^i,f_c^j \right ) \right ) ,
\end{equation}
where $\mathrm{cos}\left ( \cdot \right )$ denotes the cosine similarity between two features; $\epsilon$ represents the margin that controls the divergence among features and we set $\epsilon=0$ in the implementation. By introducing $L_{dis}$, we can increase the discrepancy among different evaluation dimensions during the network training.

Finally, the overall loss function for training is defined as: 

\begin{equation}
    L = L_{reg} + \lambda L_{dis} ,
\end{equation}
where $\lambda$ is the weighting factor, we simply set $\lambda=1$.

\begin{table}[]
    \centering
    \caption{The feature size of each module in the hypernetwork and mapping head.}
    \resizebox{\linewidth}{!}{
    \begin{tabular}{c c c c}
    \toprule
         Module & Layer & Input Size & Output Size \\ \midrule
         
            \multirow{4}{*}[-2.5ex]{Mapping Head} 
        & FC1 Layer & 224 & 112 \\ \rule{0pt}{15pt}
        & FC2 Layer & 112 & 56 \\ \rule{0pt}{15pt}
        & FC3 Layer & 56 & 28 \\ \rule{0pt}{15pt}
        & FC4 Layer & 28 & 1 \\
        \midrule
        \midrule
         
         \multirow{2}{*}{Transformation} &FC Layer &512 & 5,488  \\ \rule{0pt}{15pt}
         &Reshape &5,488 &$112\times7\times7$ \\

         \midrule
         \multirow{2}{*}{FC1 Weight Generation} & $3\times3$ Conv &$112\times7\times7$ &$512\times7\times7$  \\  \rule{0pt}{15pt}
         & Reshape &$512\times7\times7$ &$224\times112$  \\  \midrule
        \multirow{2}{*}{FC1 Bias Generation} 
        & Global Avg Pool &$112\times7\times7$ &$112\times1\times1$  \\ \rule{0pt}{15pt}
         & FC Layer &112 &112
         
         \\

    \bottomrule
    \end{tabular}}
    \label{tab:feature_size}
\end{table}

\subsection{Training Strategy}

We train the proposed HyperScore on MATE-3D for 30 epochs with a batch size of 8. During the training and testing process, all rendered images are resized into the resolution of $224\times224$. We use the Adam \cite{kingma2014adam} optimizer with weight decay $1e-4$. The learning rate is set separately as $2e-6$ and $2e-4$ for the pre-trained visual encoder and other parts (note that the textual encoder is frozen) and is reduced by a rate of 0.9 every 5 epochs.

\begin{figure*}
    \centering
    \includegraphics[width=0.95\linewidth]{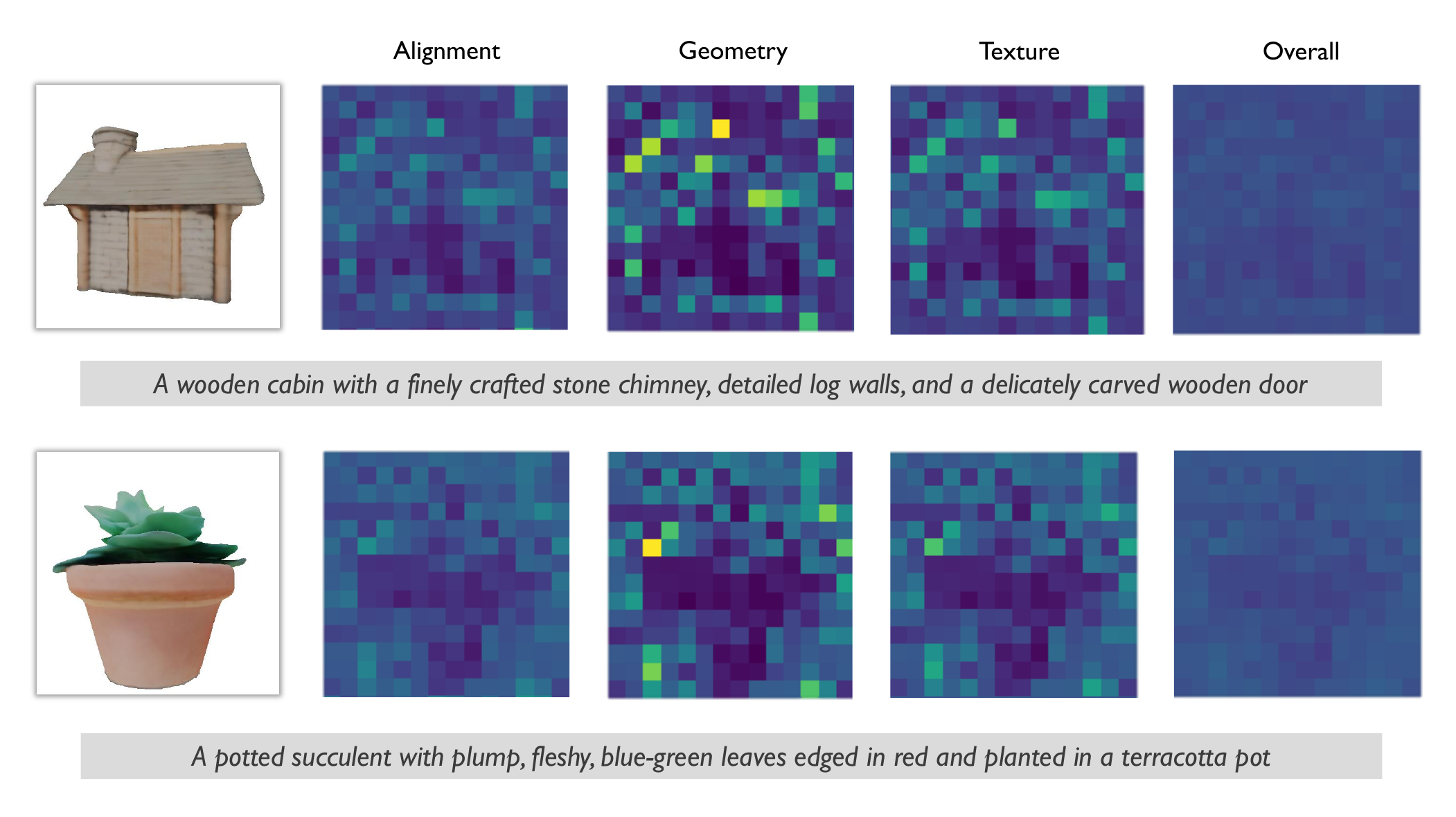}
    \caption{Visualization for the weight maps used for conditional feature fusion. }
    \label{fig:weight_visualization}
    \vspace{-0.3cm}
\end{figure*}

\section{Additional Experimental Results} \label{sec:appendix_experiment}

\subsection{Visualization}
We provide additional qualitative samples in \cref{fig:qualitative_sample_appendix} for visualization. From the figure, we can observe that HyperScore presents more accurate predictions. In comparison, MultiScore usually assigns similar scores across different dimensions, making it challenging for users to determine which factor impacts visual perception more.

Furthermore, we visualize the weight maps used for conditional feature fusion in \cref{fig:weight_visualization}. From the figure, we can observe that the weight maps differ across evaluation dimensions, demonstrating that the condition features contribute to distinguishing these dimensions. Meanwhile, we notice that the weight maps for the geometry evaluation have larger element values at the edges of objects, which may help to seize the shapes better. In comparison, the overall evaluation exhibits a more uniform distribution of the weight maps, probably because it needs to consider various factors during the evaluation process.

\begin{figure}[t]

    \centering
  \includegraphics[width=0.8\linewidth]{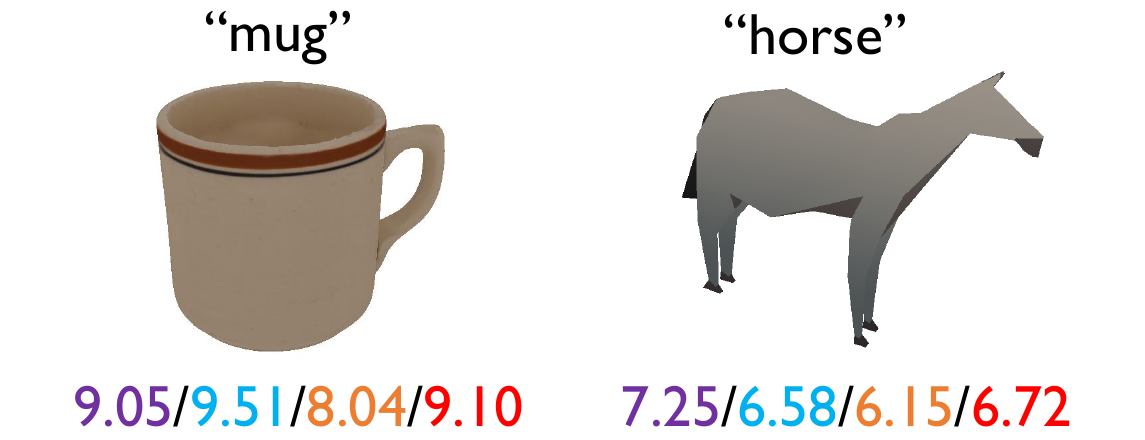}
  \caption{Evaluation results for examples in Objaverse.}
  \label{fig:sample_objaverse}
  
\end{figure}

To validate the evaluation ability for ground truth data, we download 2000 meshes from Objaverse \cite{deitke2023objaverse}and use HyperScore to access their quality. The average scores for four dimensions are [7.9, 7.6, 6.7, 7.4], which is generally higher than the 3D generation quality. \cref{fig:sample_objaverse} illustrates two evaluated examples. There also exist relatively low-quality samples in Objaverse such as ``horse''.

% Finally, we present a qualitative analysis using a prevalent tool (\textit{i.e.}, XGrad-CAM \cite{fu2020axiom}) for network explanation. As illustrated in \cref{fig:gradcam_visualization}, the XGrad-CAM maps reflect what makes the network perform decisions for different evaluation dimensions. We observe that the XGrad-CAM maps exhibit different focus areas corresponding to changes in evaluation dimensions, which validates the effectiveness of the proposed conditional learning strategy. Notably, the XGrad-CAM maps related to the geometry evaluation pay more attention to the shape and structure of objects (\textit{e.g.,} missing bones of the skeleton knight). In contrast, the texture and overall quality evaluations have a wider range of focus, likely because they require more appearance details to make decisions.

\subsection{Additional Ablation Studies}

\textbf{Ablation Study for the Aggregation Strategy.}
We test the performance of HyperScore with different aggregation strategies between the visual and textual features. Except for the used element-wise multiplication (denoted by $\odot$), we choose two other strategies, \textit{i.e.}, addition (denoted by $+$) and concatenation (denoted by $\oplus$). According to the results in \cref{tab:ablation_aggregation}, we can see that the element-wise multiplication performs best on all evaluation dimensions, and the addition and concatenation both provide relatively inferior performance, which justifies our choice for the aggregation strategy.

% \begin{figure*}
%     \centering
%     \includegraphics[width=0.9\linewidth]{image_appendix/gradcam_visualization.pdf}
%     \caption{Visual explanations generated by XGrad-CAM for different evaluation dimensions.}
%     \label{fig:gradcam_visualization}
%     \vspace{-0.3cm}
% \end{figure*}

\begin{table}[t]

  \centering
  \caption{Ablation study for the aggregation strategy between visual and textual features. Results of SRCC are reported.}

  \resizebox{\linewidth}{!}{
    \begin{tabular}{c|cccc}
    \toprule
     Aggregation  & Alignment & Geometry & Texture & Overall\\
    \midrule
    $f_{v,c}^i+ f_t^{eot}$           &0.708	&0.769	&0.783	&0.767

 \\
      $f_{v,c}^i\oplus f_t^{eot}$           
    &0.717	&0.766	&0.789	&0.767

\\
    $f_{v,c}^i\odot f_t^{eot}$      &\bf{0.739}	&\bf{0.782}	&\bf{0.811}	&\bf{0.792}

 \\ 
    \bottomrule
    \end{tabular}}%
  \label{tab:ablation_aggregation}%
  
\end{table}%

\textbf{Ablation Study for the Viewpoint Count.} To perform evaluation, we render textured meshes into $M=6$ images from six perpendicular viewpoints (\textit{i.e.}, along the positive and negative directions of the x, y, and z axes).
We further test the performance of HyperScore under different $M$ values to investigate the influence of the number of viewpoints and report the results in \cref{tab:ablation_viewpoint}. Note that the corresponding camera locations of different $M$ are declared in \cref{tab:camera_location}.
From \cref{tab:ablation_viewpoint}, it can be observed that as the $M$ value increases, the performance initially improves and then decreases. The reason may be that increasing $M$ provides more information for a better prediction when $M$ is a small number. However, when $M$ becomes large, the increase in $M$ can affect performance to some extent due to information redundancy. Meanwhile, a larger $M$ also incurs higher computational complexity. Therefore, to achieve the balance between performance and complexity, we consider $M=6$ as a suitable choice.

\begin{figure}
    \centering
    \includegraphics[width=\linewidth]{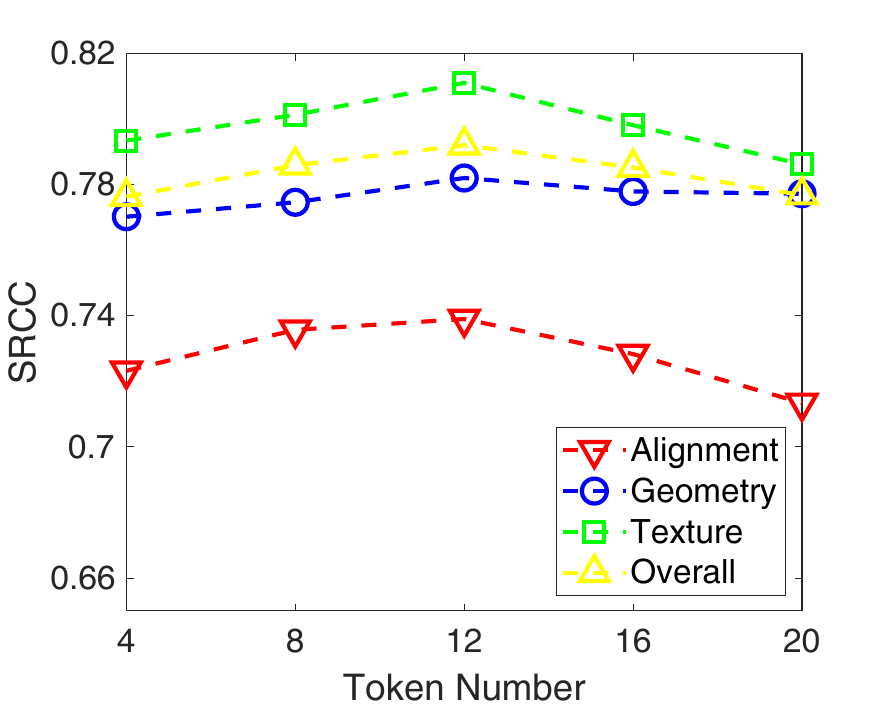}
    \caption{Performance comparison of the learnable prompt length.}
    \label{fig:plot_token_num}
    
\end{figure}

\textbf{Ablation Study for the Prompt Design.}
In our implementation, to obtain multiple condition features, we first transform the meta texts (\textit{i.e.,} \textit{``alignment quality'', ``geometry quality'', ``texture quality'', ``overall quality''}) into text tokens and then insert them into the front of $K$ learnable prompts with $L=12$ tokens. Here we further explore the impact of $L$ on the performance and illustrate the results in \cref{fig:plot_token_num}. From the figure, we see that HyperScore performs the best with $L=12$. Short prompts may constrain the learning space, potentially leading to the loss of crucial information. Conversely, too long prompts might introduce unnecessary noise, which may obscure important information and degrade performance. Therefore, balancing the prompt length is critical to optimizing the metric performance.

\begin{table}[t]

  \centering
  \caption{Ablation study for the number of rendered viewpoints. Results of SRCC are reported.}

  \resizebox{0.9\linewidth}{!}{
    \begin{tabular}{c|cccc}
    \toprule
     $M$  & Alignment & Geometry & Texture & Overall\\
    \midrule
    4   &0.716	&0.755	&0.790	&0.766

\\
    6  &\bf{0.739}	&0.782	&\bf{0.811}	&\bf{0.792}

\\
    9  &0.737	&0.780	&0.810	&0.790

\\
    12  &0.724	&\bf{0.784}	&0.805	&0.784

\\
    16 &0.724	&0.784	&0.805	&0.784

\\
    \bottomrule
    \end{tabular}}%
  \label{tab:ablation_viewpoint}%
\vspace{-0.3cm}
\end{table}%

\begin{table}[]
\caption{Camera locations of different view counts.}
\label{tab:camera_location}
\resizebox{\linewidth}{!}{
\begin{tabular}{c|c|c}
\toprule
M & Elevation Angle & Azimuth Angle \\ \midrule
4 &  $-60^{\circ},\ 60^{\circ}$ & $0^{\circ},\ 180^{\circ}$ \\ 
9 &  $-60^{\circ},\ 0^{\circ},\ 60^{\circ}$ & $0^{\circ},\ 120^{\circ},\ 240^{\circ}$ \\ 
12 &  $-60^{\circ},\ 0^{\circ},\ 60^{\circ}$ & $0^{\circ},\ 90^{\circ},\ 180^{\circ},\ 270^{\circ}$ \\ 
16 &  $-60^{\circ},\ -30^{\circ},\ 30^{\circ},\ 60^{\circ}$ & $0^{\circ},\ 90^{\circ},\ 180^{\circ},\ 270^{\circ}$ \\ 

\midrule

6 & \multicolumn{2}{c}{$\left ( 0^{\circ},0^{\circ} \right ),\ \left ( 0^{\circ},90^{\circ} \right ),\ \left ( 0^{\circ},180^{\circ} \right ),\ \left ( 0^{\circ},270^{\circ} \right ),\ \left ( 90^{\circ},0^{\circ} \right ),\ \left ( -90^{\circ},0^{\circ} \right )$}
\\ \bottomrule

\end{tabular}}
\end{table}

We further replace these learnable prompts with fixed tags and test the performance of the case. Specifically, we define the tags for the four quality dimensions as:
\begin{itemize}
    \item \textbf{Alignment}: \textit{``quantity,  attributes, position, location''};
    \item \textbf{Geometry}: \textit{``shape, size, hole, edge, surface''};
    \item \textbf{Texture}: \textit{``color, material, clarity, texture, contrast''};
    \item \textbf{Overall}: \textit{``quantity, attributes, position, location, shape, size, hole, edge, surface, color, material, clarity, texture, contrast''}.
\end{itemize}
The performance of the fixed tags is reported in \cref{tab:ablation_prompt}, where we also report the performance of only using learnable tokens without the meta texts (denoted by ``Learnable \textit{w/o} meta"). We can observe that the learnable prompts with the meta texts outperform the fixed tags and alleviate the need for meticulous prompt design. Meanwhile, introducing the meta texts also benefits network optimization.

\textbf{Ablation Study for the Loss Function.}
The proposed network is trained using the regression loss $L_{reg}$, and the feature disentangling loss $L_{dis}$. We evaluate the effect of each loss function and report the results in Table \ref{tab:ablation_loss}. From the table, we can see that only utilizing $L_{reg}$ can also achieve remarkable performance while $L_{dis}$ benefits the evaluations of all dimensions. Considering $L_{dis}$ only measures the linear similarity between features, one possible avenue for further improvement is to minimize non-linear dependence between two features, such as mutual information.

\textbf{Ablation Study for the Input Image Type.}
The proposed network only uses texture rendering as input. We further test the impact of normal maps on model performance and reported the SRCC results in \cref{tab:ablation_modality}.  It seems that feeding the normal maps into our model does not bring much improvement even for geometry evaluation, which is consistent with the conclusion in GPTEval3D. The reason for this may be that the backbones are trained on natural images rather than normal maps, so simply merging normal information impairs the performance.

\textbf{Ablation Study for the Rotation Angle.}
A good 3D evaluation metric should be robust to changes of scale and viewpoint. 
HyperScore has scale invariance because the meshes are scaled into the unit ball before rendering. Meanwhile, HyperScore is robust to rotation because the training samples are in different poses. We report the performance under different azimuth angle in \cref{tab:ablation_angle}, where the results are relatively stable.

\begin{table}[t]

  \centering
  \caption{Ablation study for the prompt type. Results of SRCC are reported.}

  \resizebox{\linewidth}{!}{
    \begin{tabular}{c|cccc}
    \toprule
     Prompt Type  & Alignment & Geometry & Texture & Overall\\
    \midrule
    Fixed          &0.732	&0.778	&0.794	&0.782

    \\
        Learnable \textit{w/o} meta  &0.725	&0.772	&0.802	&0.784
        
 \\
    Learnable           
&\bf{0.739}	&\bf{0.782}	&\bf{0.811}	&\bf{0.792}

 \\
    \bottomrule
    \end{tabular}}%
  \label{tab:ablation_prompt}%
  
\end{table}%

\begin{table}[t]

  \centering
  \caption{Ablation study for the loss function. Results of SRCC are reported.}

  \resizebox{\linewidth}{!}{
    \begin{tabular}{c|cccc}
    \toprule
     Loss  & Alignment & Geometry & Texture & Overall\\
    \midrule
    $L_{reg}$         &0.728	&0.769	&0.797	&0.782

    \\
    $L_{reg}+L_{dis}$           
&\bf{0.739}	&\bf{0.782}	&\bf{0.811}	&\bf{0.792}

 \\
    \bottomrule
    \end{tabular}}%
  \label{tab:ablation_loss}%
  
\end{table}%

\subsection{Performance Comparison on Different Prompt Categories}

In this section, we test the performance of different metrics on eight prompt categories and report the results in \cref{tab:challenge_performance}. For the metrics that need fine-tuning, we perform a leave-one-category-out evaluation, that is, testing on one category while training on the other seven categories, which can validate the generalization performance for unknown scenes.
From the table, we have the following observations: i) HyperScore achieves the best performance on the four dimensions of almost all prompt categories, demonstrating its sensitivity to fine-grained scenes. In contrast, although some metrics work well on partial categories (\textit{e.g.,} ImageReward on \textit{Basic} and ViT+FT on \textit{Spatial}), they may fail in other cases (\textit{e.g.,} ImageReward on \textit{Complex} and ViT+FT on \textit{Basic}).
ii) Almost all metrics perform better when evaluating single object generation than multiple object generation. This may be because these metrics do not model the specific relationships between objects well. iii) For the sub-categories of single and multiple object generation, most metrics perform best on \textit{Basic} and \textit{Grouped}, respectively. It is reasonable because the evaluation needs to incorporate more factors when measuring complex scenes and relationships, which presents challenges for the existing metrics.

\subsection{Performance Comparison on Different Generative Methods}

In this section, we test the performance of different metrics on eight generative methods and report the results in \cref{tab:method_performance}. For the metrics that require fine-tuning, we perform a leave-one-method-out evaluation, that is, testing on one generative method while training on other seven methods, which can validate the generalization performance for unknown generative methods.
From the table, we have the following observations: i) HyperScore achieves the best performance on the four dimensions of almost all generative methods, demonstrating its generalization capability to new generative methods. In contrast, other metrics show inconsistent results across different methods.
ii) Almost all fine-tuned metrics exhibit a noticeable decline in performance when evaluating the two recent methods, 3DTopia and One-2-3-45++. This may be because the two methods present different distortion patterns compared to other methods. More specifically, 3DTopia tends to generate a single object, leading to comparatively lower scores in multiple object generation. One-2-3-45++ tends to generate flatter geometric structures and blurry textures. These deformations, which are not seen during training, are very challenging for evaluation. In contrast, fine-tuned metrics work well on DreamFusion because many other methods inherit DreamFusion's paradigm and generate similar patterns. Therefore, it is important to consider the generalization towards new generative methods when designing text-to-3D evaluators.

\begin{table}[t]
  \caption{Ablation study for the input image types. Results of SRCC are reported.}
  \centering
  % \caption{Performance comparison with evaluators in GPTEval3D and $T^3$Bench.}
  \resizebox{0.9\linewidth}{!}{
    \begin{tabular}{c|cccc}
    \toprule
     SRCC & {Alignment} & {Geometry} & {Texture} & {Overall}    \\
    \midrule
    Texture &\bf{0.739}	&\bf{0.782}	&\bf{0.811	}&\bf{0.792}
 \\
    Normal &0.663	&0.740	&0.722	&0.732
\\
    Texture + Normal &0.684	&0.749	&0.777	&0.749
 \\
    \bottomrule
    \end{tabular}}
  \label{tab:ablation_modality}
\end{table}

\begin{table}[t]
 \caption{Ablation study for the rotation angles. Results of SRCC are reported.}
  \centering
  % \caption{Performance comparison with evaluators in GPTEval3D and $T^3$Bench.}
  \resizebox{\linewidth}{!}{
    \begin{tabular}{c|cccc}
    \toprule
     SRCC & {Alignment} & {Geometry} & {Texture} & {Overall}    \\
     \midrule
    - &\bf{0.739}	&0.782	&\bf{0.811}	&0.792

 \\
    $30^{\circ}$ &0.719	&0.783	&0.800	&0.789

\\
    $45^{\circ}$ &0.723	&\bf{0.789}	&0.802	&\bf{0.793}

\\
    $60^{\circ}$ &0.721	&0.787	&0.801	&0.791
 \\
    \bottomrule
    \end{tabular}}
  \label{tab:ablation_angle}%

\end{table}

\newpage

% \begin{table}[t]

%   \centering
%   \caption{Ablation study for the number of rendered viewpoint.}

%   \resizebox{0.9\linewidth}{!}{
%     \begin{tabular}{c|cccc}
%     \toprule
%      $\epsilon$  & Alignment & Geometry & Texture & Overall\\
%     \midrule
%     0   &0.688	&0.752	&0.777	&0.753

% \\
%     0.1  &0.725	&0.780	&0.800	&0.783

% \\
%     0.2  &0.724	&0.780	&0.804	&0.785

% \\
%     0.3  &0.733	&\bf{0.796}	&\bf{0.815}	&0.796

% \\
%     0.4 &\bf{0.736}	&0.792	&0.815	&\bf{0.798}

% \\
%     0.5 &\bf{0.736}	&0.792	&0.815	&\bf{0.798}
% \\ \midrule
%    \textit{w/o} $L_{dis}$ &\bf{0.736}	&0.792	&0.815	&\bf{0.798}
% \\
%     \bottomrule
%     \end{tabular}}%
%   \label{tab:ablation_viewpoint}%
%    \vspace{-0.4cm}
% \end{table}%

% \begin{table}[t]

%   \centering
%   \caption{Ablation study for the number of rendered viewpoint.}

%   \resizebox{0.9\linewidth}{!}{
%     \begin{tabular}{c|cccc}
%     \toprule
%      $L$  & Alignment & Geometry & Texture & Overall\\
%     \midrule
%     4   &0.688	&0.752	&0.777	&0.753

% \\
%     8 &0.725	&0.780	&0.800	&0.783

% \\
%     12  &0.724	&0.780	&0.804	&0.785

% \\
%     16  &0.733	&\bf{0.796}	&\bf{0.815}	&0.796

% \\
%     20 &\bf{0.736}	&0.792	&0.815	&\bf{0.798}

% \\
%   \midrule
%    Fixed Tags &\bf{0.736}	&0.792	&0.815	&\bf{0.798}
% \\
%     \bottomrule
%     \end{tabular}}%
%   \label{tab:ablation_viewpoint}%
%    \vspace{-0.4cm}
% \end{table}%

\begin{table*}[t]
\centering
\caption{Performance comparison (in terms of SRCC) of different evaluators on eight prompt categories. }
\label{tab:challenge_performance}
\resizebox{0.68\textwidth}{!}{
\begin{tabular}{c|cccccccc}
\toprule
\multirow{2}{*}{Metric} & \multicolumn{8}{c}{Alignment}  \\ \cline{2-9}
& Basic & Refined & Complex & Fantastic & Grouped & Action & Spatial & Imaginative \\ \midrule

CLIPScore &0.574	&0.592	&0.407	&0.421	&0.492	&0.434	&0.404	&0.591

\\
BLIPScore  &0.640	&0.520	&0.479	&0.423	&0.590	&0.476	&0.577	&0.540

\\
Aesthetic Score &0.221	&0.285	&0.283	&0.051	&0.053	&0.086	&0.013	&0.062

\\
ImageReward &0.775	&0.672	&0.544	&0.528	&0.631	&0.622	&0.626	&0.608

\\
DreamReward &0.671	&0.597	&0.427	&0.424	&0.483	&0.454	&0.448	&0.491

\\
HPS v2  &0.597	&0.348	&0.146	&0.388	&0.562	&0.250	&0.407	&0.518

\\
CLIP-IQA 	&0.058	&0.144	&0.082	&0.127	&0.042	&0.006	&0.086	&0.114
\\
Q-Align &0.349	&0.365	&0.326	&0.255	&0.224	&0.212	&0.023	&0.111

\\ 
 
ResNet50 + FT &0.625	&0.603	&0.582	&0.601	&0.548	&0.429	&0.562	&0.475

\\
ViT-B + FT  &0.552	&0.533	&0.622	&0.519	&0.609	&0.623	&\bf{0.656}	&0.483

\\
SwinT-B + FT &0.626	&0.541	&0.651	&0.536	&0.455	&0.508	&0.570	&0.556
\\
DINO v2 + FT &0.669	&0.649	&0.720	&0.665	&0.643	&0.651	&0.594	&0.598

\\
MultiScore &0.680	&\bf{0.693}	&0.710	&0.621	&0.655	&0.615	&0.625	&0.580

\\
HyperScore   &\bf{0.819}	&0.690	&\bf{0.733}	&\bf{0.705}	&\bf{0.701}	&\bf{0.681}	&0.637	&\bf{0.691}

\\

\bottomrule

\end{tabular}}

\resizebox{0.68\textwidth}{!}{
\begin{tabular}{c|cccccccc}
\toprule
\multirow{2}{*}{Metric} & \multicolumn{8}{c}{Geometry}  \\ \cline{2-9}
& Basic & Refined & Complex & Fantastic & Grouped & Action & Spatial & Imaginative \\ \midrule

CLIPScore &0.575	&0.570	&0.438	&0.404	&0.496	&0.503	&0.479	&0.602

\\
BLIPScore  &0.667	&0.491	&0.496	&0.398	&0.556	&0.510	&0.611	&0.574

\\
Aesthetic Score &0.244	&0.279	&0.268	&0.053	&0.164	&0.041	&0.088	&0.028

\\
ImageReward &0.790	&0.545	&0.558	&0.467	&0.532	&0.588	&0.585	&0.600

\\
DreamReward &0.691	&0.495	&0.473	&0.395	&0.465	&0.422	&0.469	&0.497

\\
HPS v2  &0.617	&0.355	&0.218	&0.364	&0.504	&0.334	&0.423	&0.544

\\
CLIP-IQA 	&0.107	&0.215	&0.178	&0.133	&0.092	&0.107	&0.199	&0.029
\\
Q-Align &0.423	&0.469	&0.439	&0.334	&0.444	&0.396	&0.111	&0.079

\\ 
 
ResNet50 + FT &0.693	&0.671	&0.712	&0.655	&0.641	&0.491	&0.638	&0.515

\\
ViT-B + FT  &0.692	&0.643	&0.752	&0.568	&0.704	&0.598	&0.700	&0.552

\\
SwinT-B + FT &0.720	&0.618	&0.738	&0.579	&0.558	&0.559	&0.637	&0.622

\\
DINO v2 + FT &0.81	&0.752	&0.815	&0.749	&0.741	&0.769	&0.623	&0.619

\\
MultiScore &0.733	&\bf{0.753}	&0.802	&0.670	&0.727	&0.625	&0.723	&0.634

\\
HyperScore   &\bf{0.862}	&0.745	&\bf{0.838}	&\bf{0.742}	&\bf{0.776}	&\bf{0.690}	&\bf{0.736}	&\bf{0.738}

\\

\bottomrule

\end{tabular}}

\resizebox{0.68\textwidth}{!}{
\begin{tabular}{c|cccccccc}
\toprule
\multirow{2}{*}{Metric} & \multicolumn{8}{c}{Texture}  \\ \cline{2-9}
& Basic & Refined & Complex & Fantastic & Grouped & Action & Spatial & Imaginative \\ \midrule

CLIPScore &0.620	&0.616	&0.472	&0.442	&0.514	&0.519	&0.528	&0.631

\\
BLIPScore  &0.707	&0.556	&0.586	&0.458	&0.543	&0.507	&0.665	&0.584

\\
Aesthetic Score &0.218	&0.310	&0.306	&0.108	&0.257	&0.053	&0.108	&0.030

\\
ImageReward &0.800	&0.597	&0.596	&0.510	&0.505	&0.606	&0.626	&0.618

\\
DreamReward &0.697	&0.529	&0.460	&0.446	&0.425	&0.449	&0.514	&0.523

\\
HPS v2  &0.620	&0.359	&0.217	&0.370	&0.508	&0.376	&0.464	&0.584

\\
CLIP-IQA 	&0.128	&0.228	&0.230	&0.215	&0.111	&0.165	&0.223	&0.023
\\
Q-Align &0.469	&0.532	&0.485	&0.425	&0.524	&0.474	&0.231	&0.211

\\ 
 
ResNet50 + FT &0.721	&0.703	&0.733	&0.658	&0.654	&0.686	&0.553	&0.477

\\
ViT-B + FT  &0.702	&0.682	&0.782	&0.619	&0.721	&0.692	&0.668	&0.603

\\
SwinT-B + FT &0.747	&0.617	&0.770	&0.598	&0.580	&0.624	&0.666	&0.652

\\
DINO v2 + FT &0.818	&0.765	&0.842	&0.729	&0.751	&\bf{0.777}	&0.692	&0.674

\\
MultiScore &0.776	&0.778	&0.843	&0.671	&0.735	&0.695	&\bf{0.747}	&0.691

\\
HyperScore   &\bf{0.880}	&\bf{0.792}	&\bf{0.852}	&\bf{0.745}	&\bf{0.807}	&{0.761}	&0.746	&\bf{0.732}

\\

\bottomrule

\end{tabular}}

\resizebox{0.68\textwidth}{!}{
\begin{tabular}{c|cccccccc}
\toprule
\multirow{2}{*}{Metric} & \multicolumn{8}{c}{Overall}  \\ \cline{2-9}
& Basic & Refined & Complex & Fantastic & Grouped & Action & Spatial & Imaginative \\ \midrule

CLIPScore &0.590	&0.587	&0.441	&0.433	&0.497	&0.519	&0.473	&0.621

\\
BLIPScore  &0.678	&0.508	&0.502	&0.433	&0.570	&0.507	&0.626	&0.579

\\
Aesthetic Score &0.244	&0.295	&0.305	&0.056	&0.166	&0.053	&0.059	&0.021

\\
ImageReward &0.793	&0.588	&0.572	&0.504	&0.569	&0.606	&0.616	&0.617

\\
DreamReward &0.689	&0.536	&0.467	&0.428	&0.468	&0.452	&0.488	&0.513

\\
HPS v2  &0.623	&0.347	&0.200	&0.380	&0.530	&0.376	&0.427	&0.537

\\
CLIP-IQA 	&0.096	&0.202	&0.163	&0.163	&0.064	&0.165	&0.155	&0.023
\\
Q-Align &0.424	&0.465	&0.439	&0.345	&0.406	&0.349	&0.131	&0.09

\\ 
 
ResNet50 + FT &0.690	&0.675	&0.695	&0.640	&0.611	&0.608	&0.508	&0.512

\\
ViT-B + FT  &0.670	&0.640	&0.739	&0.567	&0.664	&0.679	&0.613	&0.552

\\
SwinT-B + FT &0.710	&0.598	&0.732	&0.572	&0.519	&0.590	&0.620	&0.616
\\
DINO v2 + FT &0.786	&0.735	&0.816	&0.718	&0.704	&\bf{0.737}	&0.648	&0.622

\\
MultiScore &0.734	&0.748	&0.811	&0.665	&0.699	&0.656	&0.694	&0.643

\\
HyperScore   &\bf{0.865}	&\bf{0.760}	&\bf{0.841}	&\bf{0.750}	&\bf{0.759}	&{0.730}	&\bf{0.717}	&\bf{0.756}

\\

\bottomrule
\end{tabular}}

\end{table*}

\begin{table*}[t]
\centering
\caption{Performance comparison (in terms of SRCC) of different evaluators on eight generative methods.}
\label{tab:method_performance}
\resizebox{0.8\textwidth}{!}{
\begin{tabular}{c|cccccccc}
\toprule
\multirow{2}{*}{Metric} & \multicolumn{8}{c}{Alignment}  \\ \cline{2-9}
&DreamFusion &Magic3D &SJC	&TextMesh	&Consistent3D	&LatentNeRF	&3DTopia &One-2-3-45++
 \\ \midrule

CLIPScore &0.500	&0.337	&0.514	&0.356	&0.242	&0.431	&0.411	&0.361

\\
BLIPScore  &0.550	&0.390	&0.520	&0.443	&0.337	&0.527	&0.502	&0.337

\\
Aesthetic Score &0.076	&0.092	&0.088	&0.074	&0.057	&0.029	&0.361	&0.074

\\
ImageReward &0.677	&0.538	&0.585	&0.610	&0.508	&0.632	&\bf{0.613}	&0.547

\\
DreamReward &0.534	&0.396	&0.333	&0.545	&0.283	&0.324	&0.577	&0.464

\\
HPS v2  &0.497	&0.275	&0.316	&0.420	&0.249	&0.295	&0.340	&0.267

\\
CLIP-IQA &0.223	&0.163	&0.040	&0.179	&0.023	&0.122	&0.094	&0.137
\\
Q-Align	 &0.079	&0.093	&0.096	&0.093	&0.021	&0.231	&0.076	&0.13

\\ 
 
ResNet50 + FT &0.689	&0.508	&0.450	&0.530	&0.430	&0.401	&0.259	&0.222

\\
ViT-B + FT  &0.726	&0.582	&0.475	&0.652	&0.389	&0.396	&0.376	&0.155

\\
SwinT-B + FT &0.722	&0.560	&0.347	&0.649	&0.442	&0.462	&0.148	&0.223
\\
DINO v2 + FT &0.791	&0.633	&0.609	&0.72	&0.501	&0.472	&0.45	&0.259

\\
MultiScore &0.784	&0.670	&0.657	&0.597	&0.672	&0.493	&0.365	&0.415

\\
HyperScore   &\bf{0.846}	&\bf{0.754}	&\bf{0.724}	&\bf{0.816}	&\bf{0.724}	&\bf{0.697}	&0.599	&\bf{0.573}

\\

\bottomrule

\end{tabular}}

\resizebox{0.8\textwidth}{!}{
\begin{tabular}{c|cccccccc}
\toprule
\multirow{2}{*}{Metric} & \multicolumn{8}{c}{Geometry}  \\ \cline{2-9}
&DreamFusion &Magic3D &SJC	&TextMesh	&Consistent3D	&LatentNeRF	&3DTopia &One-2-3-45++\\ \midrule

CLIPScore &0.511	&0.323	&0.493	&0.343	&0.226	&0.418	&0.360	&0.269

\\
BLIPScore  &0.549	&0.356	&0.537	&0.431	&0.328	&0.553	&0.426	&0.316

\\
Aesthetic Score &0.159	&0.156	&0.088	&0.051	&0.015	&0.005	&0.452	&0.367

\\
ImageReward &0.665	&0.461	&0.588	&0.552	&0.433	&0.555	&0.503	&0.421

\\
DreamReward &0.513	&0.369	&0.363	&0.533	&0.237	&0.333	&0.502	&0.394

\\
HPS v2  &0.506	&0.299	&0.311	&0.429	&0.252	&0.346	&0.309	&0.331

\\
CLIP-IQA 	&0.202	&0.099	&0.014	&0.142	&0.026	&0.085	&0.099	&0.078
\\
Q-Align &0.022	&0.048	&0.118	&0.019	&0.01	&0.321	&0.073	&0.449

\\ 
 
ResNet50 + FT &0.707	&0.537	&0.482	&0.545	&0.443	&0.439	&0.429	&0.358

\\
ViT-B + FT  &0.743	&0.563	&0.582	&0.629	&0.375	&0.466	&0.546	&0.215

\\
SwinT-B + FT &0.729	&0.529	&0.382	&0.602	&0.434	&0.525	&0.369	&0.184
\\
DINO v2 + FT &0.775	&0.624	&0.681	&0.73	&0.554	&0.603	&0.69	&0.471

\\
MultiScore &0.741	&0.620	&0.595	&0.585	&0.581	&0.446	&0.588	&0.418

\\
HyperScore  &\bf{0.821}	&\bf{0.717}	&\bf{0.703}	&\bf{0.773}	&\bf{0.657}	&\bf{0.708}	&\bf{0.683}	&\bf{0.653}

\\

\bottomrule

\end{tabular}}

\resizebox{0.8\textwidth}{!}{
\begin{tabular}{c|cccccccc}
\toprule
\multirow{2}{*}{Metric} & \multicolumn{8}{c}{Texture}  \\ \cline{2-9}
&DreamFusion &Magic3D &SJC	&TextMesh	&Consistent3D	&LatentNeRF	&3DTopia &One-2-3-45++\\ \midrule

CLIPScore &0.502	&0.406	&0.509	&0.393	&0.259	&0.434	&0.384	&0.258

\\
BLIPScore  &0.542	&0.400	&0.532	&0.462	&0.330	&0.536	&0.473	&0.298

\\
Aesthetic Score &0.141	&0.122	&0.095	&0.020	&0.003	&0.047	&0.450	&0.356

\\
ImageReward &0.671	&0.489	&0.573	&0.568	&0.435	&0.535	&0.543	&0.420

\\
DreamReward &0.498	&0.377	&0.364	&0.517	&0.251	&0.281	&0.533	&0.364

\\
HPS v2  &0.516	&0.295	&0.330	&0.450	&0.266	&0.307	&0.367	&0.307

\\
CLIP-IQA 	&0.195	&0.101	&0.024	&0.119	&0.006	&0.169	&0.093	&0.007
\\
Q-Align &0.018	&0.072	&0.19	&0.021	&0.024	&0.388	&0.187	&0.44

\\ 
 
ResNet50 + FT &0.702	&0.532	&0.450	&0.553	&0.421	&0.432	&0.395	&0.412

\\
ViT-B + FT  &0.735	&0.606	&0.572	&0.636	&0.392	&0.483	&0.492	&0.220

\\
SwinT-B + FT &0.734	&0.524	&0.374	&0.617	&0.459	&0.519	&0.340	&0.171
\\
DINO v2 + FT &0.792	&0.635	&0.659	&0.736	&0.563	&0.581	&0.663	&0.423

\\
MultiScore &0.755	&0.634	&0.603	&0.598	&0.593	&0.472	&0.610	&0.448

\\
HyperScore   &\bf{0.823}	&\bf{0.736}	&\bf{0.694}	&\bf{0.775}	&\bf{0.702}	&\bf{0.694}	&\bf{0.613}	&\bf{0.623}

\\

\bottomrule

\end{tabular}}

\resizebox{0.8\textwidth}{!}{
\begin{tabular}{c|cccccccc}
\toprule
\multirow{2}{*}{Metric} & \multicolumn{8}{c}{Overall}  \\ \cline{2-9}
&DreamFusion &Magic3D &SJC	&TextMesh	&Consistent3D	&LatentNeRF	&3DTopia &One-2-3-45++ \\ \midrule

CLIPScore &0.506	&0.342	&0.517	&0.362	&0.208	&0.428	&0.399	&0.330

\\
BLIPScore  &0.548	&0.380	&0.553	&0.443	&0.300	&0.543	&0.473	&0.347

\\
Aesthetic Score &0.118	&0.149	&0.066	&0.005	&0.001	&0.029	&0.434	&0.242

\\
ImageReward &0.678	&0.504	&0.602	&0.577	&0.440	&0.572	&0.555	&0.517

\\
DreamReward &0.522	&0.391	&0.356	&0.548	&0.237	&0.332	&0.544	&0.462

\\
HPS v2  &0.507	&0.291	&0.337	&0.436	&0.241	&0.318	&0.336	&0.322

\\
CLIP-IQA 	&0.206	&0.135	&0.023	&0.156	&0.005	&0.111	&0.104	&0.028
\\
Q-Align &0.038	&0.013	&0.154	&0.024	&0.048	&0.297	&0.124	&0.327

\\ 
ResNet50 + FT &0.703	&0.543	&0.470	&0.535	&0.525	&0.432	&0.357	&0.302

\\
ViT-B + FT  &0.731	&0.597	&0.561	&0.635	&0.388	&0.439	&0.460	&0.190

\\
SwinT-B + FT &0.733	&0.567	&0.363	&0.639	&0.441	&0.499	&0.297	&0.230
\\
DINO v2 + FT &0.791	&0.65	&0.686	&0.724	&0.543	&0.567	&0.61	&0.363

\\
MultiScore &0.769	&0.659	&0.635	&0.597	&0.613	&0.451	&0.508	&0.444

\\
HyperScore  &\bf{0.839}	&\bf{0.745}	&\bf{0.730}	&\bf{0.794}	&\bf{0.698}	&\bf{0.718}	&\bf{0.638}	&\bf{0.634}

\\

\bottomrule
\end{tabular}}

\end{table*}

\end{document}